\documentclass{ieeetj}
\usepackage{amsmath,amssymb,amsfonts}
\usepackage{algorithmic}
\usepackage{graphicx,color}
\usepackage{textcomp}
\usepackage{xcolor}
\usepackage{hyperref}
\hypersetup{hidelinks=true}
\usepackage{algorithm,algorithmic}
\def\BibTeX{{\rm B\kern-.05em{\sc i\kern-.025em b}\kern-.08em
    T\kern-.1667em\lower.7ex\hbox{E}\kern-.125emX}}
\AtBeginDocument{\definecolor{tmlcncolor}{cmyk}{0.93,0.59,0.15,0.02}\definecolor{NavyBlue}{RGB}{0,86,125}}

\usepackage[caption=false]{subfig}
\usepackage{array}
\usepackage{booktabs}
\usepackage{tikz}
\usepackage[export]{adjustbox}
\usepackage{multirow}
\usepackage{siunitx}
\usepackage{url}
\usepackage{hyperref}
\usepackage{gensymb}
\usepackage{amsmath, amssymb}
\usepackage{amsfonts}
\usepackage{systeme}
\usepackage{algorithm}
\usepackage{threeparttable} %
\usepackage{multirow}
\usepackage{colortbl}
\usepackage[style=ieee,mincitenames=1,maxcitenames=2,maxbibnames=100]{biblatex}

\newcommand\fig[1]{Fig.~\ref{#1}}
\newcommand\sect[1]{Section~\ref{#1}}
\newcommand\tab[1]{Table~\ref{#1}}

\newcommand\catabot{\emph{Catabot}}
\newcommand{\datasetSize}{\num[group-separator={,}]{11561}} %
\newcommand{\opensetSize}{\num[group-separator={,}]{1376}} %
\newcommand{\closesetSize}{\num[group-separator={,}]{10185}} %

\newcommand\pointpillars{PointPillars}
\newcommand\second{SECOND}
\newcommand\pointRCNN{PointRCNN}
\newcommand\pvRCNN{PV-RCNN}
\newcommand\voxelRCNN{Voxel-RCNN}
\newcommand\tedSingle{TED-S}

\newcommand\pointpainting{PointPainting}
\newcommand\clocs{CLOCs}
\newcommand\focalconvf{Focal Conv-F}
\newcommand\tedMulti{TED-M}

\DeclareSIUnit\rpm{RPM}

\newcommand\arirevised[1]{\textcolor{black}{#1}}
\newcommand\monrevised[1]{\textcolor{black}{#1}}
\newcommand\revised[1]{\textcolor{black}{#1}} %
\newcommand\secrevised[1]{\textcolor{black}{#1}} %
\long\def\invis#1{}

\def\OJlogo{\vspace{-4pt}IEEE Transactions on Field Robotics}
\def\seclogo{\vspace{10pt}IEEE Transactions on Field Robotics}

\def\authorrefmark#1{\ensuremath{^{\textbf{#1}}}}

\newcommand\copyrighttext{%
  \footnotesize \textcopyright 2025 IEEE. Personal use of this material is permitted.
  Permission from IEEE must be obtained for all other uses, in any current or future
  media, including reprinting/republishing this material for advertising or promotional
  purposes, creating new collective works, for resale or redistribution to servers or
  lists, or reuse of any copyrighted component of this work in other works.
  DOI: \href{https://doi.org/10.1109/TFR.2025.3602937}{10.1109/TFR.2025.3602937}
  }
\newcommand\copyrightnotice{%
\begin{tikzpicture}[remember picture,overlay]
\node[anchor=south,yshift=80pt,xshift=75pt] at (current page.south) {\fbox{\parbox{\dimexpr\textwidth-\fboxsep-\fboxrule\relax}{\copyrighttext}}};
\end{tikzpicture}%
}

\begin{document}
\reviseddate{Preprint -- Special Issue on ICRA 2024 Workshop on Field Robotics}
\markboth{}{Jeong and Chadda {et al.}}

\title{SeePerSea: Multi-modal Perception Dataset of In-water Objects for Autonomous Surface Vehicles}

\author{Mingi Jeong\authorrefmark{*1}, Arihant Chadda\authorrefmark{*2}, Ziang Ren\authorrefmark{3}, Luyang Zhao\authorrefmark{1}, Haowen Liu\authorrefmark{4}, \\ Aiwei Zhang\authorrefmark{1},  Yitao Jiang\authorrefmark{1}, Sabriel Achong\authorrefmark{1}, Samuel Lensgraf\authorrefmark{5}, Monika Roznere\authorrefmark{6}, and Alberto Quattrini Li\authorrefmark{1}}
\affil{Department of Computer Science, Dartmouth College, Hanover, NH USA}
\affil{IQT Labs, Tysons, VA USA}
\affil{Department of Computer Science, Columbia University, New York, NY USA}
\affil{Department of Computer Science, University of Maryland College Park, College Park, MD USA}
\affil{The Institute for Human and Machine Cognition and The University of West Florida, Pensacola, FL USA}
\affil{School of Computing, Binghamton University, Binghamton, NY USA}
\corresp{Corresponding author: Mingi Jeong and Alberto Quattrini Li (email: mingi.jeong.gr@dartmouth.edu, alberto.quattrini.li@dartmouth.edu).}
\authornote{$^*$ These authors contributed equally to the paper. This work is supported in part by the Burke Research Initiation Award, NSF CNS-1919647, 2144624, OIA1923004, and NOAA NH Sea Grant.}

\begin{abstract}
This paper introduces the first publicly accessible labeled multi-modal perception dataset for autonomous maritime navigation, focusing on in-water obstacles within the aquatic environment to enhance situational awareness for Autonomous Surface Vehicles (ASVs). This dataset, collected over 4 years and consisting of diverse objects encountered under varying environmental conditions, aims to bridge the research gap in ASVs by providing a multi-modal, annotated, and ego-centric perception dataset, for object detection and classification. We also show the applicability of the proposed dataset by training \revised{and testing current} deep learning-based open-source perception algorithms that have shown success in the autonomous ground vehicle domain. \revised{With the training and testing results, we discuss open challenges for existing datasets and methods, identifying future research directions.} We expect that our dataset will contribute to the development of future marine autonomy pipelines and marine (field) robotics. This dataset is open source and found at  \url{https://seepersea.github.io/}.
\end{abstract}


\maketitle
\copyrightnotice

\section{INTRODUCTION}
\begin{figure}[t!]
    \centering
    \includegraphics[width=0.9\columnwidth]{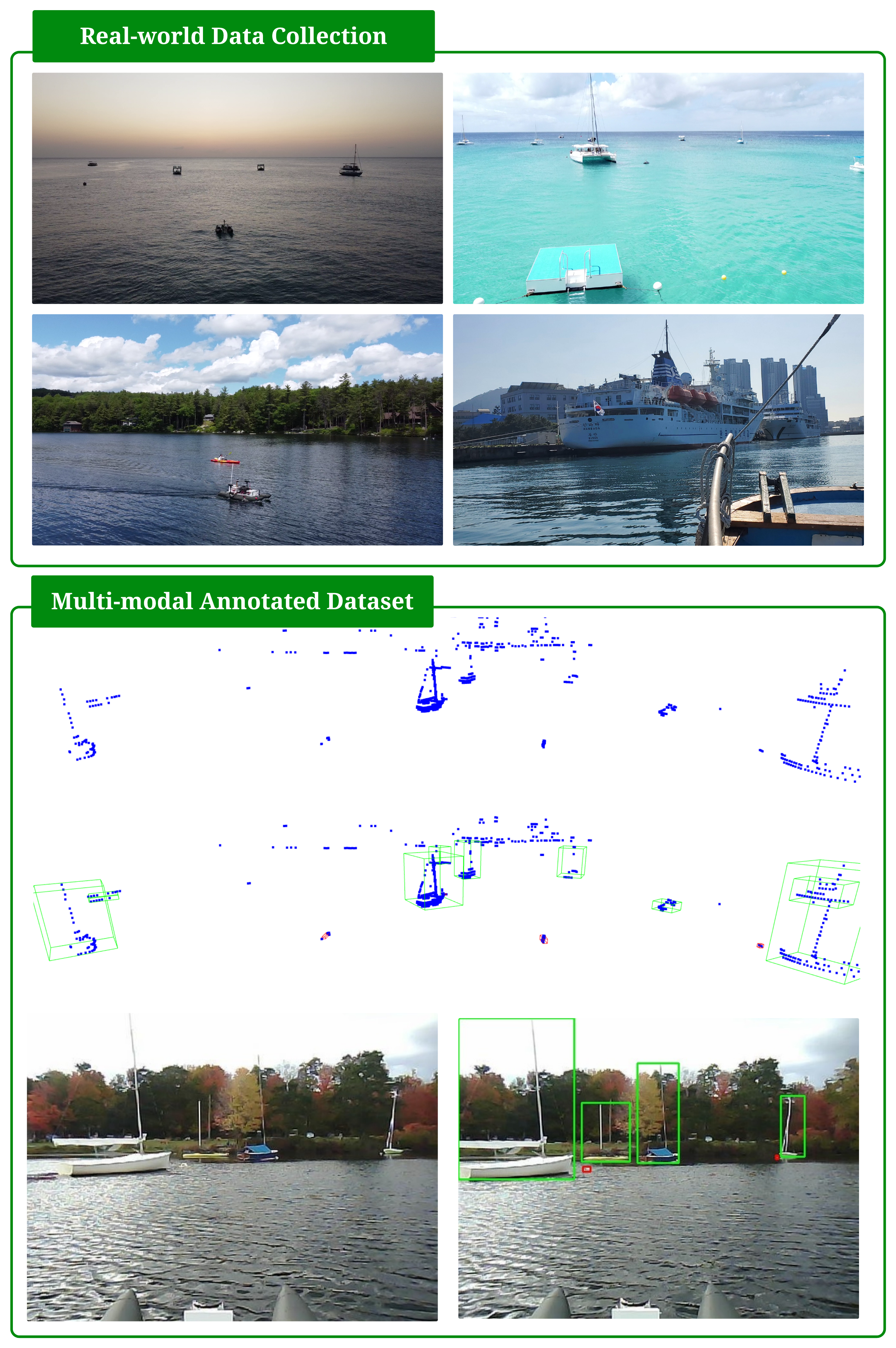}
    \caption{Real-world data collection of in-water objects by ASV and human-driven boat in operation at different geographic locations and conditions. We provide multi-modal annotated dataset (LiDAR and RGB camera) for marine autonomy.}
    \label{fig:beauty}
\end{figure}

\IEEEPARstart{L}earning-based, multi-modal algorithms have shown terrestrial domain success for self-driving cars on the road to autonomy. The precondition(s) to this success fundamentally rest on the availability of relevant, labeled datasets \cite{KITTI-raw-2013, waymo-2020, nuscenes-2019}. Equivalent success in marine Autonomous Surface Vehicles (ASVs) is, unsurprisingly, hampered by the lack of relevant multi-modal perception datasets. Thus, the goal of this paper is to \textbf{create the first publicly available labeled, multi-modal 3D perception dataset for autonomous maritime navigation} (\fig{fig:beauty}). This dataset, consisting of in-water obstacles, aims to enhance ASVs' situational awareness. Situational awareness is a foundational task that undergirds autonomy, which is increasing in importance given the focus on ASVs for tasks such as environmental monitoring and automated transportation. This importance will only grow as marine trade increases to $90\%$ of {the share of} world trade \cite{UN-2022} and, accordingly, the expected size of the ASV market will grow to $2.7B$ USD by $2032$ \cite{USV-2024}. 

\revised{Understanding the locations of static and dynamic objects in the aquatic domain (\textbf{object detection}) and determining the types of these objects (\textbf{object classification}) are crucial tasks for \emph{data association}---to understand the speed and heading of approaching objects. Such processes are integral for \emph{navigational decision-making}, i.e., collision avoidance. However, aquatic domain challenges, including (1) unstructured navigational environments and (2) the limited maneuverability of marine vehicles, raise the importance of \textit{early} and \textit{accurate} state estimation of in-water obstacles for safe and efficient navigation that minimize detection errors (e.g., false negatives). Among human error-driven marine accidents, over $70\%$ are attributed to improper situational awareness \cite{humanerror-review2021}. Consequently, marine vehicles, even   human-driven vessels, naturally rely on \textbf{multi-modal} data for situational awareness, which aligns with the regulations (e.g., rule 5 \textit{look-out})  explicitly covered by the maritime Rules of the Road \cite{colreg}}.

\revised{
The scarcity of multimodal labeled 3D perception datasets for ASVs is attributed to the high operational costs and the extensive labeling effort required \cite{massmind-2023}. Among the few existing datasets in the aquatic domain, the open-source  ones primarily consist of either (1) \textbf{single-modality} data that is typically image-based \cite{vais-2015, MARVEL-2016, seaships-2018, wsodd-2019, seasaw-2022, MODD1-2015, MODD2-2018, MODS-2022}, or (2) multiple modalities but lacking \textbf{object labels} across modalities~\cite{lin-maritime-lidar-2022, comp-maritime-review-2024}, which are essential for ground-truth evaluation \cite{MIT-data2021}. This absence of multi-modal  and ground-truth annotations significantly hinders the development of crucial ASV capabilities, as noted in \cite{MIT-data2021, MODS-2022}}.

\revised{Accordingly, we release the first multi-modal labeled maritime dataset.} \monrevised{Our dataset includes expeditions} \arirevised{from 2021 to 2024 using our ASV platform \catabot{} and a human-driven vessel} in different locations (United States, Barbados, and South Korea) covering various environments (both sea and fresh water), conditions (e.g., dusk, daylight), and encounters (e.g., head-on, crossing) with various objects. The proposed dataset includes navigation-oriented \arirevised{three class (ship, buoy, and other) labeled objects} \arirevised{for detection and classification}. \monrevised{We selected these labels} according to the international traffic rule \cite{colreg} and buoyage system \cite{Admiralty_2018}. \monrevised{In summary, } the dataset is composed of \datasetSize~frames of LiDAR \arirevised{point clouds} and RGB images. We also \arirevised{demonstrate} the \arirevised{utility} of the proposed dataset using deep learning-based open-source perception algorithms -- both single-modality and multi-modal fusion -- that have shown success \arirevised{in the terrestrial domain}, with both quantitative and qualitative evaluations: highlighting success in some scenarios, but also current gaps.

We release our dataset publicly  (\url{https://seepersea.github.io/}) for the community \arirevised{and expect it} will have the following contributions:
\begin{itemize}
    \item SeePerSea, the first LiDAR-camera dataset in aquatic environments with object labels across two modalities, will foster the development of robust fusion perception pipelines for ASV autonomy.
    \item SeePerSea, covering various environments and day conditions, will help ensure that developed perception pipelines are increasingly generalizable.
\end{itemize}

Overall, the SeePerSea dataset will contribute to the development of \arirevised{state-of-the-art} marine autonomy pipeline\arirevised{s} and \arirevised{accelerate the future of} marine (field) robotics.

The structure of this paper is as follows. \sect{sec:related-work} discusses datasets both in the ground and maritime domains. \sect{sec:dataset-generation} describes how the data was collected, annotated, and structured. \sect{sec:dataset-characteristic} provides an analysis of the dataset characteristic. \sect{sec:benchmarks} presents the results from current deep learning pipelines trained on the provided dataset and \sect{sec:discussion} discusses lessons learned and current gaps. Finally, \sect{sec:conclusion} summarizes the paper and highlights future work.  

\begin{table*}
\centering
\caption{Comparison of the state-of-the-art dataset in the maritime domain.}
\label{tab:comparison}
\resizebox{\textwidth}{!}{%
\begin{threeparttable}
\begin{tabular}{|l|cc|c|c|cc|c|c|}
\hline
\textbf{Dataset} &
  \multicolumn{2}{c|}{\textbf{Modality}} &
  \textbf{\begin{tabular}[c]{@{}c@{}}Object\\ Label\end{tabular}} &
  \textbf{\begin{tabular}[c]{@{}c@{}}On-board\\ Data\end{tabular}} &
  \multicolumn{2}{c|}{\textbf{Area}} &
  \textbf{Application} &
  \textbf{Sensors} \\ \cline{2-3} \cline{6-7}
 &
  \multicolumn{1}{c|}{\textbf{Image}} &
  \textbf{Range} &
   &
   &
  \multicolumn{1}{c|}{\textbf{Coastal}} &
  \textbf{Fresh} &
   &
   \\ \hline \hline
\raggedright MassMIND \cite{massmind-2023} &
  \multicolumn{1}{c|}{Y} &
   &
  Y &
  Y &
  \multicolumn{1}{c|}{Y} &
  Y &
  Object Segmentation &
  IR cam \\ \hline
\raggedright \begin{tabular}[c]{@{}l@{}}MaSTr1325,\\ MODD \cite{MaSTr1325-2019, MODD1-2015, MODD2-2018, MODS-2022}\end{tabular} &
  \multicolumn{1}{c|}{Y} &
   &
  Y &
  Y &
  \multicolumn{1}{c|}{Y} &
   &
  Object Segmentation &
  RGB cam, IMU \\ \hline
\raggedright VAIS \cite{vais-2015} &
  \multicolumn{1}{c|}{Y} &
   &
  Y &
   &
  \multicolumn{1}{c|}{Y} &
   &
  Object Classification &
  IR cam, RGB cam \\ \hline
\raggedright MARVEL \cite{MARVEL-2016} &
  \multicolumn{1}{c|}{Y} &
   &
  Y &
   &
  \multicolumn{1}{c|}{N/A*} &
  N/A &
  Object Classification &
  RGB cam \\ \hline
\raggedright SeaShips \cite{seaships-2018} &
  \multicolumn{1}{c|}{Y} &
   &
  Y &
   &
  \multicolumn{1}{c|}{Y} &
   &
  \textbf{\begin{tabular}[c]{@{}c@{}}Object Detection, \\ Object Classification\end{tabular}} &
  RGB cam \\ \hline
\raggedright WSODD \cite{wsodd-2019} &
  \multicolumn{1}{c|}{Y} &
   &
  Y &
   &
  \multicolumn{1}{c|}{Y} &
  Y &
  \textbf{\begin{tabular}[c]{@{}c@{}}Object Detection,\\ Object Classification\end{tabular}} &
  RGB cam \\ \hline
\raggedright USVInland \cite{Cheng2021AreWR} &
  \multicolumn{1}{c|}{Y} &
  Y &
   &
  Y &
  \multicolumn{1}{c|}{} &
  Y &
  \begin{tabular}[c]{@{}c@{}}SLAM, \\ Water segmentation, \\ Stereo matching\end{tabular} &
  \begin{tabular}[c]{@{}c@{}}LiDAR, Stereo cam, \\ RADAR, IMU\end{tabular} \\ \hline
\raggedright NTNU \cite{hetero-track-2022} &
  \multicolumn{1}{c|}{Y} &
  Y &
   &
  Y &
  \multicolumn{1}{c|}{Y} &
   &
  Object Tracking** &
  LiDAR, RADAR, EO and IR cam \\ \hline
\raggedright Pohang \cite{Pohang-2023} &
  \multicolumn{1}{c|}{Y} &
  Y &
   &
  Y &
  \multicolumn{1}{c|}{Y} &
   &
  SLAM &
  \begin{tabular}[c]{@{}c@{}}LiDAR, Stereo cam, \\ AHRS, GPS, \\ IR cam, RADAR\end{tabular} \\ \hline
\raggedright \textbf{Ours} &
  \multicolumn{1}{c|}{\textbf{Y}} &
  \textbf{Y} &
  \textbf{Y} &
  \textbf{Y} &
  \multicolumn{1}{c|}{\textbf{Y}} &
  \textbf{Y} &
  \textbf{\begin{tabular}[c]{@{}c@{}}Object Detection, \\ Object Classification\end{tabular}} &
  \textbf{\begin{tabular}[c]{@{}c@{}}LiDAR, RGB cam, \\ IMU, GPS\end{tabular}} \\ \hline
\end{tabular}
\begin{tablenotes}
\footnotesize
\item[*] The images contain ships but collected by data mining from web sources.
\item[**] The public data contains trajectories of detected vehicles, not the raw data of sensors.
\end{tablenotes}
\end{threeparttable}
}
\end{table*}

\section{Related Work}\label{sec:related-work}

\arirevised{Self-driving car d}  atasets focused on 3D perception, including \cite{KITTI-raw-2013, nuscenes-2019, waymo-2020},   \arirevised{have been} crucial for progress in \arirevised{terrestrial} robotic perception, especially for tasks like object detection, classification, segmentation, and tracking. These collections frequently feature a range of sensor\arirevised{s}  , employing either individual or combined data from cameras, LiDAR, and RADAR.   \arirevised{Given the importance of these datasets}, \monrevised{there is a push} to develop specialized datasets for the marine domain to support the advancement of marine autonomy.

Maritime object detection and classification datasets mainly consist of a \textbf{single sensor modality}, i.e., camera sensors,   \arirevised{used for} different purposes. Key datasets include the first visible and infrared ship image  \arirevised{dataset} for autonomous navigation compliance \cite{vais-2015}, a large-scale maritime dataset with over 2 million images detailing vessel information from a community site \cite{MARVEL-2016}, and a dataset of common ship types from coastal surveillance \cite{seaships-2018}.
\cite{wsodd-2019} introduced more variety with different water surface objects. However, most datasets were from stationary platforms, not from an \textbf{ego-centric perspective}. A significant onboard camera dataset exists \cite{seasaw-2022} but is not public. Public datasets \cite{MaSTr1325-2019, MODD1-2015, MODD2-2018, MODS-2022} consist of several annotated videos collected by a real ASV platform, but these primarily focus on object segmentation with four classes -- sea (water), sky, environment, obstacle -- lacking differentiation of in-water objects like buoys and ships. \cite{massmind-2023} presents a Long Wave Infrared (LWIR) dataset with categories including sky, water, obstacle, but still limited to a \textbf{single modality}.

Several \textbf{multi-modal} datasets \cite{Cheng2021AreWR, Pohang-2023, hetero-track-2022} are available, targeting different aspects of marine perception but not directly focusing on \textbf{object detection} and \textbf{classification}. \cite{Cheng2021AreWR} covers inland waterway scenes using LiDAR, stereo cameras, RADAR, GPS, and IMUs,   \arirevised{for} water segmentation, SLAM, and stereo matching. \cite{hetero-track-2022} combines data from 10 cameras, RADAR, and LiDAR for object tracking. \cite{Pohang-2023} collects data from a diverse set of sensors over a \SI{7.5}{km} route, aiming at SLAM and docking. \tab{tab:comparison} provides an overview of the discussed datasets compared to ours. This lack of datasets in the marine domain, specifically missing   \arirevised{the key} situational awareness \arirevised{tasks previously described}, hampers progress in marine autonomy  .

\begin{figure*}[]
    \begin{minipage}{0.49\columnwidth}
        \subfloat{
            \includegraphics[width=\textwidth]{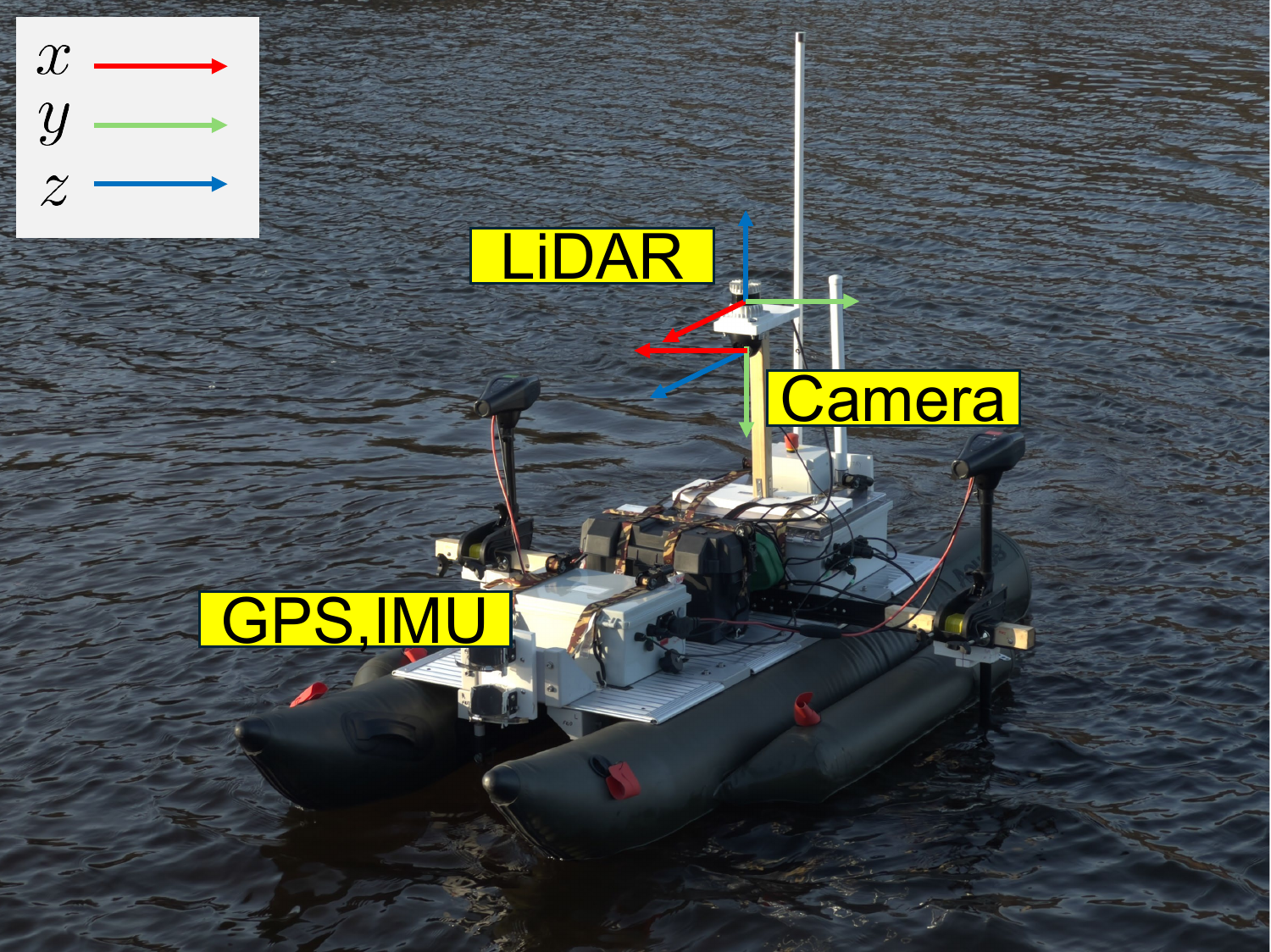}
        }
    \end{minipage}
    \begin{minipage}{0.49\columnwidth}
        \subfloat{

            \includegraphics[width=\textwidth]{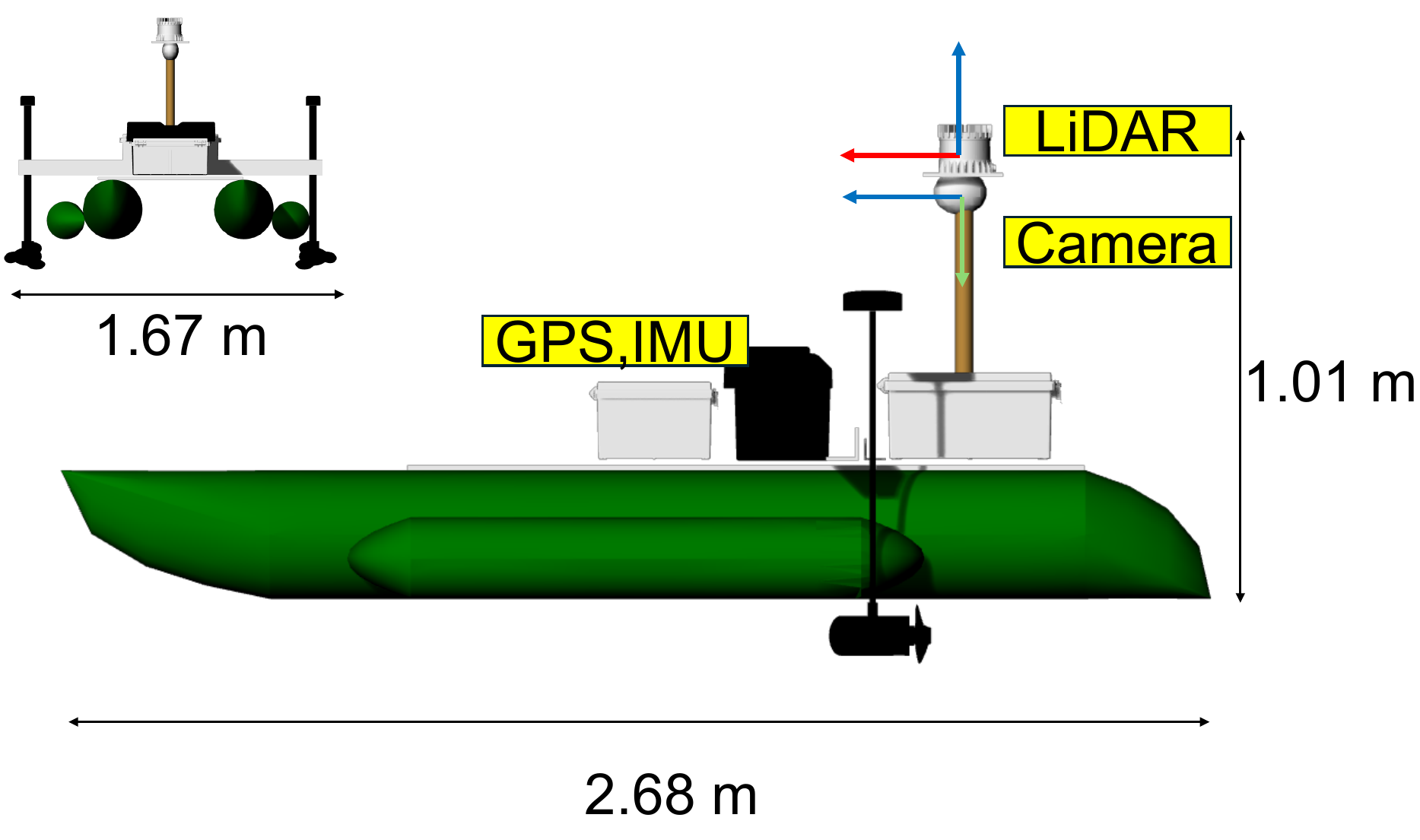}
        }
    \end{minipage}
    \begin{minipage}{0.54\columnwidth}
        \subfloat{

            \includegraphics[width=\textwidth]{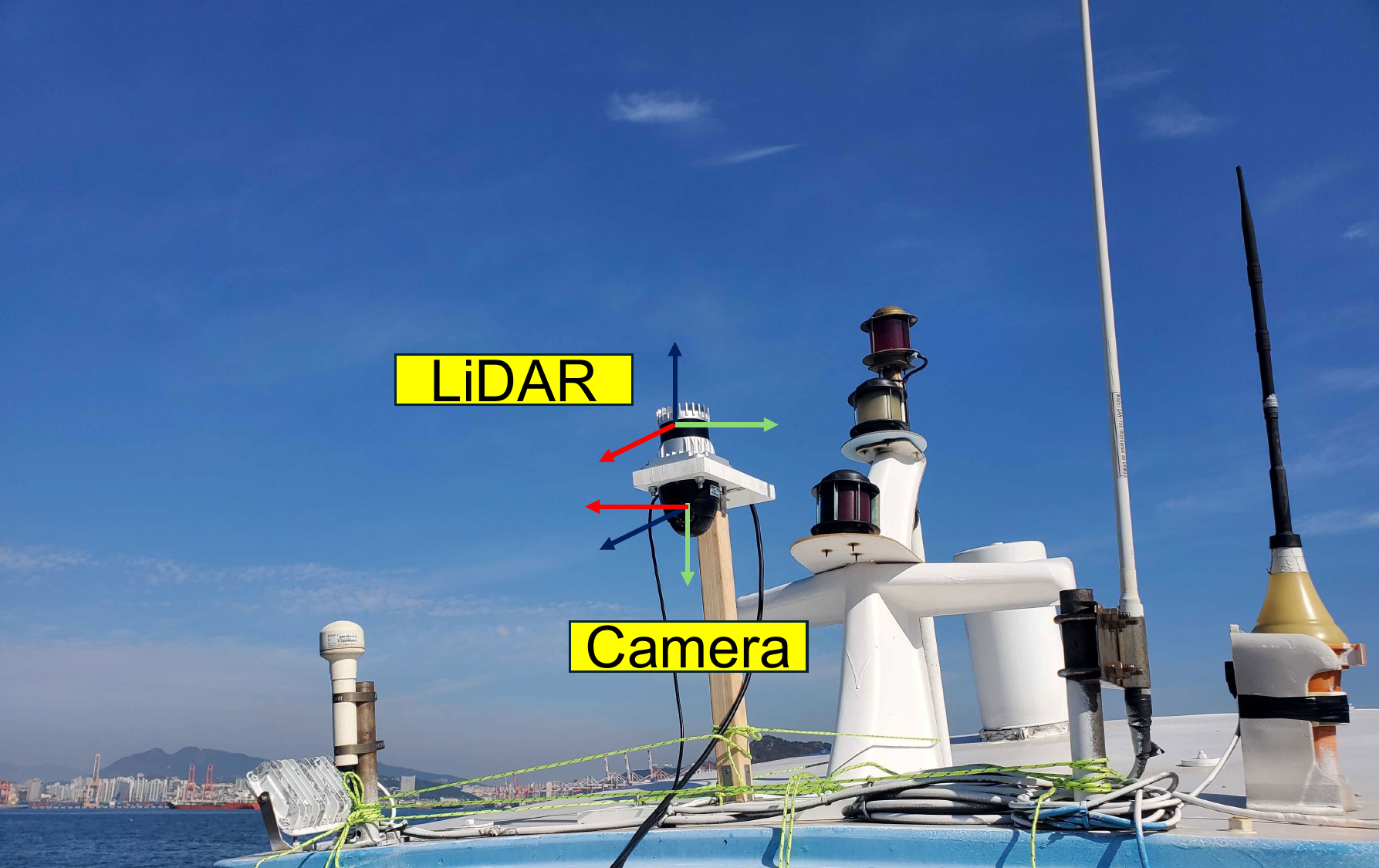}
        }
    \end{minipage}
    \begin{minipage}{0.45\columnwidth}
        \subfloat{

            \includegraphics[width=\textwidth]{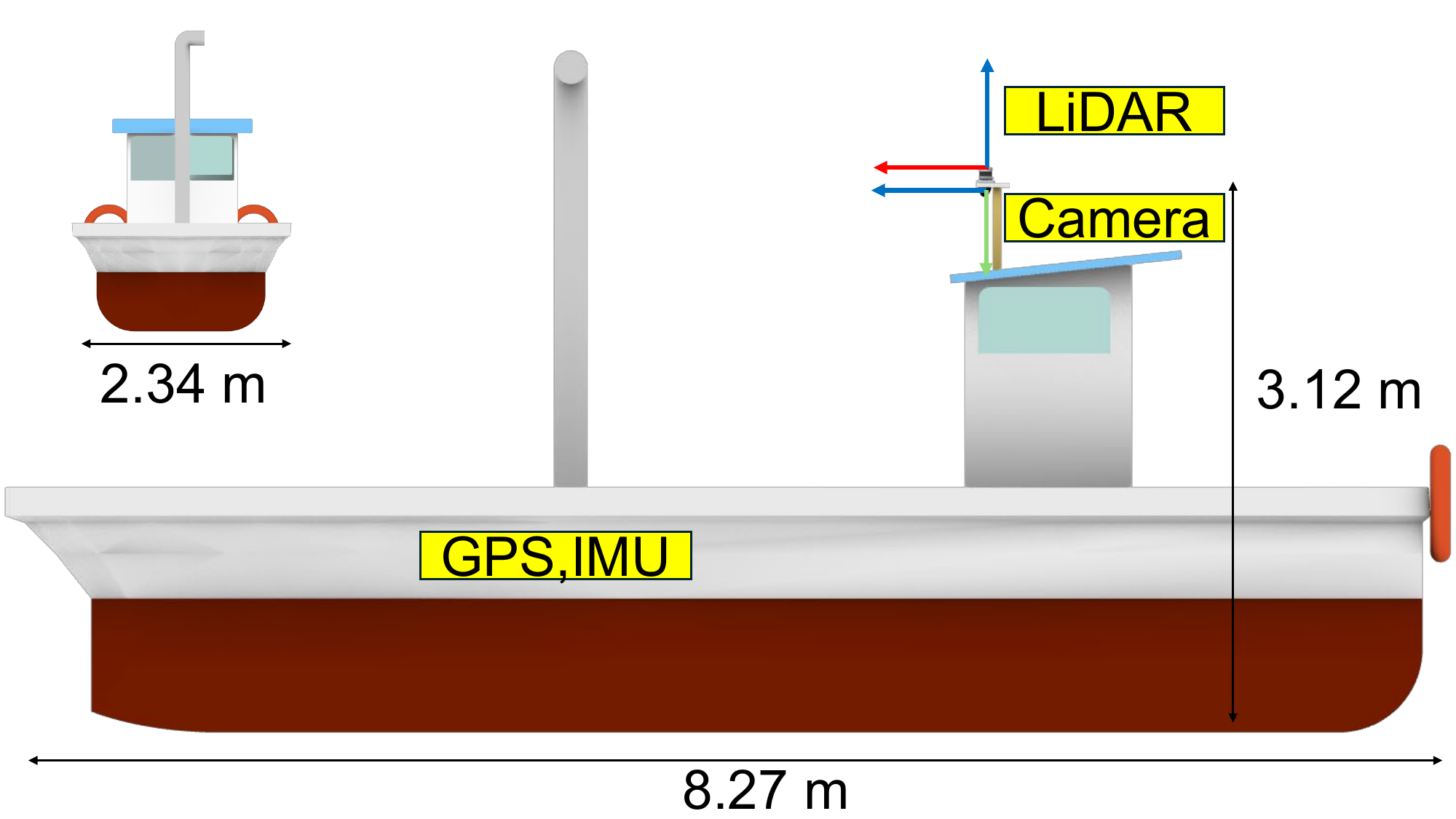}
        }
    \end{minipage}
        \begin{minipage}{0.49\columnwidth}
        \subfloat{

            \includegraphics[width=\textwidth]{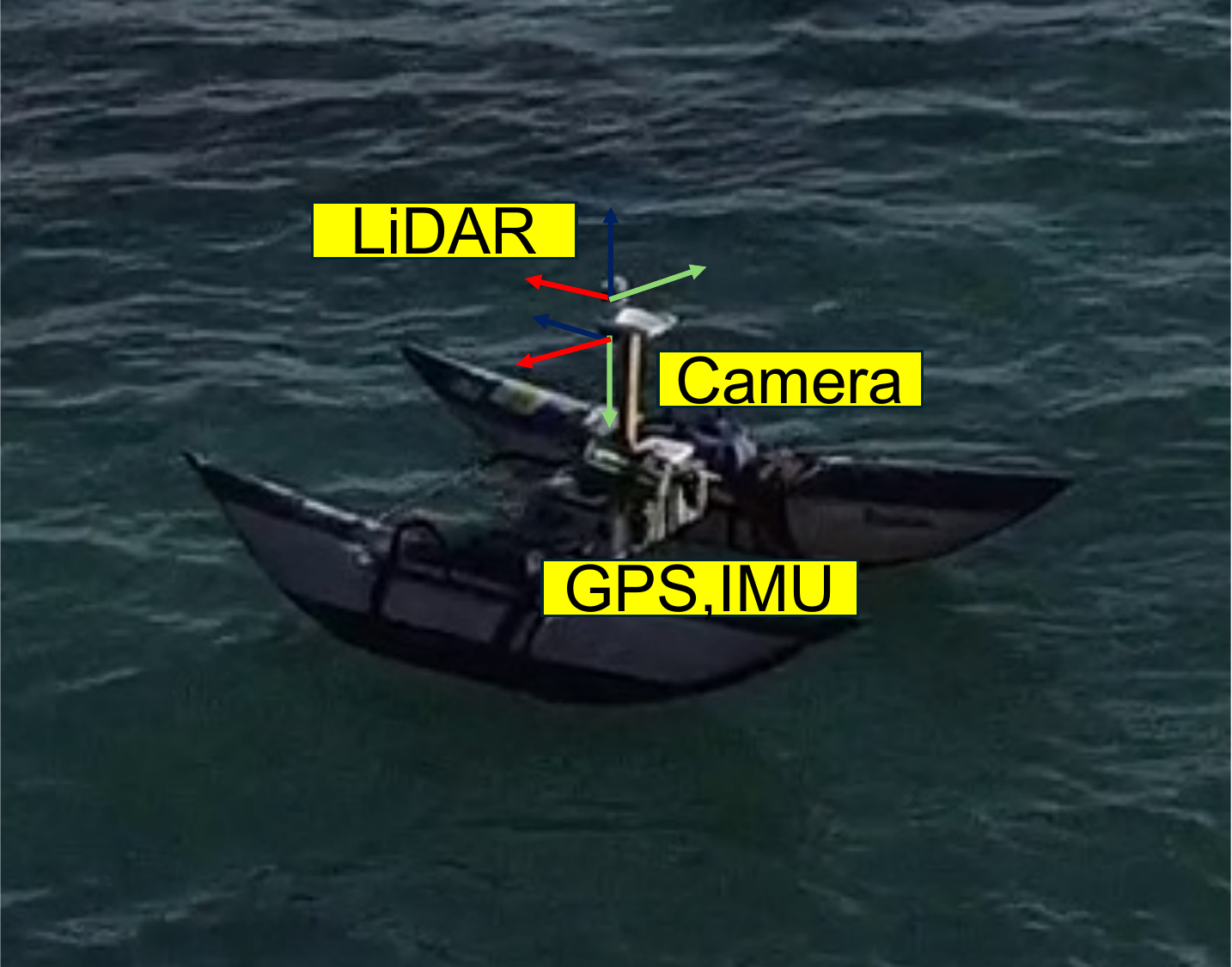}
        }
    \end{minipage}
    \begin{minipage}{0.52\columnwidth}
        \subfloat{

            \includegraphics[width=\textwidth]{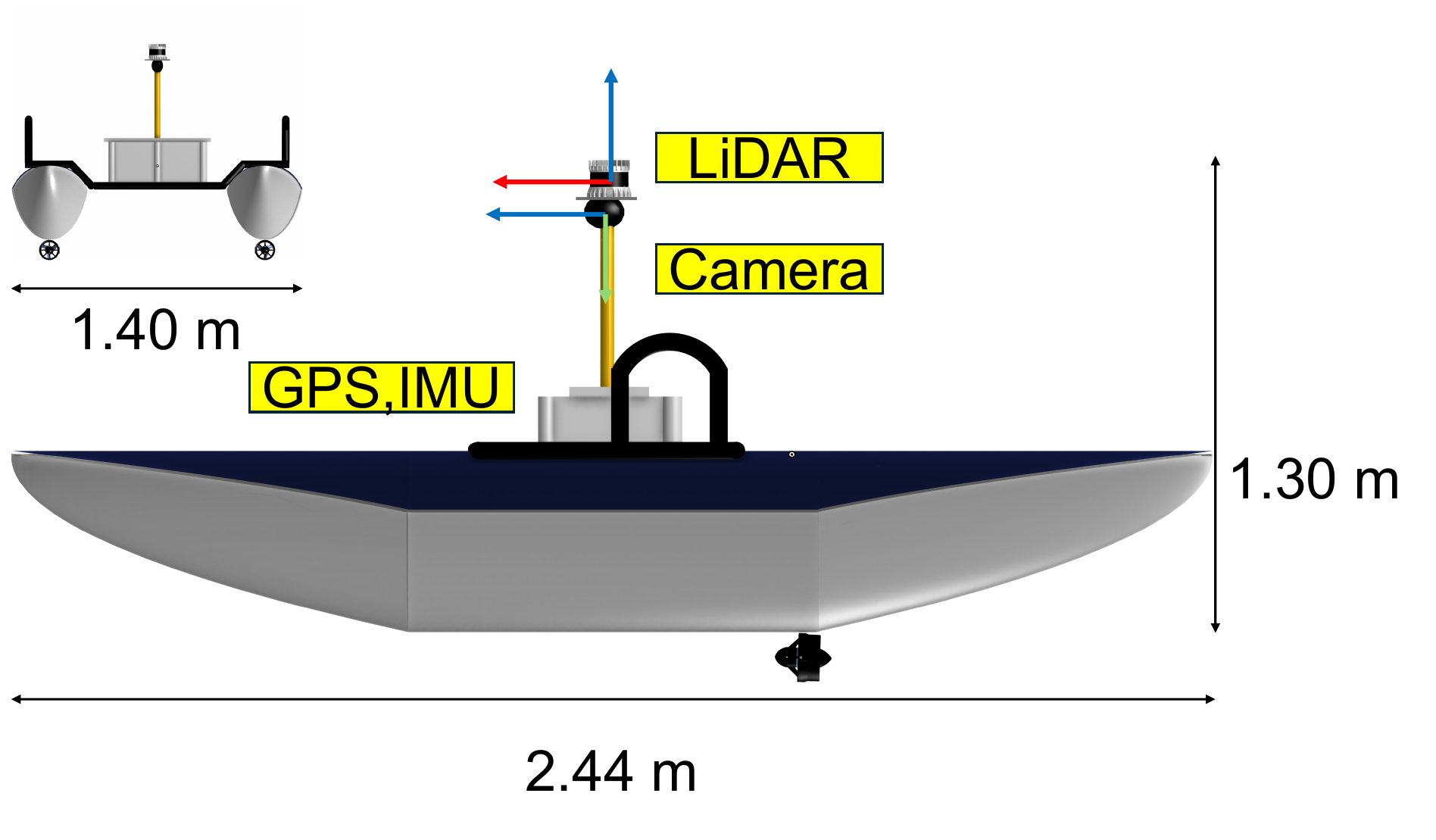}
        }
    \end{minipage}
    \begin{minipage}{0.54\columnwidth}
        \subfloat{

            \includegraphics[width=0.8\textwidth]{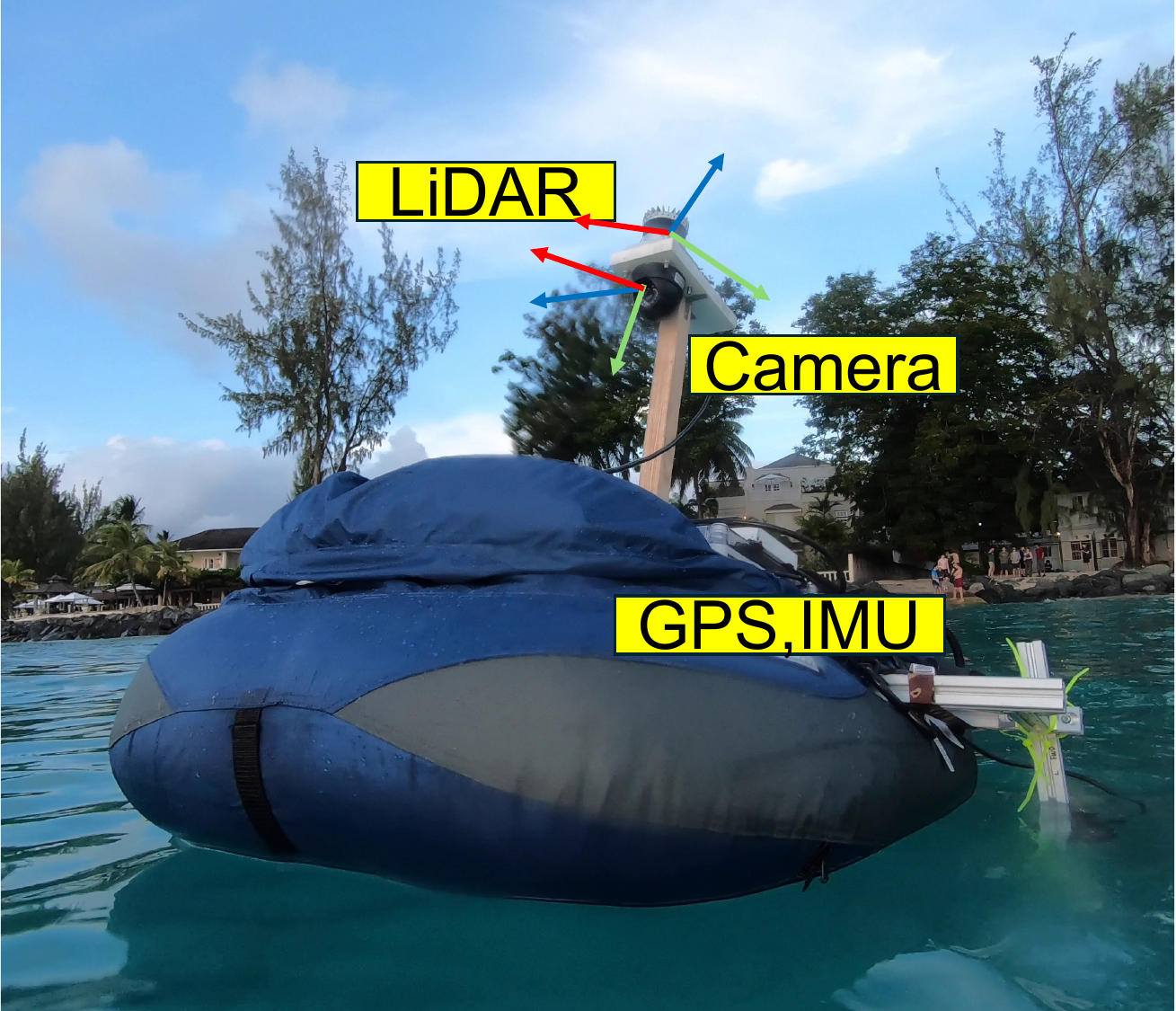}
        }
    \end{minipage}
    \begin{minipage}{0.45\columnwidth}
        \subfloat{

            \includegraphics[width=\textwidth]{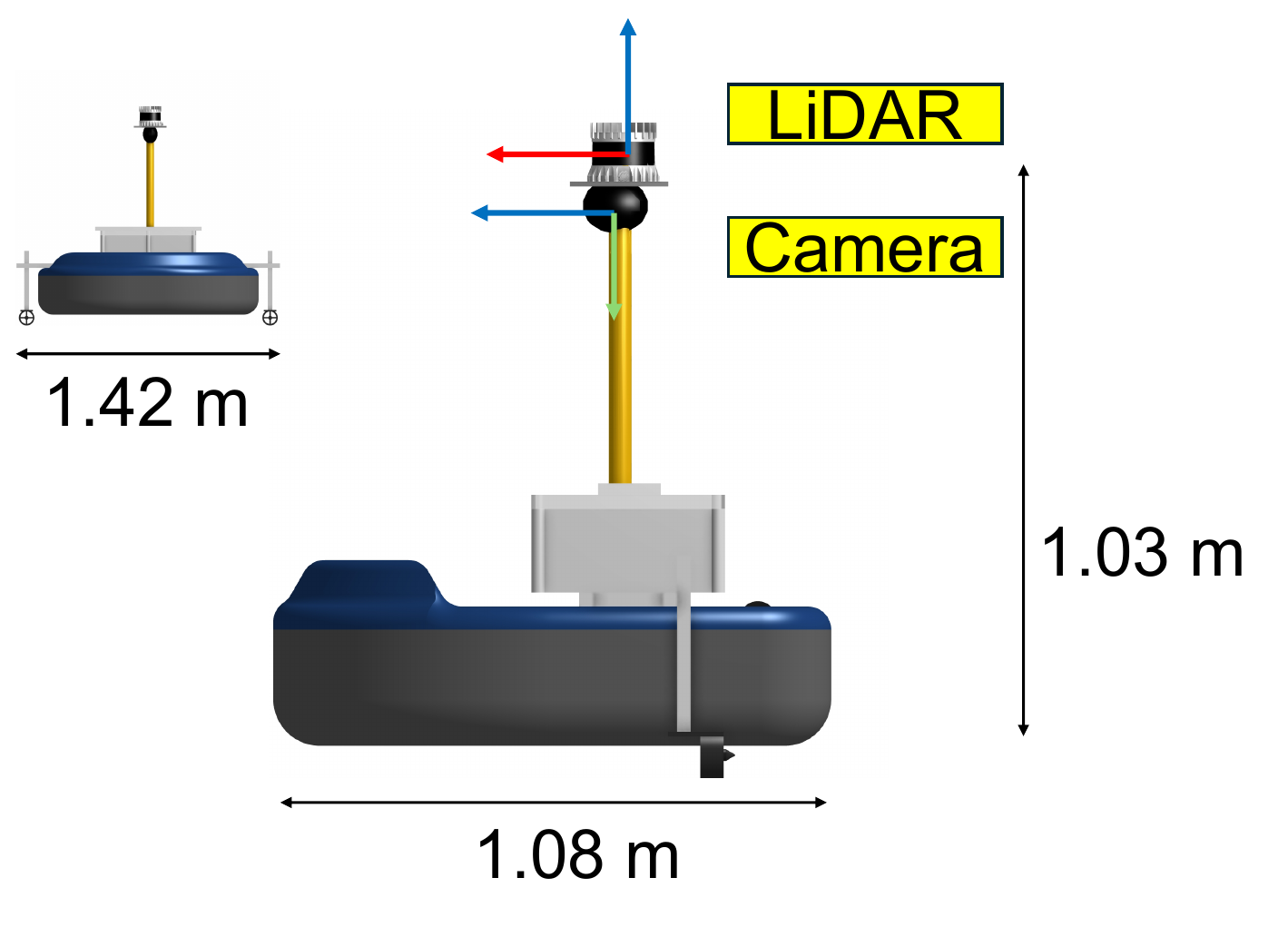}
        }
    \end{minipage}
    \caption{Data collection platform (\textit{top left}): our custom ASV \catabot 2, (\textit{top right}): human-driven ship equipped with sensors,  (\textit{bottom left}): our custom ASV \catabot 1, (\textit{bottom right}): our custom ASV \catabot 5.}
    \label{fig:asv-config}
\end{figure*}
\section{Dataset Generation}\label{sec:dataset-generation}

\subsection{Sensor Configurations}
As shown in \fig{fig:asv-config}, we used our custom ASV \catabot~(in three different configurations) and a human-driven boat installed with a sensor platform. The different configurations allow us to collect diverse data that includes different vehicle dynamics. 
The \catabot~dimensions range from $\SI{1.08}{\meter}$ to $\SI{2.68}{\meter}$ long, and from  $\SI{1.40}{\meter}$ to $\SI{1.67}{\meter}$ wide. The human-driven boat is $\SI{8.27}{\meter}$ long, $\SI{2.34}{\meter}$ wide. \monrevised{Both include a} Global Positioning System (GPS) / Compass and Inertial Measurement Unit (IMU) with a flight controller unit, installed   \arirevised{at} the center line of the vehicle, to record proprioceptive data. We used a low-cost u-blox M8N GPS$/$Compass module.   \arirevised{T}he flight controller hardware \arirevised{we used w}as \arirevised{a} \textit{Pixhawk 4} \arirevised{coupled with a} 32-Bit Arm Cortex-M7 \arirevised{microcontroller}  \arirevised{with a} $\SI{216}{\mega\Hz}$  \arirevised{clock speed and} $\SI{2}{\mega\byte}$ \arirevised{of flash} memory  \arirevised{and} $\SI{512}{\kilo\byte}$ \arirevised{of} RAM.

For exteroceptive data, we installed a RGB camera (Full-HD 1080P with CMOS OV2710 image sensor that can support Infrared (IR) during the nighttime) and a $64$ channel LiDAR (Ouster OS1-64 gen2). The two exteroceptive sensors were \monrevised{located}   \arirevised{at} the center line of the vehicles to ensure a sufficient horizontal field of view (Camera -- \SI{91.8}{\degree}; LiDAR -- \SI{360}{\degree} except for the blind sector \arirevised{due to occlusion caused} by the vehicle structure) and vertical field of view (Camera -- $\SI{75.5}{\degree}$; LiDAR -- \SI{45}{\degree}). The LiDAR has a range of $\SI{120}{\meter}$ with \arirevised{a} horizontal resolution \arirevised{of} $\SI{0.35}{\degree}$ and vertical resolution \arirevised{of} $\SI{0.7}{\degree}$, while the camera sensor \monrevised{has a} $\qtyproduct{640 x 480}{}$ \arirevised{pixel} resolution. 

\begin{figure}[b!]
\centering
        \subfloat[]{
            \includegraphics[height=1.2in]{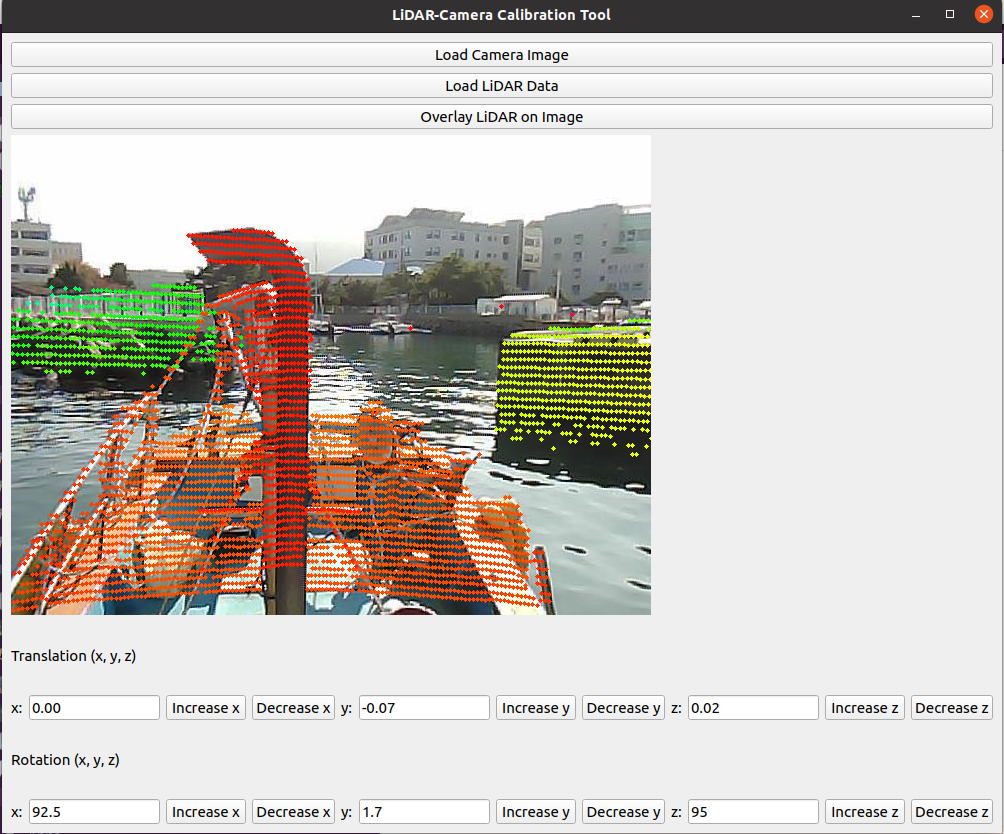}
            \label{fig:calib-check}
        }
    \centering
        \subfloat[]{
            \includegraphics[height=1.2in]{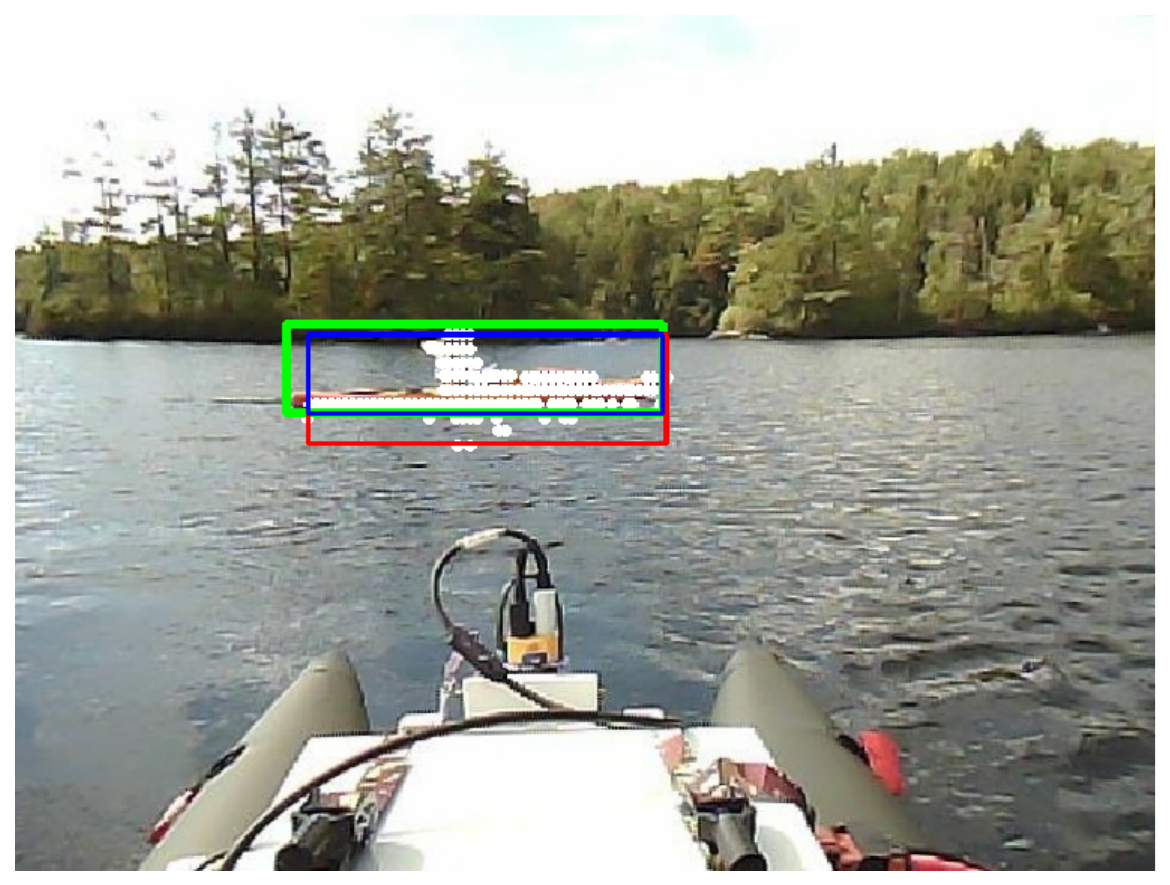}
            \label{fig:tool-overlay}
        }
    \caption{Sensor suite calibration and annotation checking tool. (a) LiDAR and camera extrinsic calibration; and (b) a point cloud (white) overlaid on the corresponding RGB image to check consistency over labels (green: image label, red: point cloud label, blue: intersection).}
    \label{fig:calibration-tool}
\end{figure}

We performed   intrinsic calibration \arirevised{of each sensor} and an extrinsic calibration between camera and LiDAR based on \cite{extrinsic-2022, auto-extrinsic-2021}. We provide a custom tool for checking the extrinsic calibration parameters and overlay of multi-modal data as shown in \fig{fig:calib-check}. We report the result of the calibration parameters per each sequence of the dataset. 

\subsection{Data Collection and Processing}

We used a companion computer system (Intel NUC) and recorded proprioceptive (GPS, Compass, IMU) and exteroceptive (RGB camera, LiDAR) data via the Robot Operating System (ROS). Our Intel NUC computer with Ubuntu 18.04 installed has an Intel Core i7-8559U Processor (8M Cache, up to $\SI{4.50}{\giga\Hz}$\arirevised{)} with $\SI{1}{\tera\byte}$ \arirevised{of} storage. \monrevised{The heterogeneous sensors operate} at different time frequencies: \arirevised{we used a} camera with a frequency of $\SI{30}{\Hz}$ and LiDAR with a frequency of $\SI{10}{\Hz}$.

\begin{figure*}[]
    \begin{minipage}{0.49\columnwidth}
        \subfloat[Geographic locations]{
            \includegraphics[width=\textwidth]{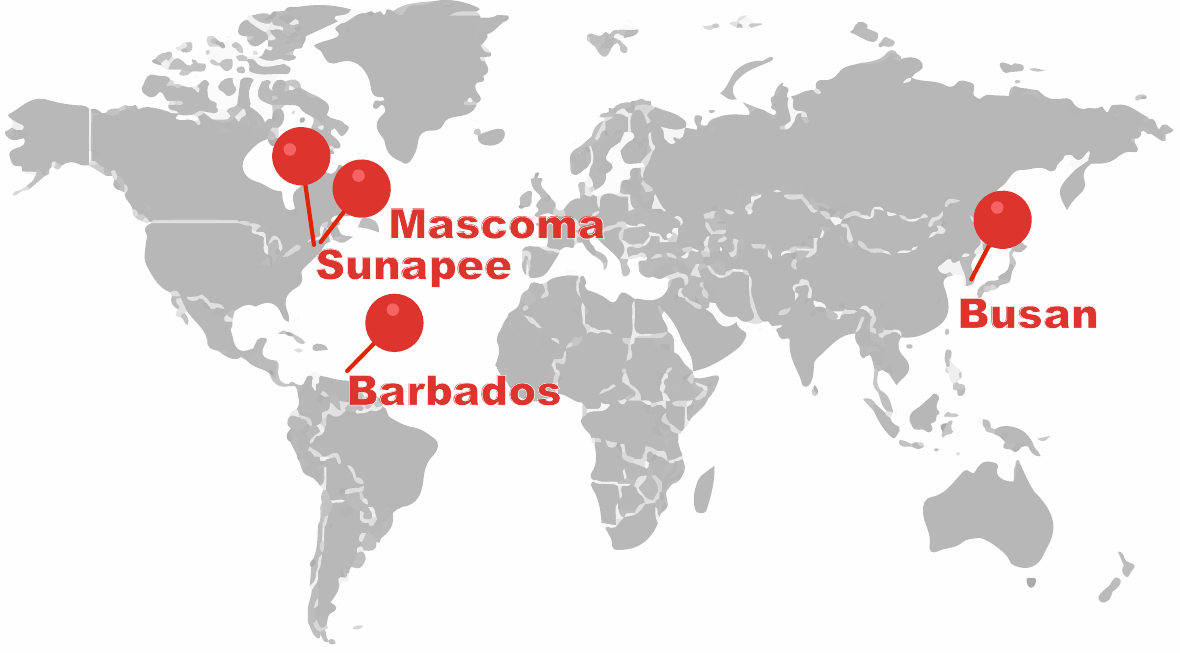}
        }
    \end{minipage}
    \hfill
    \begin{minipage}{0.40\columnwidth}
        \subfloat[Sea -- Barbados 1]{
            \includegraphics[width=\textwidth]{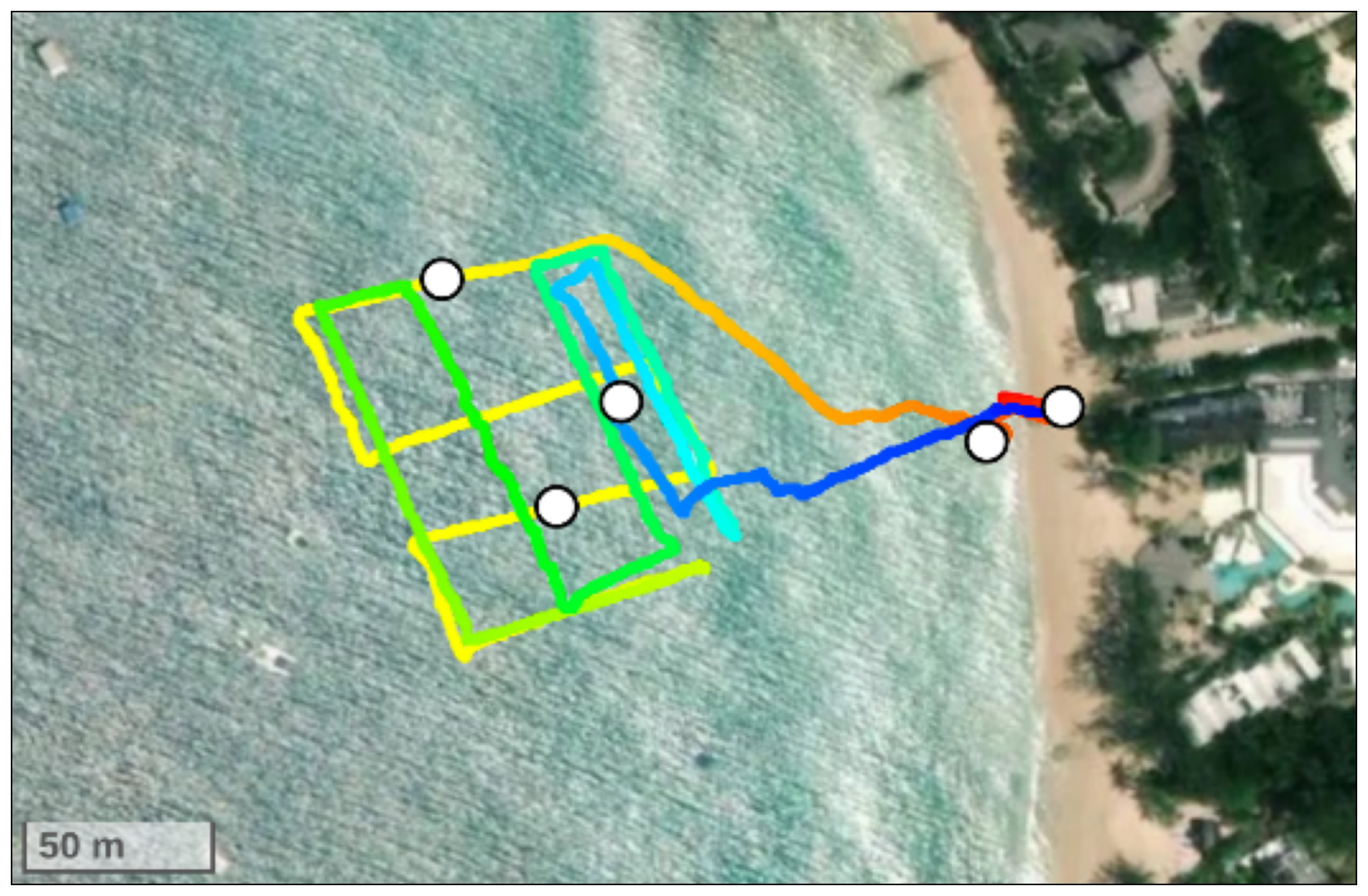}
        }
    \end{minipage}
    \hfill
    \begin{minipage}{0.535\columnwidth}
        \subfloat[Lake -- Mascoma 1]{
            \includegraphics[width=\textwidth]{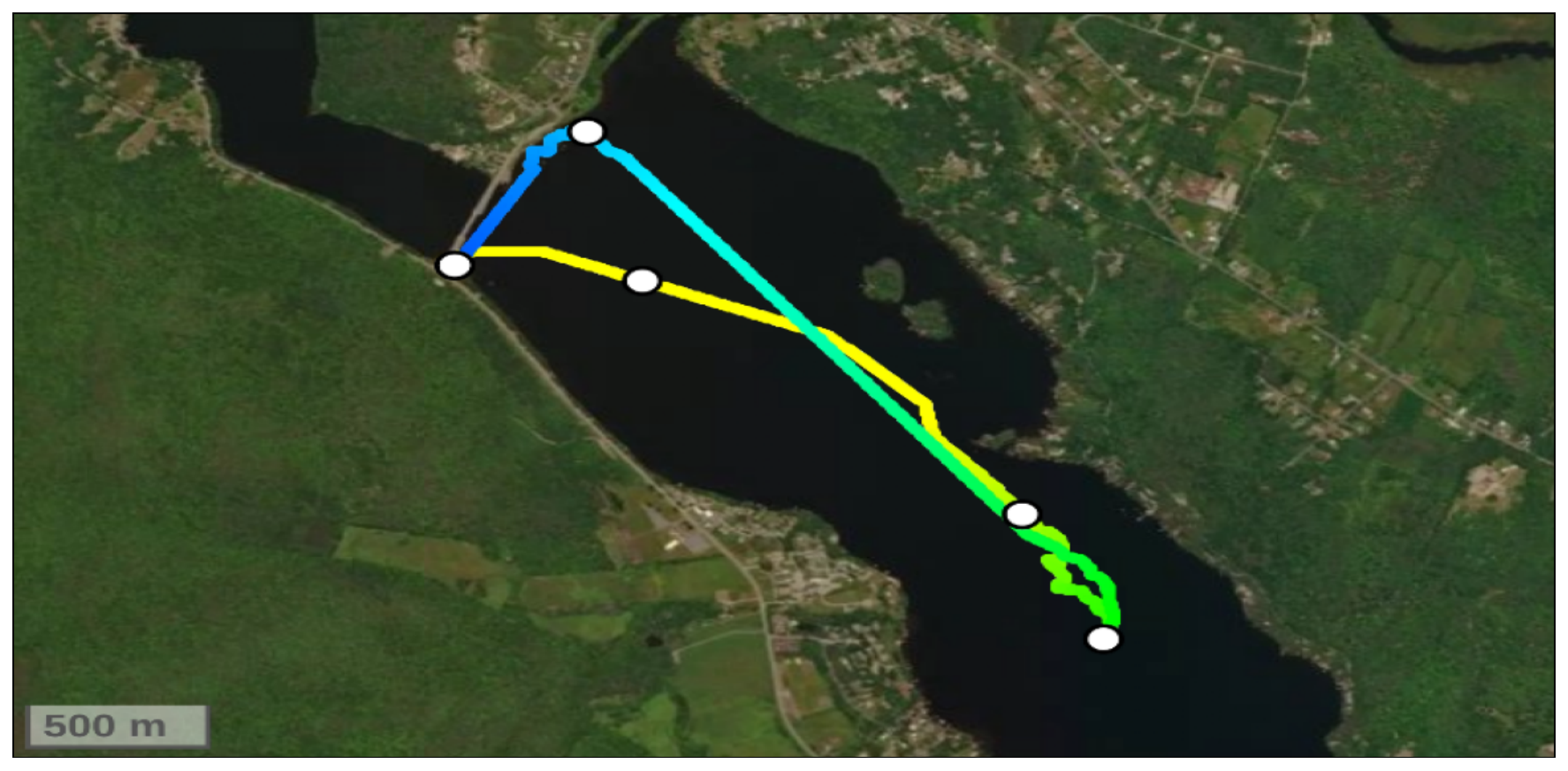}
        }
    \end{minipage}
    \hfill
    \begin{minipage}{0.535\columnwidth}
        \subfloat[Lake -- Sunapee 1]{
            \includegraphics[width=\textwidth]{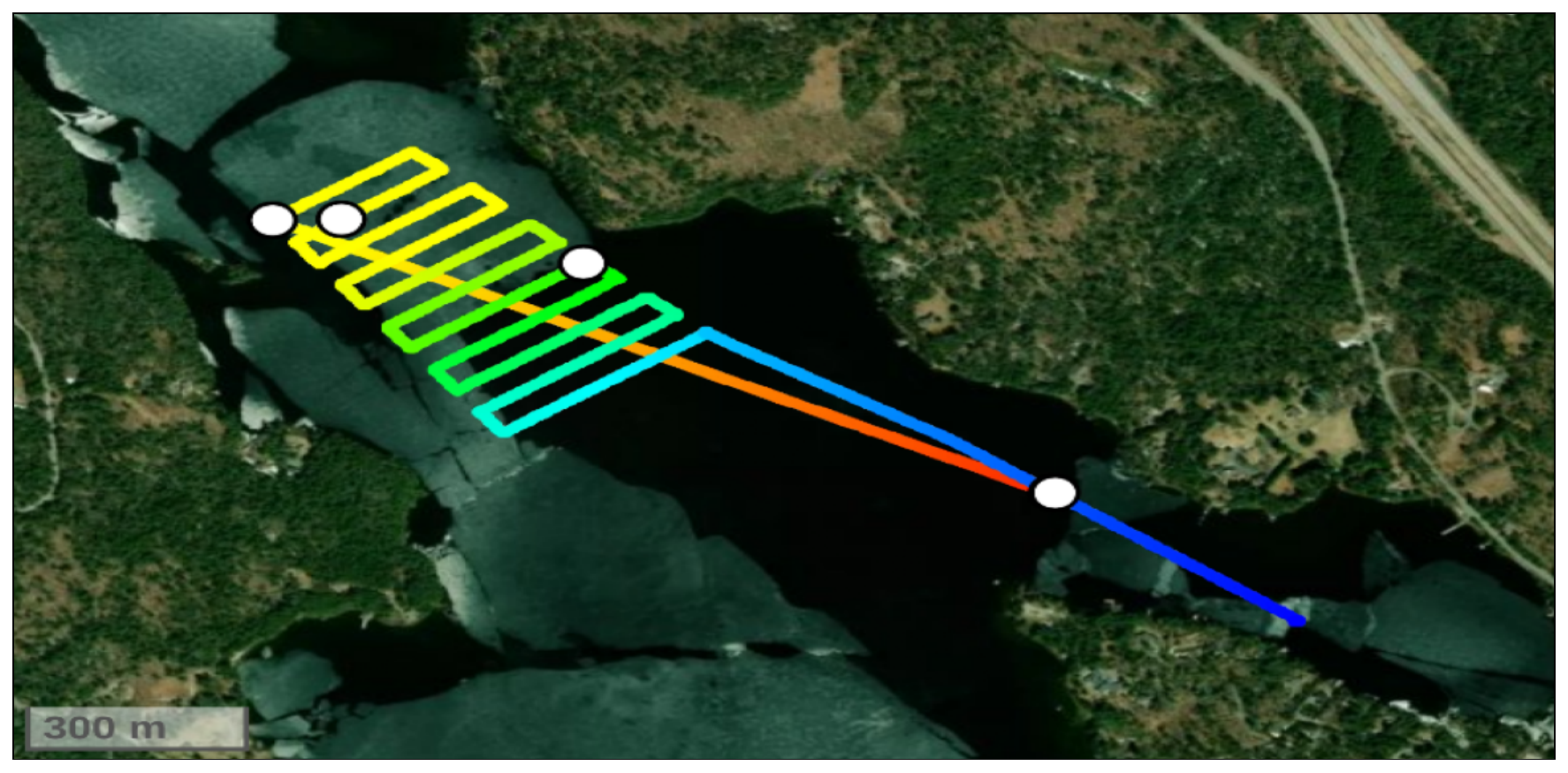}
        }
    \end{minipage}
    \hfill
    \begin{minipage}{0.49\columnwidth}
        \subfloat[Sea -- Busan]{
            \includegraphics[width=\textwidth]{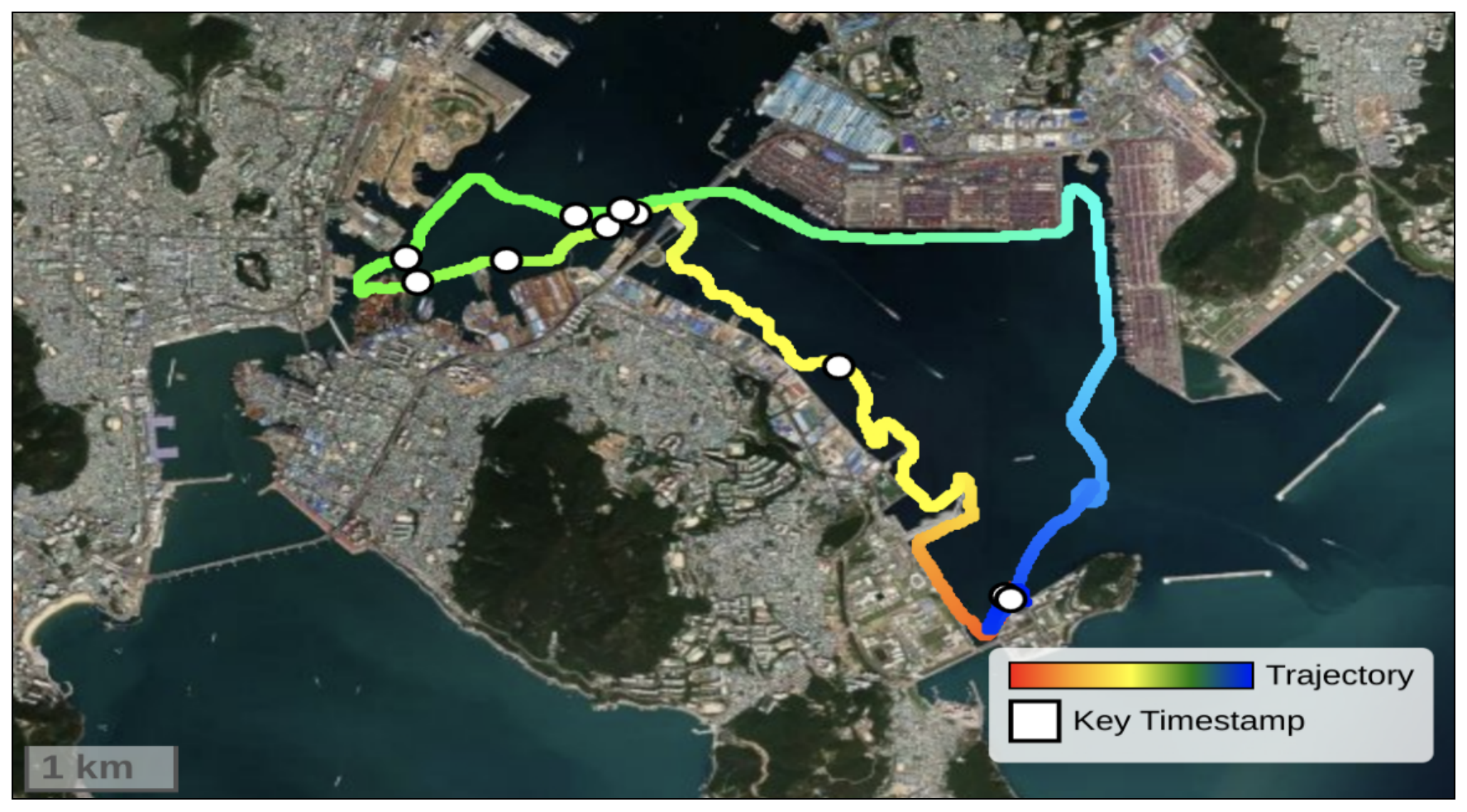}
        }
    \end{minipage}
    \hfill
    \begin{minipage}{0.40\columnwidth}
        \subfloat[Sea -- Barbados 2]{
            \includegraphics[width=\textwidth]{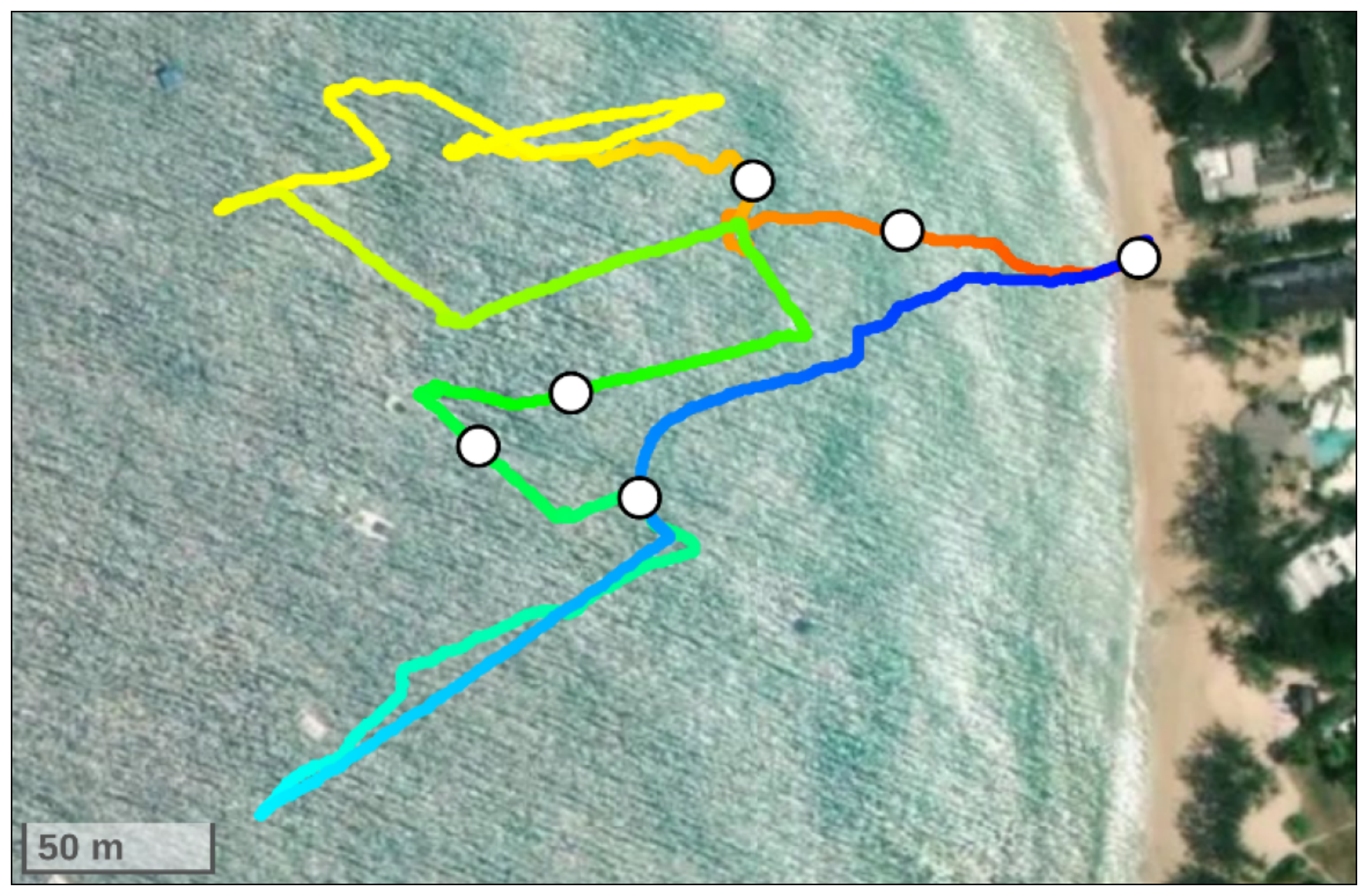}
        }
    \end{minipage}
    \hfill
    \begin{minipage}{0.535\columnwidth}
        \subfloat[Lake -- Mascoma 2]{
            \includegraphics[width=\textwidth]{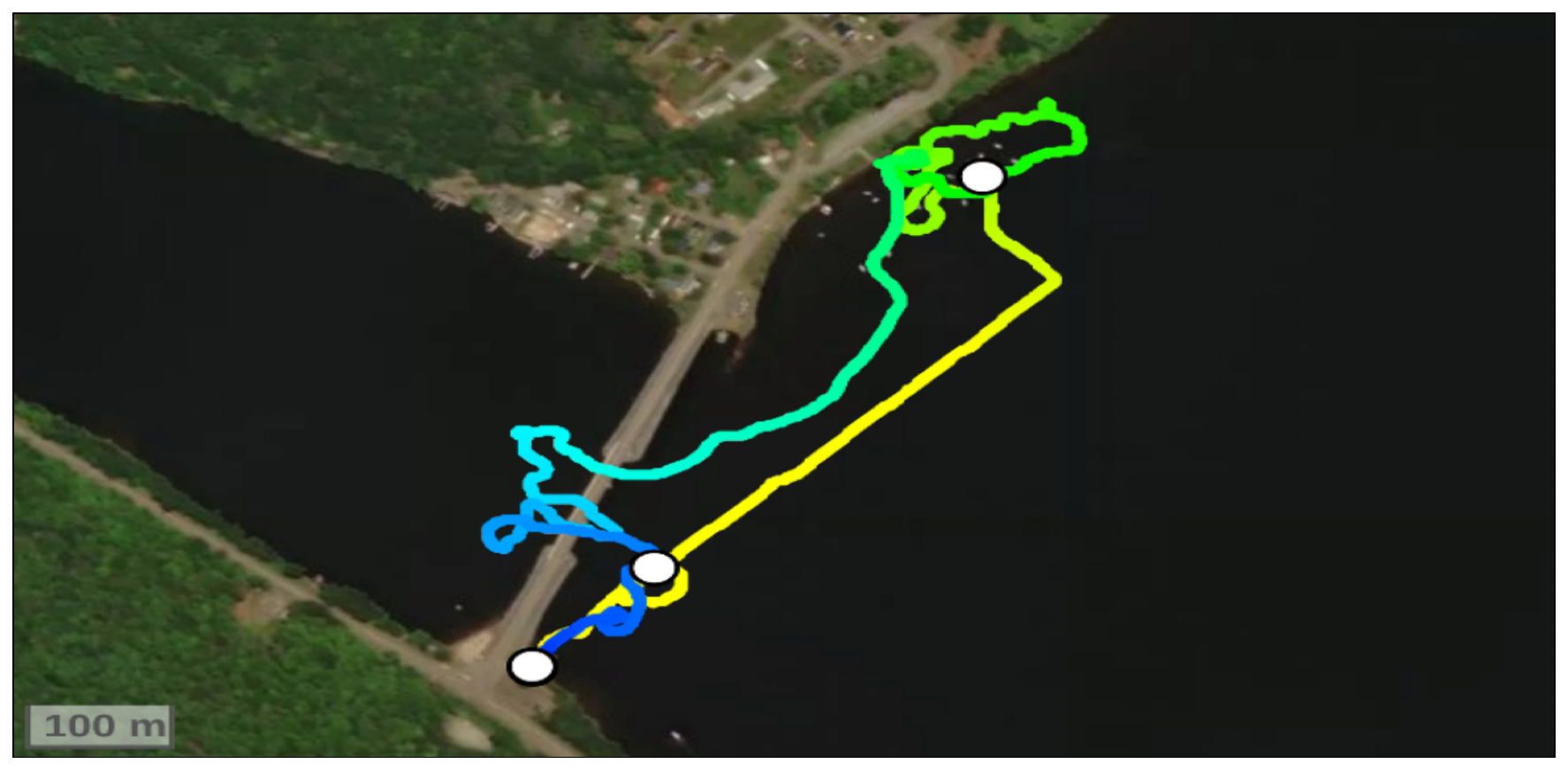}
        }
    \end{minipage}
    \hfill
    \begin{minipage}{0.535\columnwidth}
        \subfloat[Lake -- Sunapee 2]{
            \includegraphics[width=\textwidth]{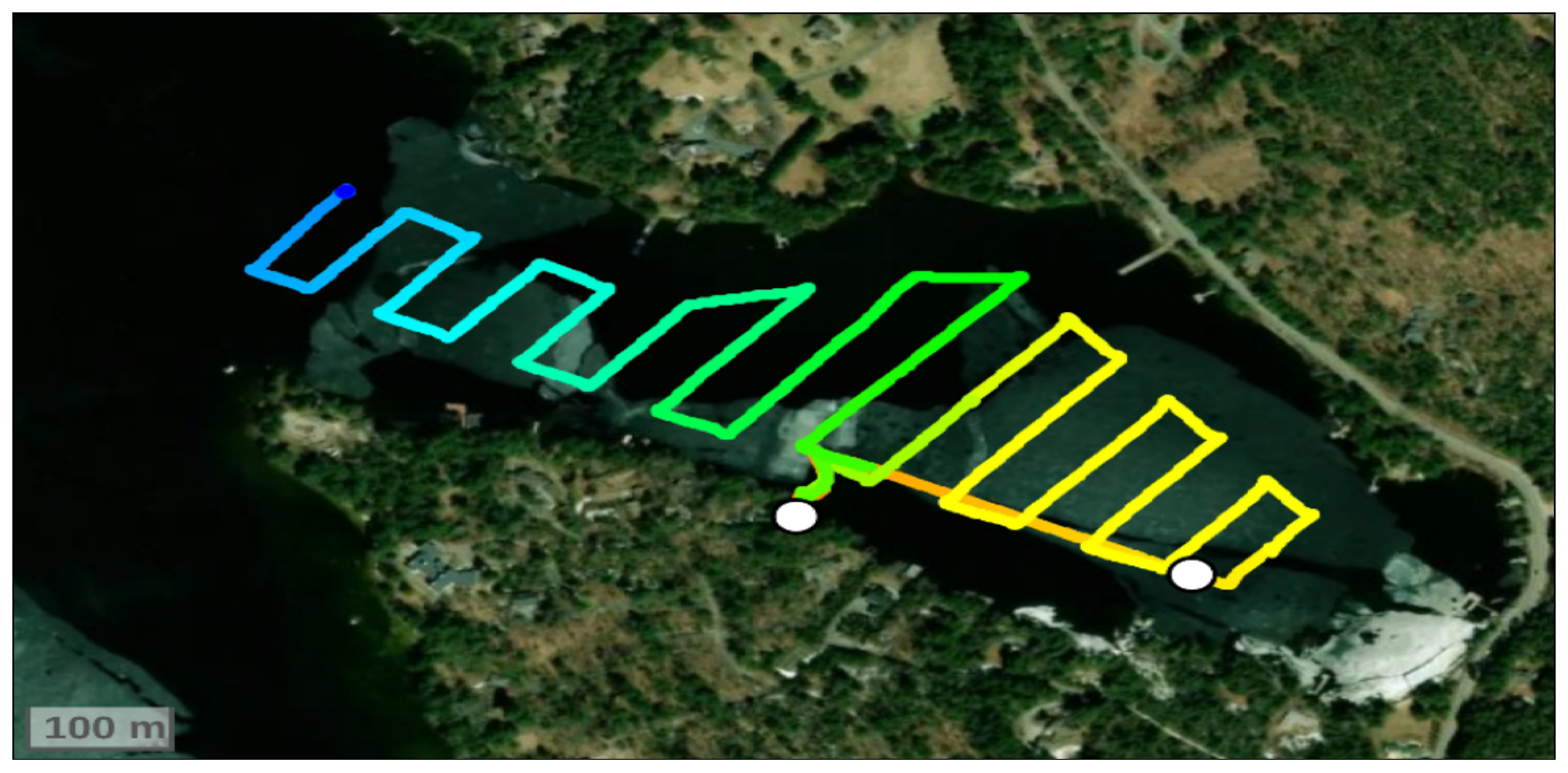}
        }
    \end{minipage}
    \caption{Data collection trajectories in different locations, navigating from \textit{red} to \textit{blue}. \textit{white} points are key frames with objects encountered and corresponding annotations in the dataset, defined as `sequences'.}
    \label{fig:asv-trajectory}
\end{figure*}

We collected relevant data in $\{ \textit{sea}, \textit{fresh} \}$ waters with varying environmental conditions $\{\textit{dusk},  \textit{day},  \textit{night}\}$. We controlled the ASV   \arirevised{via} either (1) autonomous waypoint following or (2) manual driving, while we manually navigated the human-driven boat. \fig{fig:asv-trajectory} shows the trajectories during data collection. Our dataset covers \arirevised{collections conducted between}  $2021$ to $2024$ in different geographic locations: Lake Sunapee, NH, USA; Lake Mascoma, NH, USA; Busan Port, South Korea; Holetown, Barbados.

We post-process the camera and LiDAR data by extracting raw image\arirevised{s} and point cloud\arirevised{s}   under time synchronization \arirevised{using the}  \texttt{MessageFilter} package \cite{message-filter-ros}.

\subsection{Groundtruth Generation} \label{sec:annotation}

We provide annotations of three in-water object classes based on the domain knowledge and navigation-oriented categorization: \textbf{ship}, \textbf{buoy}, and \textbf{other}, within the camera's field of view as well as the LiDAR's field of view (FoV). More specifically, (1) the \textbf{ship class} represents all marine vehicles defined according to the international traffic rule \cite{colreg} as ``every description of watercraft used or capable of being used as a means of transportation on water'', including examples such as power-driven vessels, fishing boats, kayaks, yachts, sailboats; (2) the \textbf{buoy class} represents floating objects as defined by the International Maritime Buoyage System \cite{Admiralty_2018} and includes any artificial objects serving as ``aids to navigation'', like cardinal, lateral, safe water, isolated danger, and special buoys with varying colors and shapes, such as ball and pillar types; and (3) the \textbf{other class} represents any in-water objects that can be risky to maritime navigation, for example, floating docks, fishing nets. We provide ontology documentation for labeling \arirevised{annotation consistency and dataset usage}  .

For the images, we used the 3rd party Amazon AWS \arirevised{Mechanical Turk} annotation service in addition to the annotation by   team members \arirevised{using the open-source Anylabeling \cite{anylabeling} tool and model-assisted labeling using Meta Research's Segment Anything Model (SAM) \cite{kirillov2023segment}}. %
For the LiDAR point clouds, we adapted an open-source labeling tool \cite{3d-bat-2019} for our purpose. We first conducted manual annotations and then resized them to bounding boxes that tightly contains the point cloud within it, while maintaining the yaw of the manually annotated bounding boxes. 
For both, we ran three rounds of annotation review by the expert team members for quality control.

\begin{figure}[t!]
    \centering
    \includegraphics[width=.9\columnwidth]{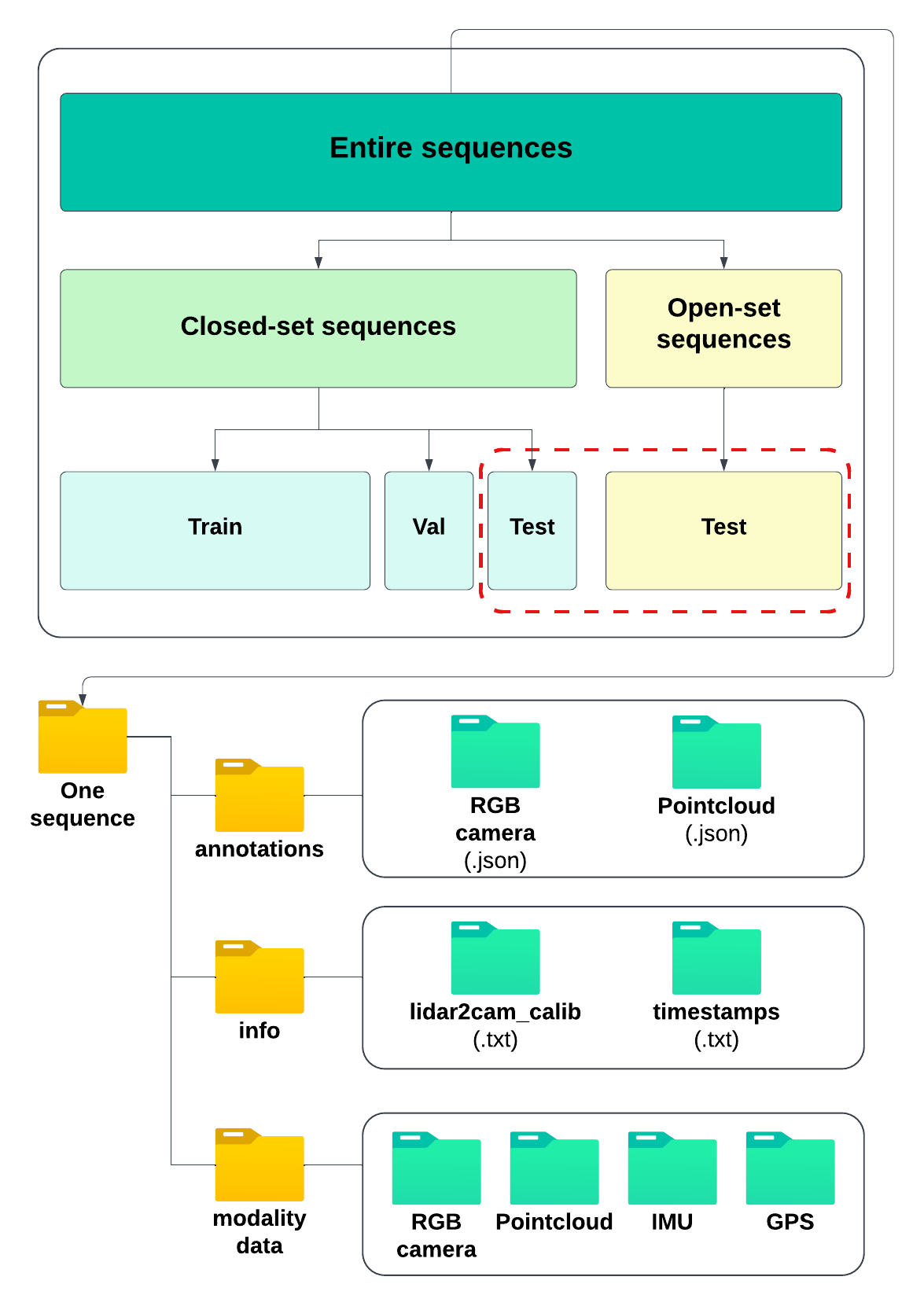}
    \caption{Overall dataset structure.}
    \label{fig:dataset-structure}
\end{figure}

\begin{figure*}[t!]
    \centering
    \begin{minipage}[t]{0.49\columnwidth}
        \subfloat{
           \includegraphics[width=\linewidth]{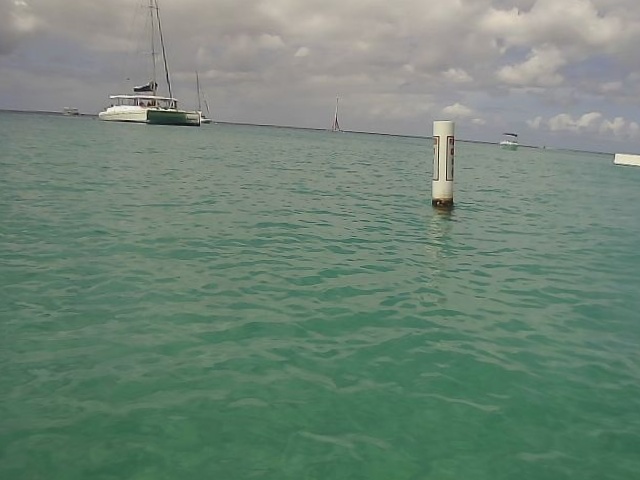}
        }
    \end{minipage}
    \hfill %
    \begin{minipage}[t]{0.49\columnwidth}
        \subfloat{
           \includegraphics[width=\linewidth]{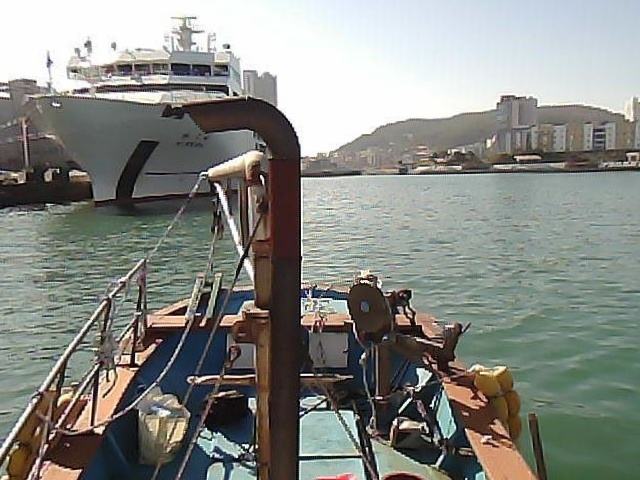}
        }
    \end{minipage}
    \hfill %
    \begin{minipage}[t]{0.49\columnwidth}
        \subfloat{
           \includegraphics[width=\linewidth]{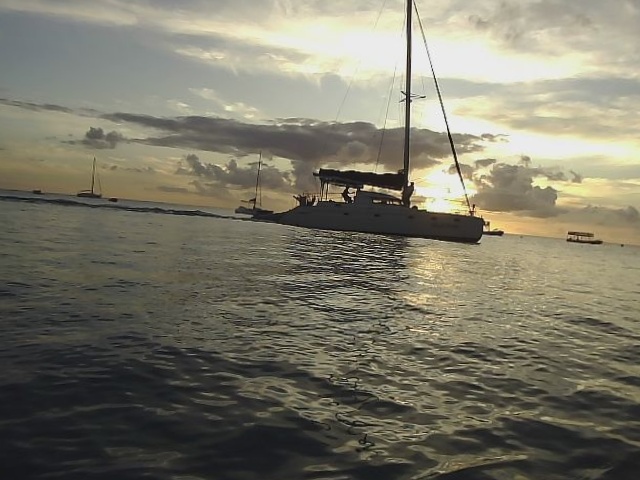}
        }
    \end{minipage}
    \hfill %
    \begin{minipage}[t]{0.49\columnwidth}
        \subfloat{
           \includegraphics[width=\linewidth]{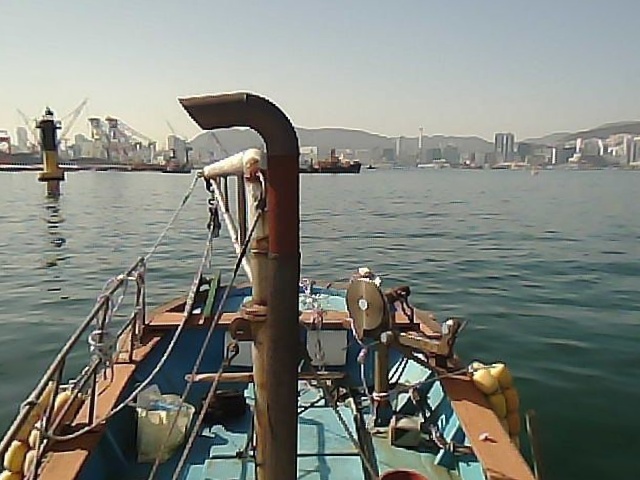}
        }
    \end{minipage}
    \hfill %
    \begin{minipage}[t]{0.49\columnwidth}
        \subfloat[Class Buoy -- Pillar]{
           \includegraphics[height=1.2in]{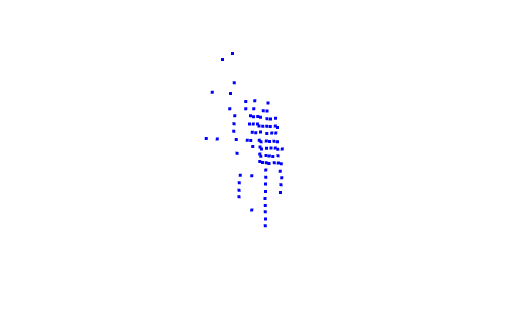}
        }
    \end{minipage}
    \hfill %
    \begin{minipage}[t]{0.49\columnwidth}
        \subfloat[Class Ship -- Large vessel]{
           \includegraphics[height=1.2in]{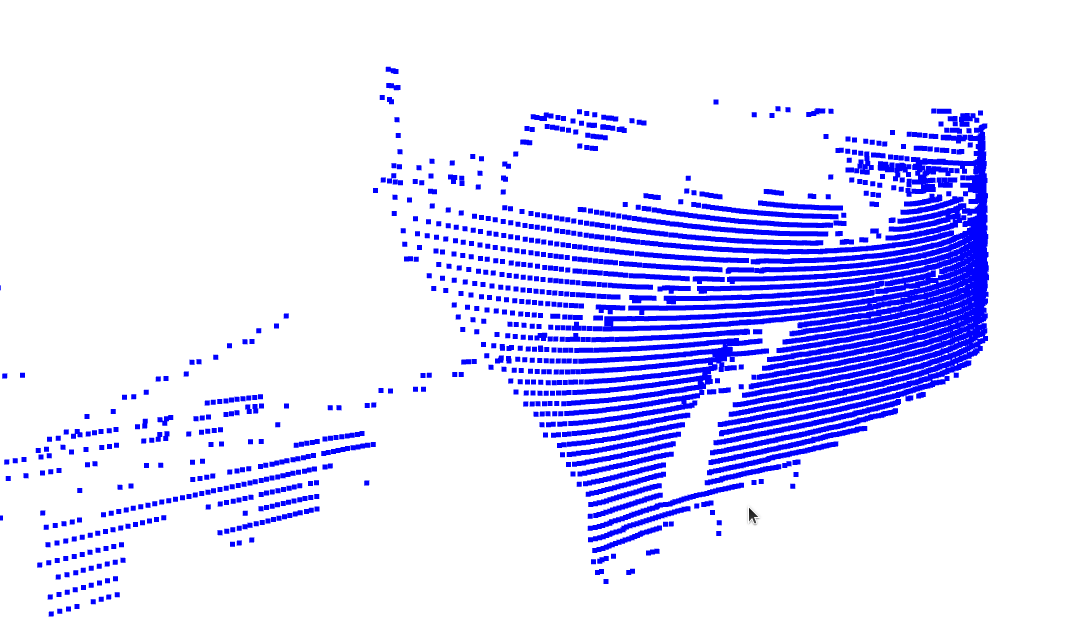}
        }
    \end{minipage}
    \hfill %
    \begin{minipage}[t]{0.49\columnwidth}
        \subfloat[Class Ship -- Yacht]{
           \includegraphics[height=1.2in]{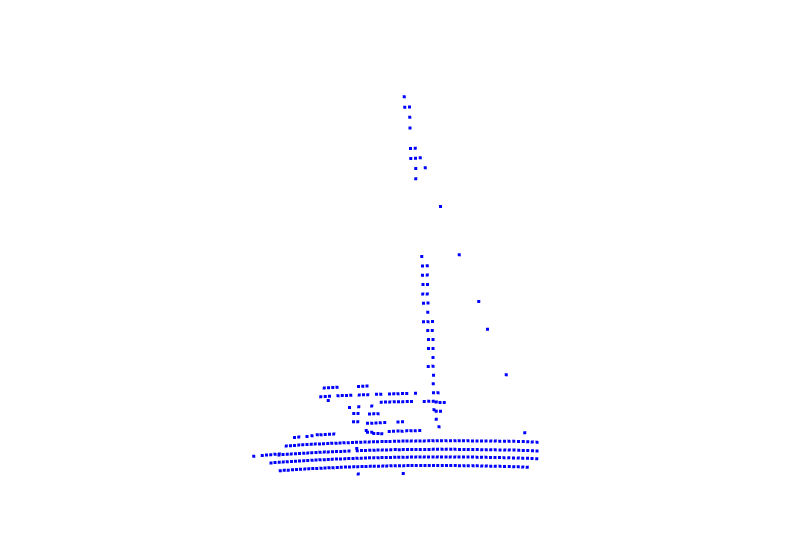}
        }
    \end{minipage}
    \hfill %
    \begin{minipage}[t]{0.49\columnwidth}
        \subfloat[Class Buoy -- Cardinal]{
           \includegraphics[height=1.2in]{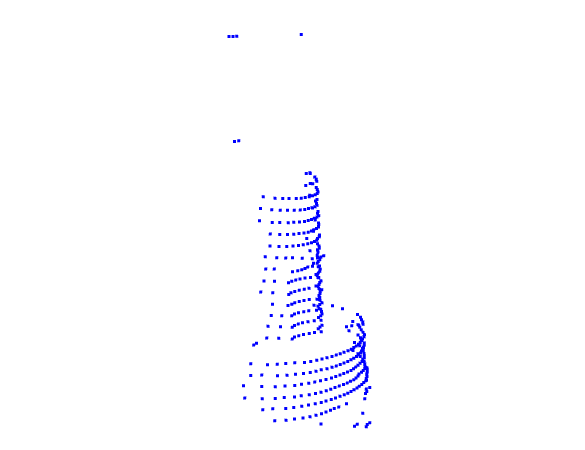}
        }
    \end{minipage}
    \hfill %
    \begin{minipage}[t]{0.49\columnwidth}
        \subfloat{
           \includegraphics[width=\linewidth]{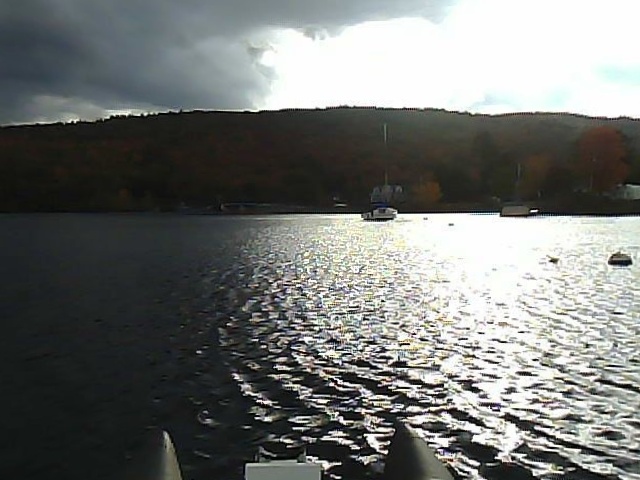}
        }
    \end{minipage}
    \hfill %
    \begin{minipage}[t]{0.49\columnwidth}
        \subfloat{
           \includegraphics[width=\linewidth]{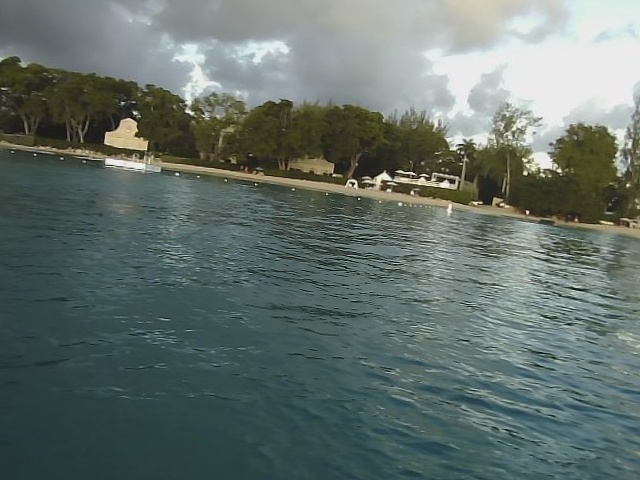}
        }
    \end{minipage}
    \hfill %
    \begin{minipage}[t]{0.49\columnwidth}
        \subfloat{
           \includegraphics[width=\linewidth]{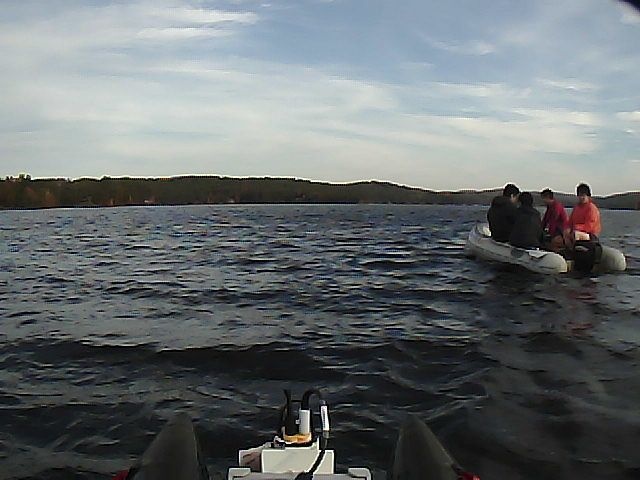}
        }
    \end{minipage}
    \hfill %
    \begin{minipage}[t]{0.49\columnwidth}
        \subfloat{
           \includegraphics[width=\linewidth]{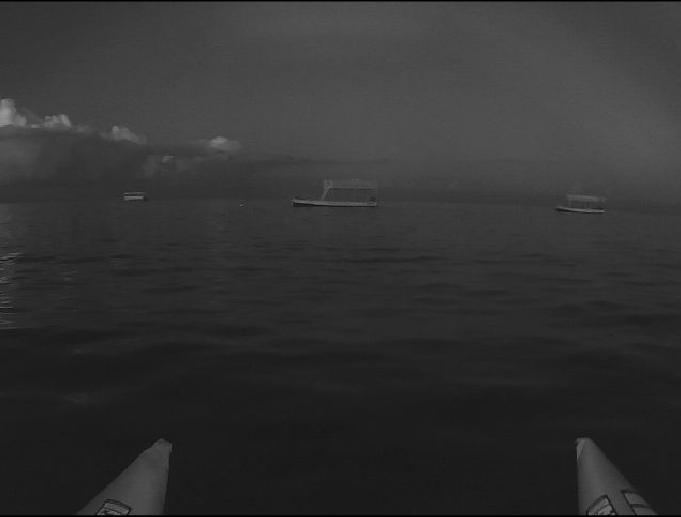}
        }
    \end{minipage}
    \hfill %
    \begin{minipage}[t]{0.49\columnwidth}
        \subfloat[Class Buoy -- Ball]{
           \includegraphics[height=1.2in]{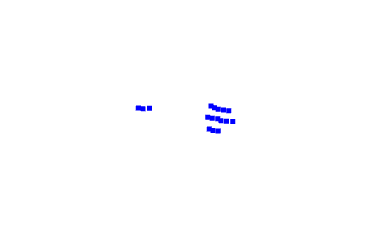}
        }
    \end{minipage}
    \hfill %
    \begin{minipage}[t]{0.49\columnwidth}
        \subfloat[Class Other -- floating dock]{
           \includegraphics[height=1.2in]{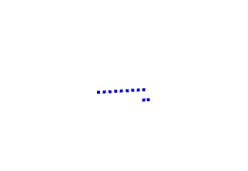}
        }
    \end{minipage}
    \hfill %
    \begin{minipage}[t]{0.49\columnwidth}
        \subfloat[Class Ship -- raft with people]{
           \includegraphics[height=1.2in]{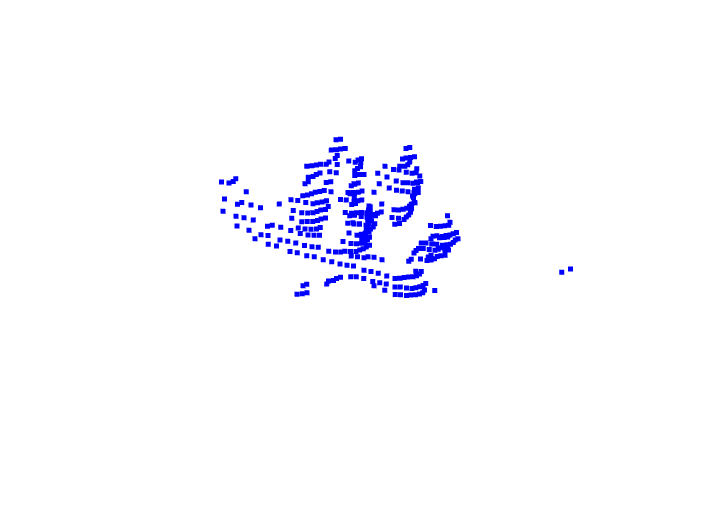}
        }
    \end{minipage}
    \hfill %
    \begin{minipage}[t]{0.49\columnwidth}
        \subfloat[Class Ship -- boat by Infrared]{
           \includegraphics[height=1.2in]{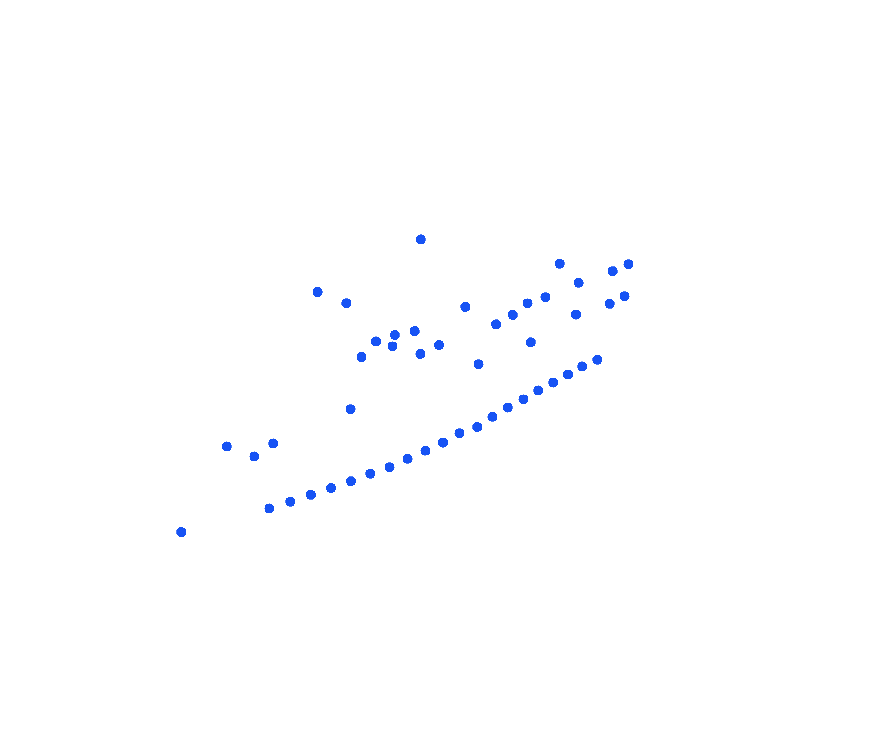}
        }
    \end{minipage}
    \caption{In-water objects under varying environmental conditions in our dataset. (\textit{top}): images; (\textit{bottom}): point clouds. Note that the view angle of the point clouds is adjusted for the best visualization, regardless of the corresponding image of the object.}
    \label{fig:collected-data}
\end{figure*} 

We provide the label format in a standardized way along with converter implementations, such as YOLO format, KITTI format, unified normative, so that users can apply the dataset to different applications. The point cloud label contains $\{x, y, z, dx, dy, dz, \textit{yaw}, \textit{class}\}$ information. We only provide the yaw angle, assuming the roll and pitch remain approximately zero. Even if in rough water conditions this assumption might not hold, roll and pitch information is typically not necessary for ASV 2D navigation. For consistency of labeling in one frame of an image and a point cloud with its quality, we used a custom tool to extract the same object across the modalities (\fig{fig:tool-overlay}). For the KITTI label format, we consider the annotation of an object as valid, only if they are located within FoV of both camera and LiDAR, following KITTI benchmark guideline~\cite{KITTI-raw-2013}.

\begin{figure}
    \centering
    \includegraphics[width=\columnwidth]{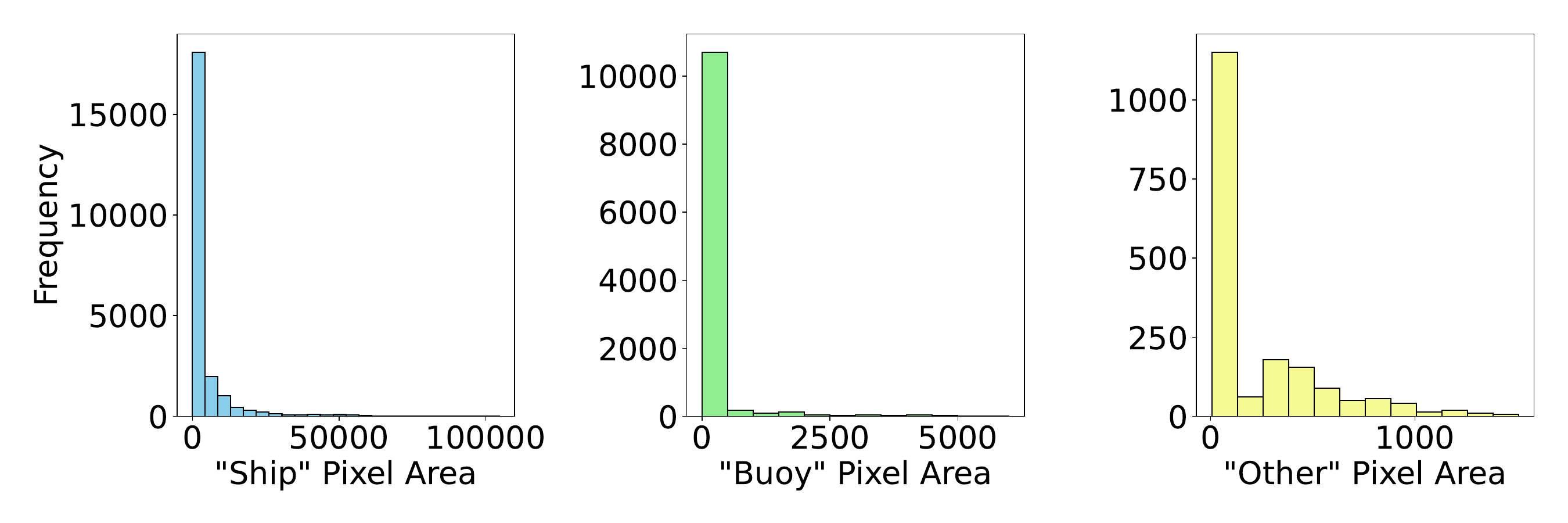}
    \includegraphics[width=\columnwidth]{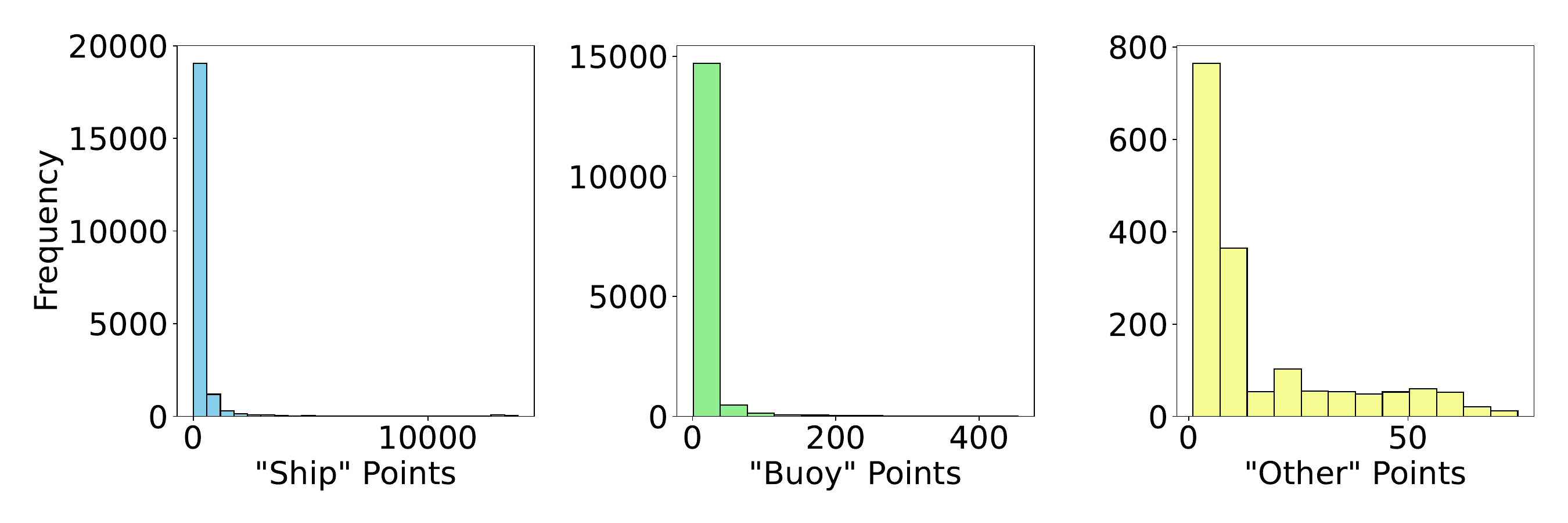}
    \caption{The distribution of labeled object pixel area in the RGB Camera modality (top) and of LiDAR points (bottom) by class. Both pixel area and number of LiDAR points exhibit similar distributions.}
    \label{fig:pixel-area-point-count}
\end{figure}

\subsection{Dataset Structure} \label{sec:dataset-structure}
\fig{fig:dataset-structure} shows the overall structure of our dataset, divided into three subsets: \emph{train}, \emph{validation}, and \emph{test}. We define a sequence as a 60-second event involving object encounters at a specific geographical location, including Barbados, Busan, Lake Sunapee, and Lake Mascoma. For each sequence, we establish subdirectories based on annotations, information, and sensor modalities. \revised{In addition, we categorize sequences into \emph{closed-set} (used for training and evaluation) and \emph{open-set} (excluded from training and used only for evaluation).}

Given the geographical coverage of the dataset---$\{\textit{\textbf{sea}: Barbados, Busan},\, \textit{\textbf{fresh}: Sunapee, Mascoma} \}$---we first construct the \emph{open-set} by selecting one sequence from each location, resulting in a total of four sequences and \opensetSize ~frames total. Each selected sequence was chosen to reflect a challenging condition specific to its environment: Sunapee features multiple kayaks at far distances; Mascoma includes many boats and buoys (more than 10) under water surface glare; Barbados captures a sunset scenario; and Busan contains both a buoy and a boat approaching from a distance to close proximity. \secrevised{For the remaining \closesetSize~frames, which form the \emph{closed-set}, at the sequence level, we randomly shuffle and split into \emph{train}, \emph{validation}, and \emph{test} subsets using a $0.70, 0.15, 0.15$ ratio.} This partitioning strategy enables fair quantifiable evaluation of both model performance and generalization capabilities for learning-based algorithms~\cite{eccv-open-set-2022}.

\section{Dataset Characteristics} \label{sec:dataset-characteristic}

\subsection{Dataset Composition} \label{sec:dataset-composition}
As shown in \fig{fig:collected-data}, our maritime perception dataset consists of various objects in water under varying conditions collected by the sensor platforms onboard ASVs or onboard a human-driven ship. This annotated, ego-perspective dataset is the first in the maritime domain, to the best of our knowledge  \arirevised{with} sufficiently large number of annotated frames (total \datasetSize).    \arirevised{W}e believe it   \arirevised{will} be useful for training, validat  \arirevised{ing}, and   \arirevised{benchmarking} maritime perception.

\begin{table}[t!]
    \caption{Labeled objects by class present in the dataset in the RGB image modality and the LiDAR modality.}
    \centering
    \begin{tabular}{|l|l|l|l|}
    \hline
    \textbf{Class Name} & Ship  & Buoy  & Other \\ \hline
    \textbf{Image Obj. Count}      & 22874 & 11337 & 1833  \\ \hline
    \textbf{LiDAR Obj. Count}      & 22251 & 15692 & 1636  \\ \hline
    \end{tabular}
    \label{tab:object_counts}
\end{table}

\tab{tab:object_counts} shows the annotated class breakdown in the RGB camera data and LiDAR data, where the predominant class in both modalities is ``ship,'' followed by ``buoy,'' and then ``other.''   \revised{While both modalities of the dataset exhibit a class imbalance between the ``ship'' annotations and the other two classes, this imbalance naturally reflects the characteristics of coastal navigation environments represented in the dataset. Training performant learning-based models using this dataset may require strategies to address this natural imbalance---see \sect{sec:discussion}.} We characterize the annotation resolution, made via 2D and 3D bounding boxes, based on its pixel area (\fig{fig:pixel-area-point-count}(top)) and the number of LiDAR points (\fig{fig:pixel-area-point-count}(bottom)) respectively. This resolution is inherently limited by the underlying sensor resolution as well as other confounders related to the modality (e.g., illumination for RGB cameras) and others related to the maritime domain (e.g., in-water dynamics). Still, this approach gives insight into the amount of available sensor information upon which to detect and classify objects. 

For the majority of objects, the annotation resolution is in the lowest bin, where the ship class has the highest average pixel area (mean: $4197.1$, standard deviation: $10194.2$, median: $794.0$), followed by other (mean: $157.4$, standard deviation: $551.7$, median: $28.0$), and, finally, buoy (mean: $218.3$, standard deviation: $301.2$, median: $38.0$). Generally, the point cloud data follows the same trend where ships have the highest average point-cloud points (mean: $360.1$, standard deviation: $1477.2$, median: $37.0$), followed by other (mean: $15.8$, standard deviation: $17.8$, median: $8.0$), and, then, buoy (mean: $11.0$, standard deviation: $35.6$, median: $2.0$). Of note is the long-tailed nature of the distributions in \fig{fig:pixel-area-point-count}, meaning that there is a large amount of heterogeneity within the same class.  

In terms of environmental conditions, the data is composed of 79.9\% for ``day'', 14.9\% for ``dusk'', and 5.2\% for ``night''. 
\revised{Dusk and night are imbalanced given the challenges in collecting data during that time. While we envision future work expanding the dataset to include a broader range of lighting conditions, we provide suggestions in \sect{sec:discussion} to address this challenge together with the class imbalance challenge.}

\subsection{Dataset Complexity}
As described in detail below, we propose novel metrics (e.g., BEVE-P, BEVE-V, DVE) in addition to existing metrics (e.g., image entropy, occlusion percentage) in the literature that quantitatively evaluate the \arirevised{dataset's} characteristic\arirevised{s}  \arirevised{with respect to} the maritime domain \monrevised{to help analyze future} benchmark algorithms.

\begin{figure}[t!]
    \centering
    \includegraphics[width=0.8\columnwidth,trim=3cm 0 0 1cm]{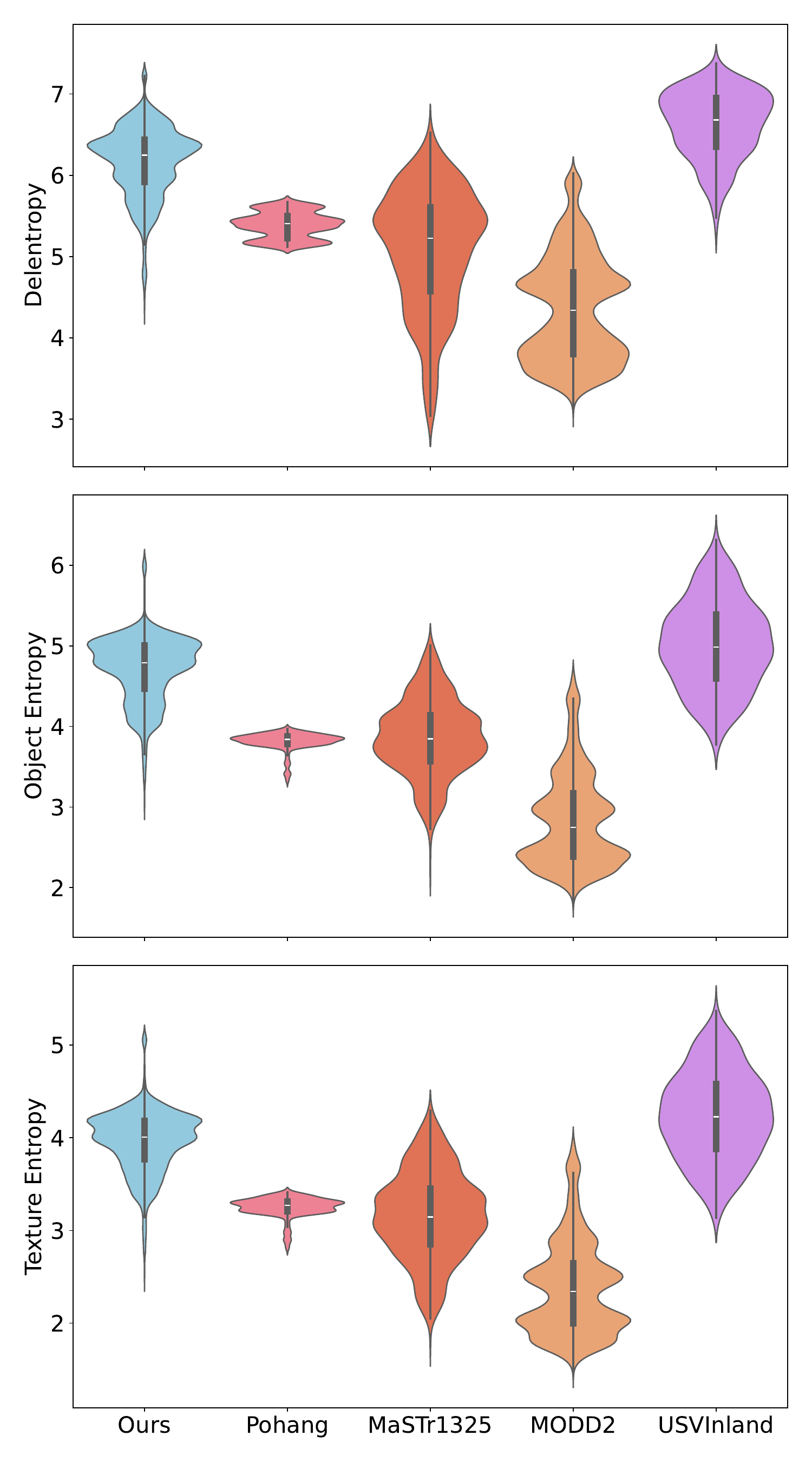}
    \caption{\secrevised{RGB image complexity comparisons between our dataset (\textit{light blue}) and four other maritime perception image datasets Pohang \cite{Pohang-2023}, MaSTr1325 \cite{MaSTr1325-2019}, MODD2 \cite{MODD2-2018}, and USVInland~\cite{Cheng2021AreWR}.}}
    \label{fig:image-delentropy}
\end{figure}

\subsubsection{Image Complexity}
Image entropy indicates the variation or complexity of an image at the grayscale distribution. In general, a low value corresponds to less edges and corners and possibly fewer interesting features, while a high value corresponds to an image with significant amount of texture.

We evaluate image complexity with three entropy metrics: delentropy, object-level entropy, and texture-level entropy. For \textbf{delentropy} metric, we first applied the Sobel operator to approximate the gradients along the vertical and horizontal directions and afterwards calculated the Shannon entropy. We take inspiration from the evaluation criterion in the work by~\cite{alvarez2023mimir}, an underwater dataset -- where image-based object detection algorithms typically implement some preliminary edge detection processing. Note, instead of the Sobel filter, another edge detection algorithm, such as Canny Edge detector, can work as well. The traditional \textbf{object entropy} and \textbf{texture entropy} metrics are similar in that they are directly calculating the Shannon entropy, but with different sized template discs -- object-level with a disc of 10 pixel radius and feature-level with a disc of 5 pixel radius. Here, there is no prior applied edge-detection based filter.

\fig{fig:image-delentropy} depicts the results of image complexity, according to the above three entropy metrics, for our dataset as well as for four other comparison datasets: Pohang~\cite{Pohang-2023}, MaSTR1325~\cite{MaSTr1325-2019}, MODD2~\cite{MODD2-2018}, and USVInland~\cite{Cheng2021AreWR}. Compared to the Pohang dataset, our dataset includes more diverse imagery scenes. On the other hand, the image complexity of the USVInland dataset is comparable to our dataset -- not surprising, given the various textures of nearby trees, rocks, tunnels, and houses in inland waters. While the MaSTR1325 and MODD2 datasets (both from the same authors) have a greater range of complexity compared to our dataset -- much of their images have small objects (relative to image size) and due to observable off white-balancing, the pixel intensity values are within a smaller range -- leading to many images corresponding to low entropy values. Our dataset shows a wide diversity of images, and with better on-camera white-balancing, our images have greater pixel intensity variations.

\subsubsection{LiDAR Complexity}

\begin{figure}[b!]
    \centering
    \includegraphics[width=\columnwidth]{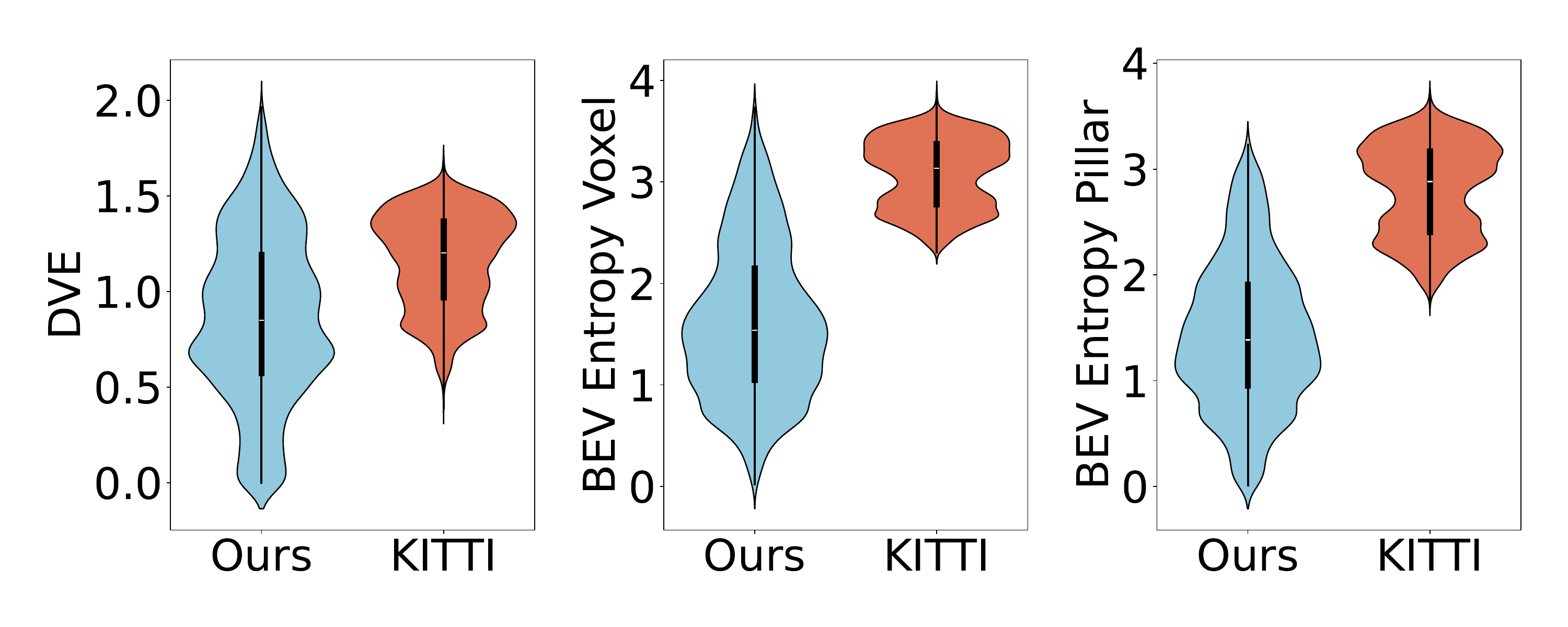}
     \caption{Point cloud complexity comparison between our dataset (\textit{light blue}) and KITTI \cite{KITTI-raw-2013} (\textit{red-orange}). Across DVE, BEV Entropy Voxel, and BEV Entropy Pillar, our dataset shows a greater range of point cloud complexities.   }
    \label{fig:lidar-entropy}
\end{figure}

\revised{
We introduce three entropy-based metrics to evaluate the spatial complexity of LiDAR-derived point clouds: Birds-Eye-View Entropy with Pillars (BEVE-P) and Voxels (BEVE-V), and Distance Variability Entropy (DVE). These metrics gauge how the point distribution spans discretized bins (pillars, voxels, or distance intervals), offering a detailed view of spatial variability in LiDAR data. Lower entropy values indicate that scene objects are more concentrated and clustered within a certain region. Conversely, higher entropy values indicate that there are objects more widely distributed or densely spread across the sensor range. A dataset with variation in entropy values represents its richness and complexity of the data.}

\begin{table*}[t!]
\caption{Performance breakdown of 2 benchmark 2D image object detectors by class. mAP is reported via the aggregated IoU threshold from (0.5 to 0.95) per class.   }
\centering
\begin{tabular}{|l|cc|cc|cc|cc|}
\hline
\multirow{2}{*}{\textbf{Model}} & \multicolumn{2}{c|}{\textbf{Aggregated mAP (0.5:0.95)}} & \multicolumn{2}{c|}{\textbf{``ship'' mAP (0.5:0.95)}} & \multicolumn{2}{c|}{\textbf{``buoy'' mAP (0.5:0.95)}} & \multicolumn{2}{c|}{\textbf{``other'' mAP (0.5:0.95)}} \\ \cline{2-9} & \textbf{\hspace{5mm}Val} & \textbf{Test} & \textbf{\hspace{4mm}Val} & \textbf{Test} & \textbf{\hspace{4mm}Val} & \textbf{Test} & \textbf{\hspace{4mm}Val} & \textbf{Test} \\ \hline\hline
YOLOv9~\cite{YOLOv9-2024} & \textbf{\hspace{5mm}0.54} & \textbf{0.42} & \textbf{\hspace{4mm}0.83} & \textbf{0.65} & \textbf{\hspace{4mm}0.42} & \textbf{0.34} & \textbf{\hspace{4mm}0.36} & \textbf{0.27} \\ \hline
RT-DETR~\cite{zhao2024detrsbeatyolosrealtime} & \hspace{5mm}0.21 & 0.16 & \hspace{4mm}0.45 & 0.36 & \hspace{4mm}0.13 & 0.10 & \hspace{4mm}0.04 & 0.01 \\ \hline

\end{tabular}
\label{tab:image-class-comparison}
\end{table*}

\begin{table*}[h!]
\caption{\revised{Comparison of validation and test results for LiDAR-based benchmarks (IoU thresholds of 0.7 and 0.5) for \textit{ship} class objects. \textit{Green} highlights the best performance across fusion methods, while \textit{yellow} indicates the best performance among LiDAR-only methods.}}
\centering
\begin{tabular}{|l|l|cc|cc!{\vrule width 1.2pt}cc|cc|}
\hline
\multirow{2}{*}{\textbf{Model}} & \multirow{2}{*}{\textbf{Modality}} & \multicolumn{2}{c|}{\textbf{BEV AP (0.7)}} & \multicolumn{2}{c!{\vrule width 1.2pt}}{\textbf{BEV AP (0.5)}} & \multicolumn{2}{c|}{\textbf{3D AP (0.7)}} & \multicolumn{2}{c|}{\textbf{3D AP (0.5)}} \\ \cline{3-10} 
                                &                                   & \textbf{Val}         & \textbf{Test}         & \textbf{Val}         & \textbf{Test}         & \textbf{Val}         & \textbf{Test}         & \textbf{Val}         & \textbf{Test}         \\ \hline\hline
PointPillars~\cite{Lang2018PointPillarsFE} & LiDAR-only                    & 28.01                & 17.32                & 57.22                & 50.77                & 4.23                 & 3.14                 & 30.30                & 30.07                \\ \hline
SECOND~\cite{second-2018}                 & LiDAR-only                    & 34.29 & 27.68 & 56.95 & 52.36 & 8.93                 & 10.17                & 40.70                & 40.67                \\ \hline
PointRCNN~\cite{point-rcnn-2019}          & LiDAR-only                    & 3.24                 & 3.11                 & 23.93                & 21.48                & 0.32                 & 0.43                 & 2.91                 & 2.66                 \\ \hline
PV-RCNN~\cite{pv-rcnn-2020}               & LiDAR-only                    & 19.11                & 9.99                 & 42.40                & 38.65                & 3.79                 & 9.09                 & 23.64                & 16.54                \\ \hline
Voxel-RCNN~\cite{deng2020voxel-rcnn}      & LiDAR-only                    & 33.36                & 27.50                & 54.55                & 50.69                & 12.90 & 13.46 & 41.96 & 43.03 \\ \hline
TED-S~\cite{TED-2023}                  & LiDAR-only                    & \cellcolor{yellow!20}\textbf{49.64} & \cellcolor{yellow!20}\textbf{37.36} & \cellcolor{yellow!20}\textbf{70.56} & \cellcolor{yellow!20}\textbf{55.09} & \cellcolor{yellow!20}\textbf{36.88} & \cellcolor{yellow!20}\textbf{26.99} & \cellcolor{yellow!20}\textbf{60.82} & \cellcolor{yellow!20}\textbf{46.10} \\ \hline\hline
PointPainting~\cite{pointpainting-2020}   & Fusion                        & 30.51                & 25.42                & 57.92                & 46.10                & 10.05                & 12.95                & 42.54                & 37.67                \\ \hline
CLOCs~\cite{CLOCs-2020}                   & Fusion                        & 32.02                & 21.76                & 56.68                & 49.47                & 9.69                 & 8.28                 & 45.38                & 41.10                \\ \hline
Focal Conv-F~\cite{focal-sparse-2022}     & Fusion                        & 37.48                & \cellcolor{green!20}\textbf{36.36} & \cellcolor{green!20}\textbf{61.69} & \cellcolor{green!20}\textbf{54.55} & 19.83                & 15.58                & 47.79                & \cellcolor{green!20}\textbf{45.45} \\ \hline
TED-M~\cite{TED-2023}                & Fusion                        & \cellcolor{green!20}\textbf{50.32} & 32.40 & 54.05 & 43.64 & \cellcolor{green!20}\textbf{30.24} & \cellcolor{green!20}\textbf{27.69} & \cellcolor{green!20}\textbf{53.87} & 42.91 \\ \hline
\end{tabular}
\label{tab:lidar-comparison}
\end{table*}

\textbf{Birds-Eye-View Entropy} \revised{with Pillar/Voxel (BEVE-P, BEVE-V): These metrics measure \textit{point cloud complexity} based on a discretized representation (pillar or voxel bins) of the LiDAR data. Formally, we define the metric as:
\begin{equation}
\textrm{BEV} = - \sum_{i=1}^{M} \left[\frac{k_i}{K} \log \left(\frac{k_i}{K}\right)\right],
\end{equation}
where $i$ indexes each pillar or voxel, $M$ is the total number of pillars or voxels, $K$ represents the total number of points in the frame, and $k_i$ is the count of points in each respective pillar or voxel. A lower BEV indicates that there is a concentration of objects in fewer bins; while a higher BEV suggests that there is a broader distribution of objects across multiple bins, reflecting a richer spatial arrangement.}

\textbf{Distance Variability Entropy (DVE):} \revised{This metric evaluates \textit{point cloud complexity} based on radial distance from the LiDAR sensor, measuring how points are distributed across predefined radial distance intervals. We define the metric as:
\begin{equation}
\textrm{DVE} = -\sum_{i=1}^{R} \left[\frac{n_i}{N} \log \left(\frac{n_i}{N}\right)\right],
\end{equation}
where $i$ indexes each predefined radial distance interval, $R$ is the total number of distance intervals, $N$ is the total number of points within the frame, and $n_i$ is the number of points in each respective radial distance ring. A lower DVE suggests that points lie within fewer radial bands (indicating a simpler or more concentrated layout), whereas a higher DVE indicates that points are spread more extensively across different distances (denoting a more complex, broad-ranging scene)}.

\fig{fig:lidar-entropy} shows the proposed complexity metrics of the dataset within the collected point clouds, indicating that we have varying spatial distributions of in-water objects.

\section{Perception Benchmarks}\label{sec:benchmarks}

We ran the perception benchmark on our proposed datasets on detection tasks, i.e.,  
\textbf{object detection} and \textbf{object classification} and developed the necessary conversion tools. 
We used a computer equipped with an Intel i7-7820X 8-core \SI{3.6}{GHz} processor, \SI{32}{GB} RAM, and NVIDIA GPU RTX 3090 Ti with \SI{24}{GB} VRAM. 
We evaluated benchmark algorithms, offering insights into the applicability \arirevised{and adaptability} of these benchmarks  in the maritime domain. 

\subsection{Image-based benchmarks} 

While many real-time RGB image object detection approaches exist, we selected two representative models to provide a benchmarking of this dataset upon: YOLOv9 \cite{YOLOv9-2024} and RT-DETR \cite{zhao2024detrsbeatyolosrealtime}. We specifically benchmark using real-time detectors as the ego-centric ASV perspective of this dataset lends itself to use in real-time, on-board object detection use cases. Based on that criteria, we selected a model from the popular You Only Look Once (YOLO)  object detector lineage, which uses a Convolutional Neural Network (CNN) backbone approach and a newer Transformer-backbone approach based on the Detection Transformer \cite{detr_10.1007/978-3-030-58452-8_13} (DETR), that was adapted for real-time (RT) use. 

\begin{figure*}
        \subfloat{
		\includegraphics[width=0.32\textwidth,trim={3cm 5.5cm 3cm 5.5cm},clip]{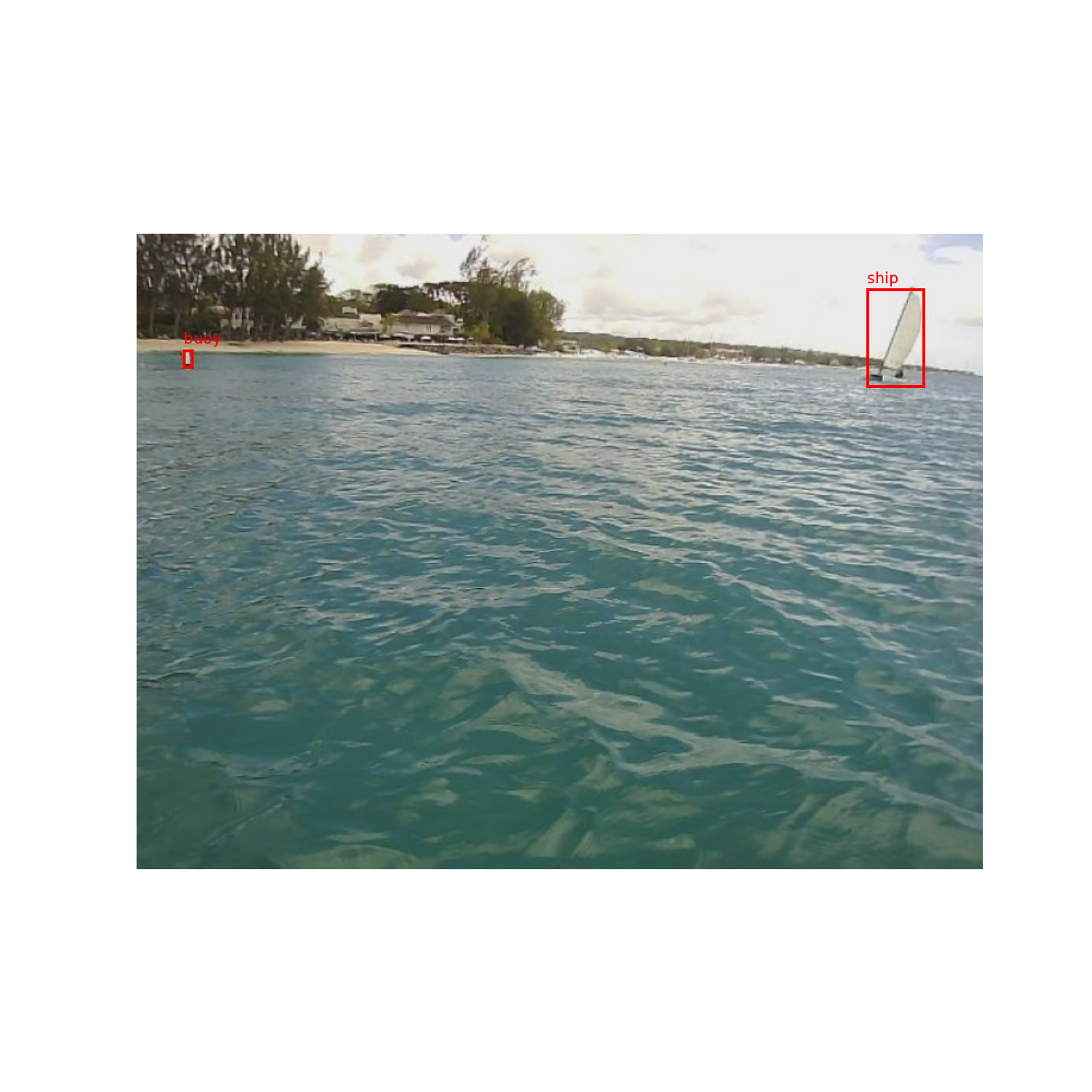}
        }
	~%
        \subfloat{
		\includegraphics[width=0.32\textwidth,trim={3cm 5.5cm 3cm 5.5cm},clip]{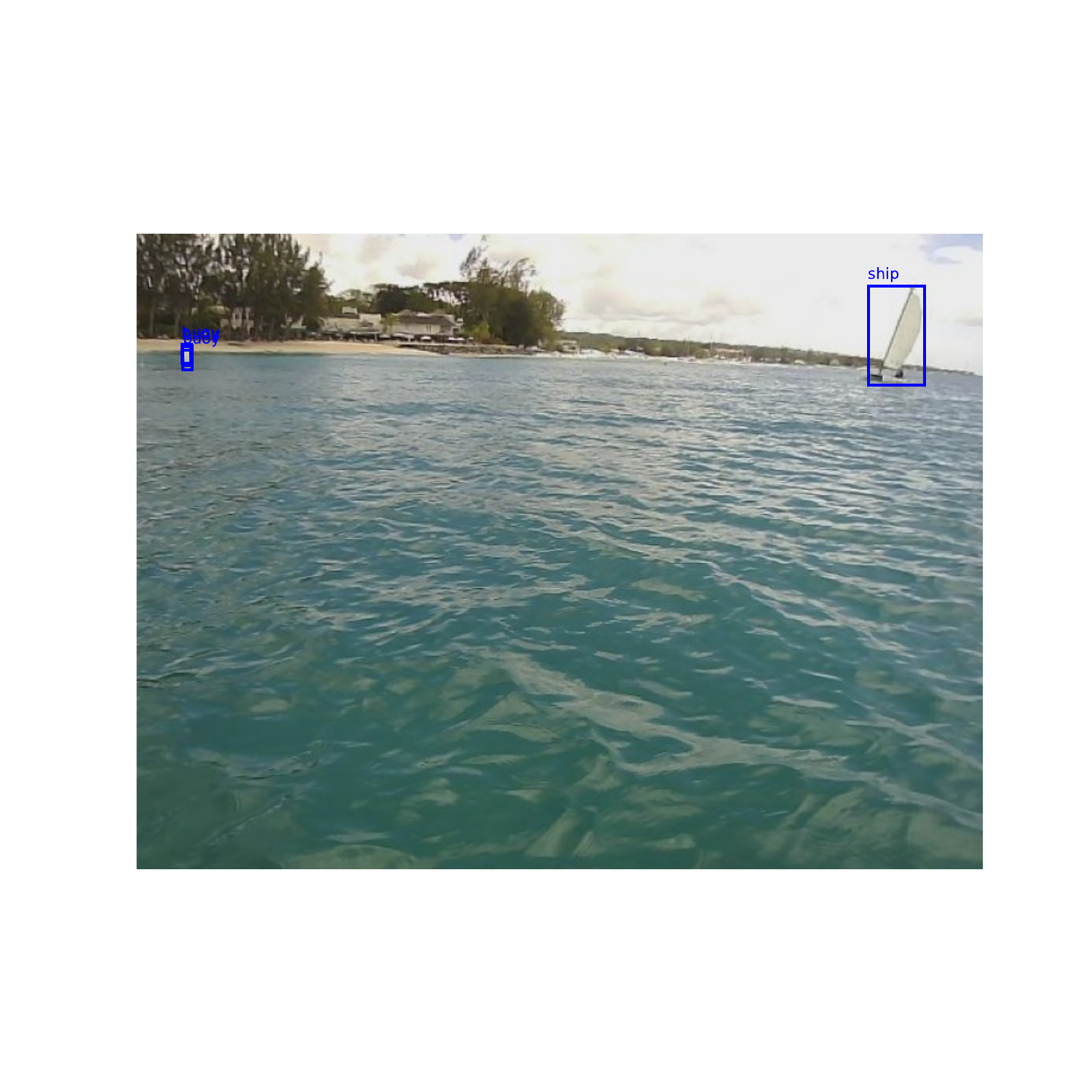}
        }
	~%
        \subfloat{
		\includegraphics[width=0.32\textwidth,trim={3cm 5.5cm 3cm 5.5cm},clip]{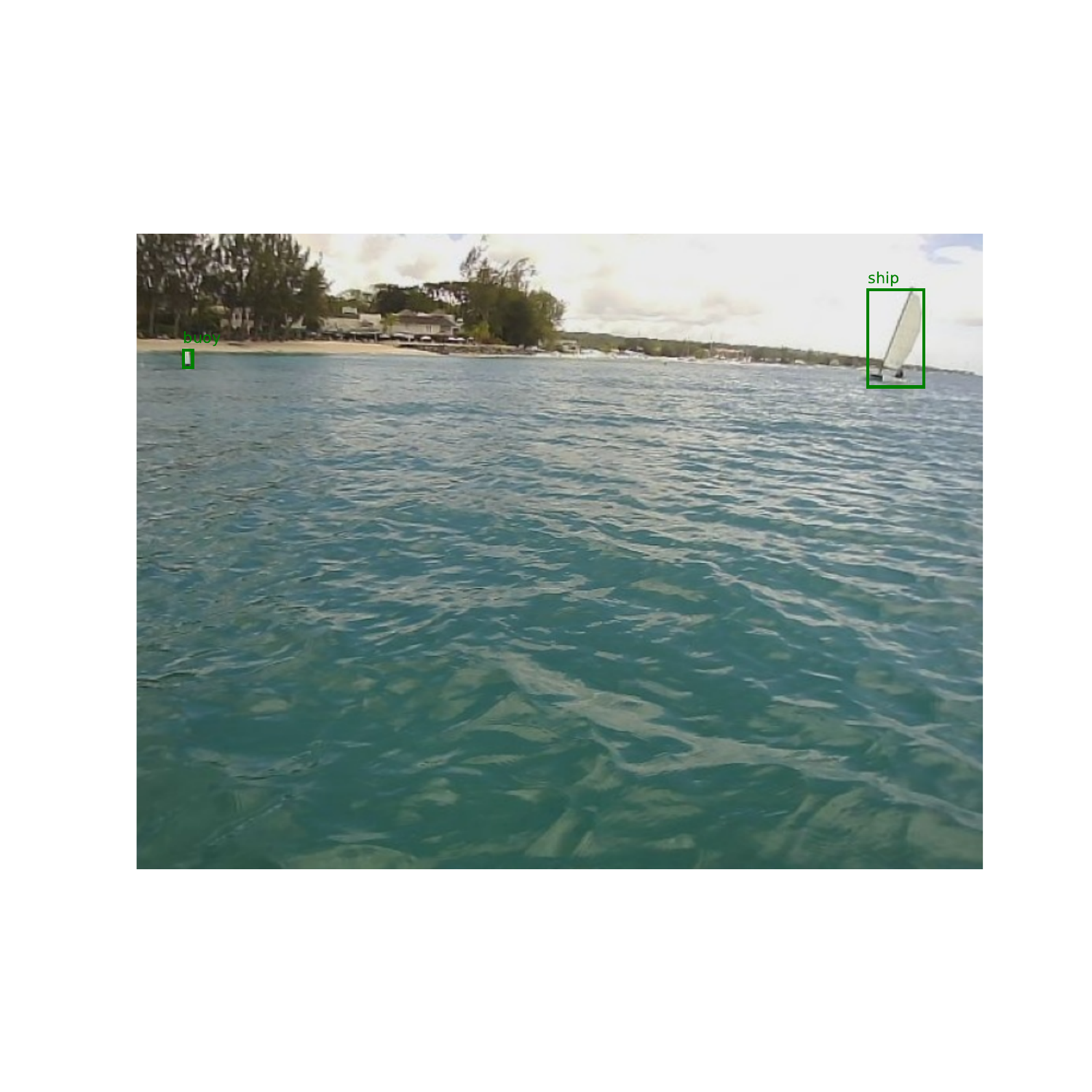}
        }

		\subfloat{
        \includegraphics[width=0.32\textwidth,trim={3cm 5.5cm 3cm 5.5cm},clip]{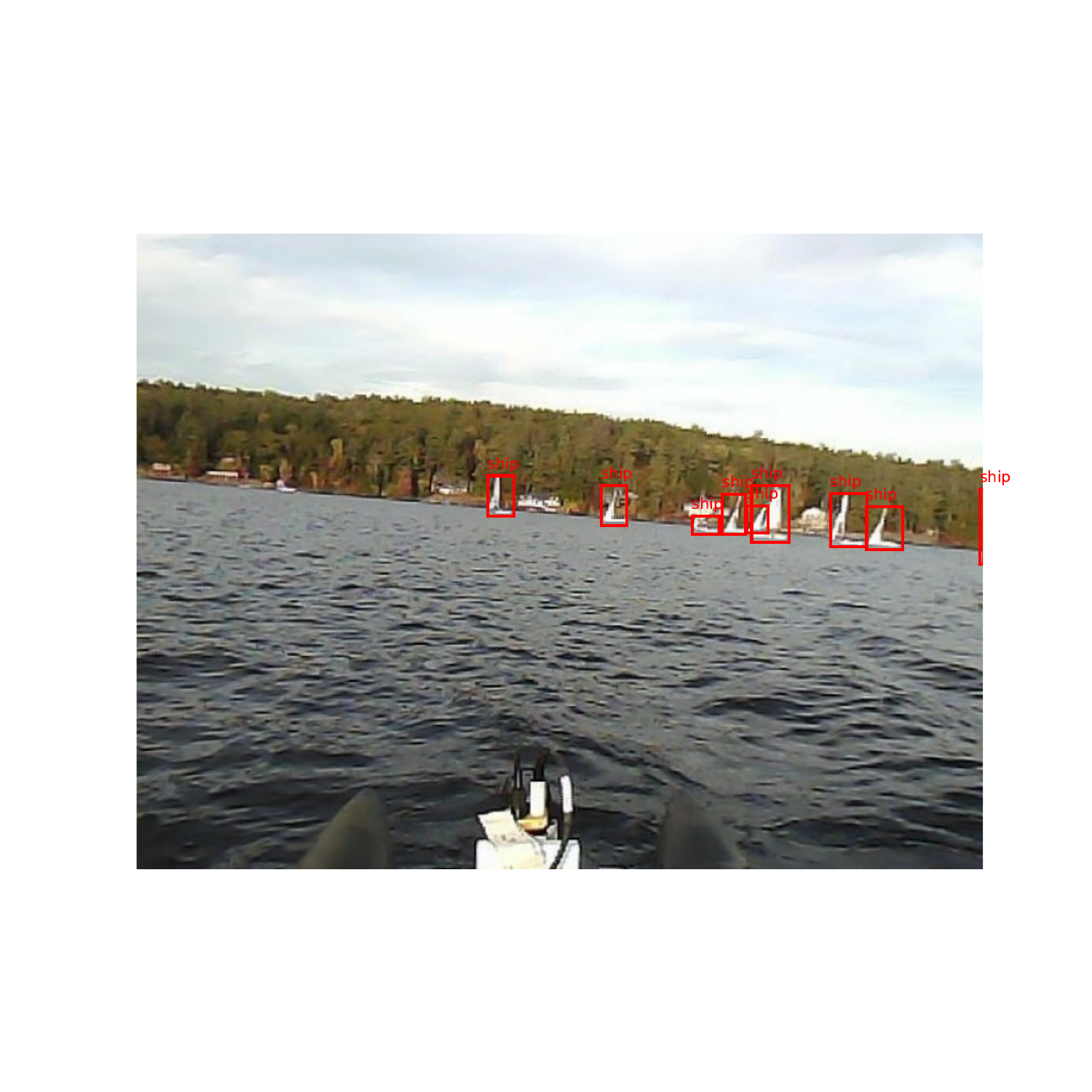}
    }
	~%
    \subfloat{
		\includegraphics[width=0.32\textwidth,trim={3cm 5.5cm 3cm 5.5cm},clip]{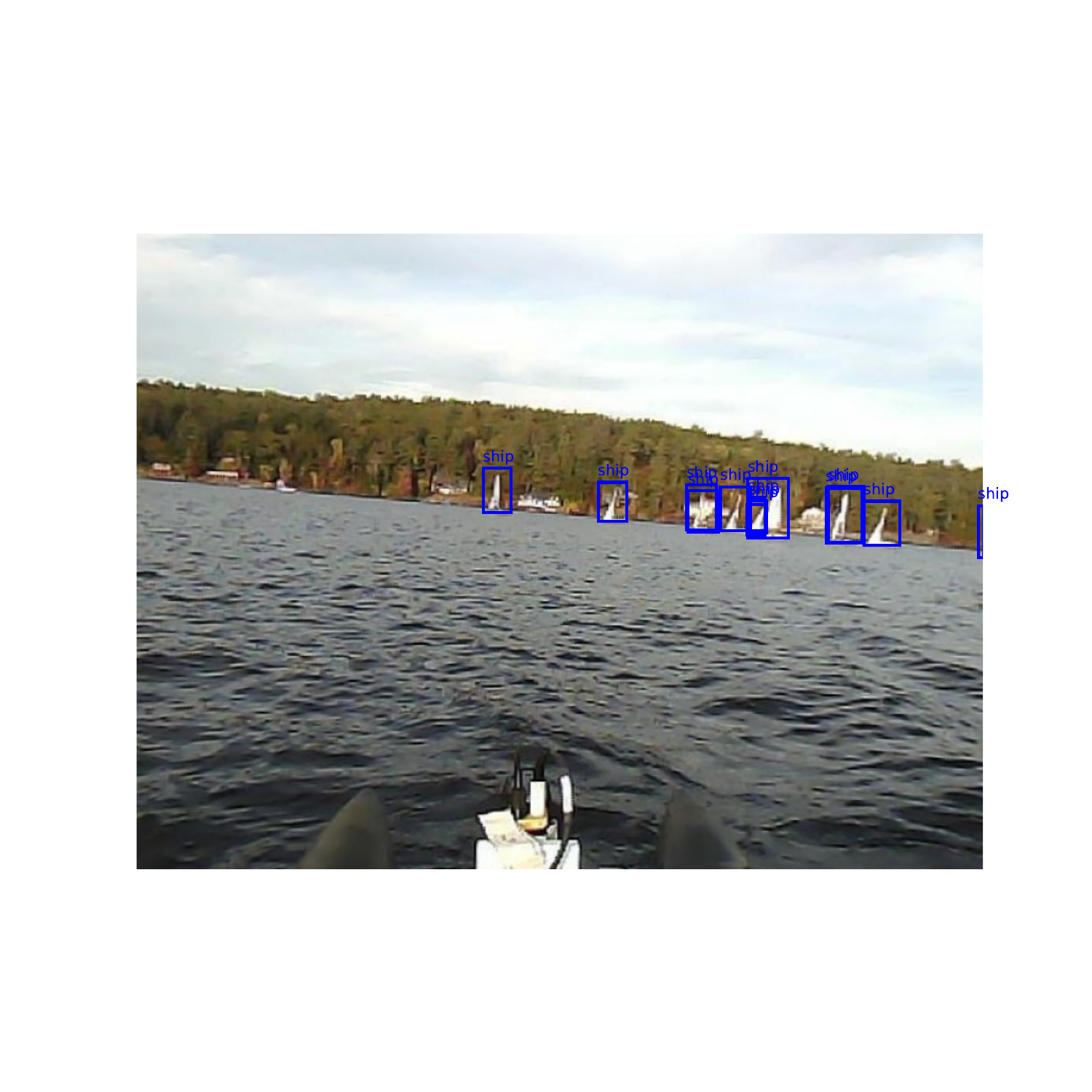}
    }
	~%
    \subfloat{
		\includegraphics[width=0.32\textwidth,trim={3cm 5.5cm 3cm 5.5cm},clip]{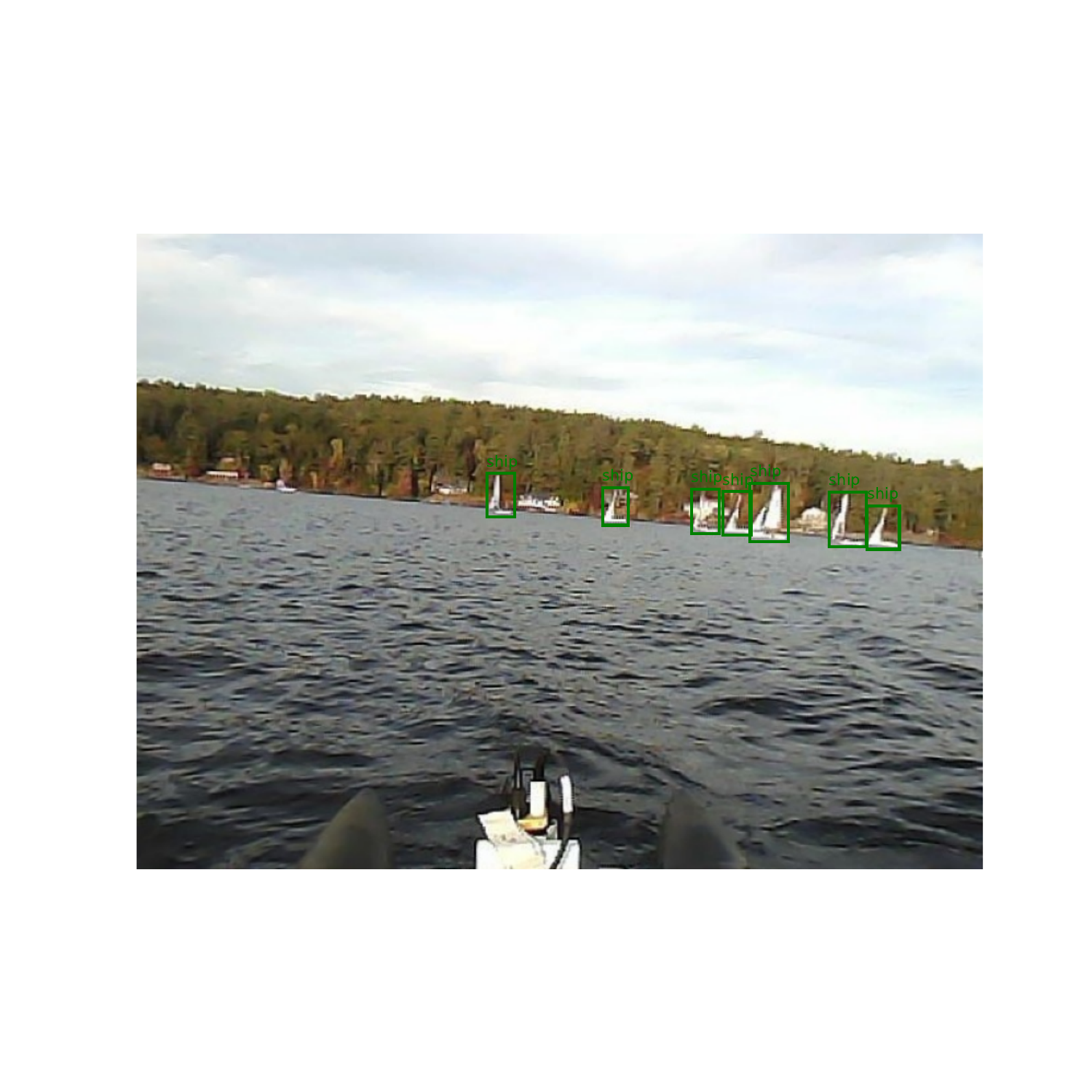}
    }
		    \subfloat{
		\includegraphics[width=0.32\textwidth,trim={3cm 5.5cm 3cm 5.5cm},clip]{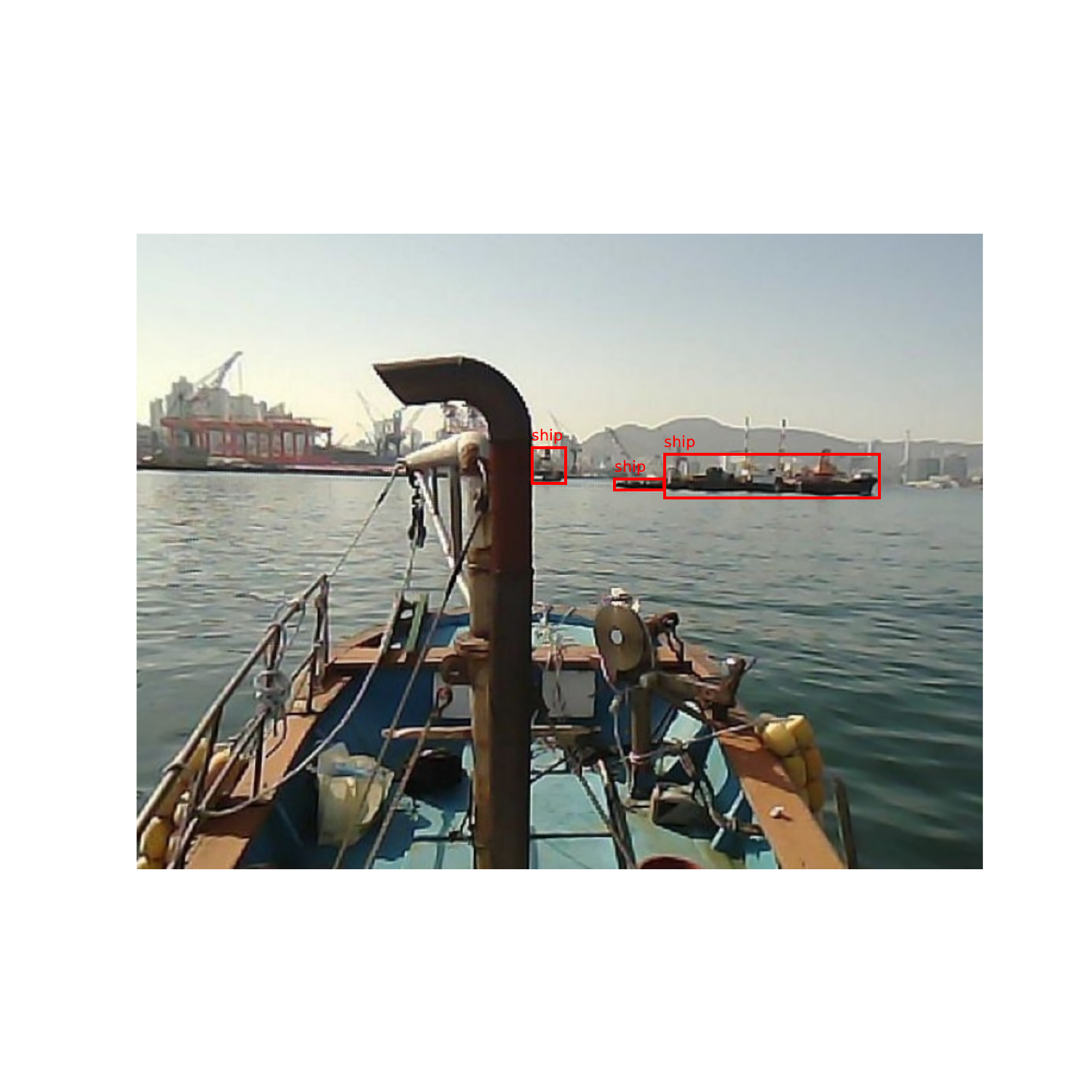}
        }
	~%
	    \subfloat{
		\includegraphics[width=0.32\textwidth,trim={3cm 5.5cm 3cm 5.5cm},clip]{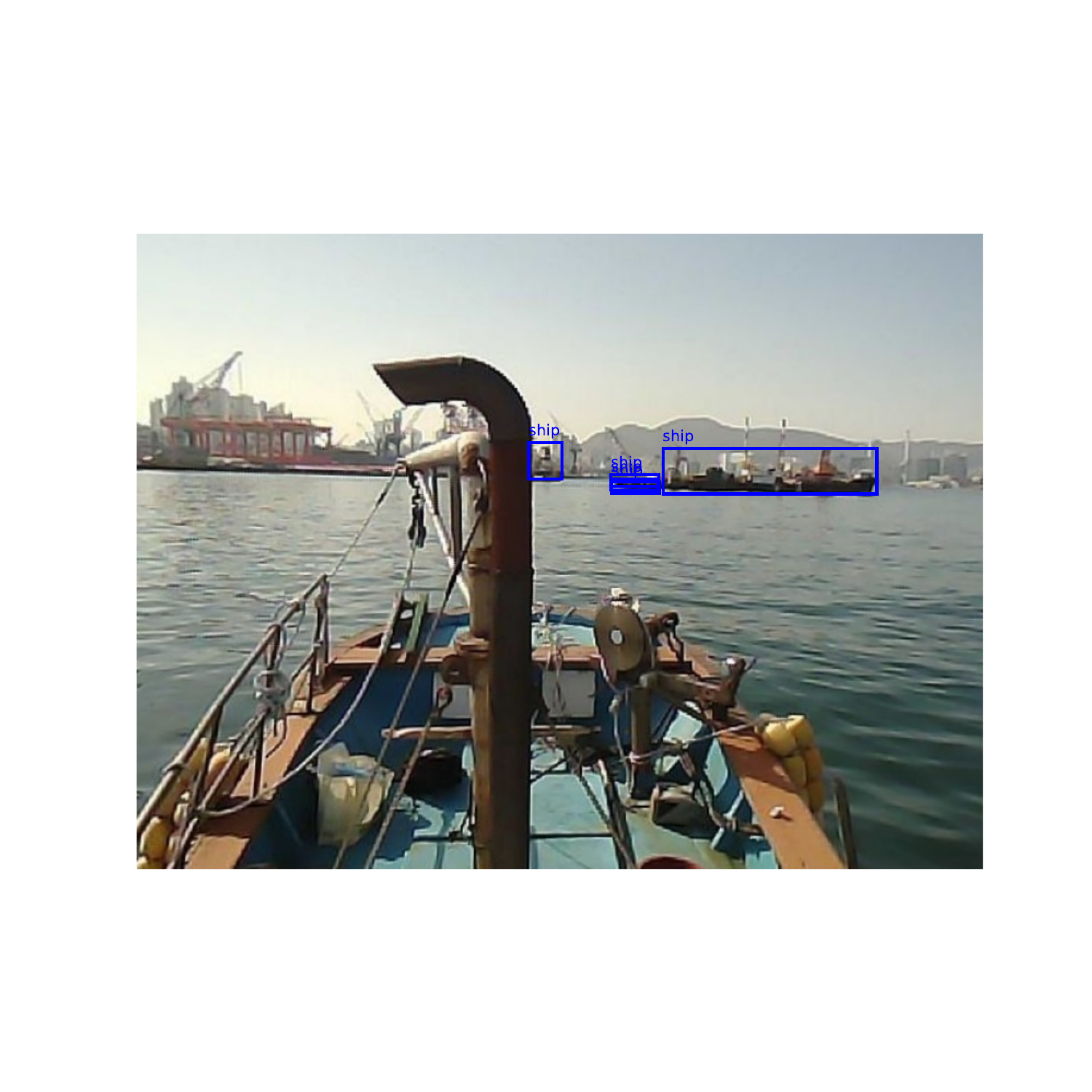}
        }
	~%
        \subfloat{
		\includegraphics[width=0.32\textwidth,trim={3cm 5.5cm 3cm 5.5cm},clip]{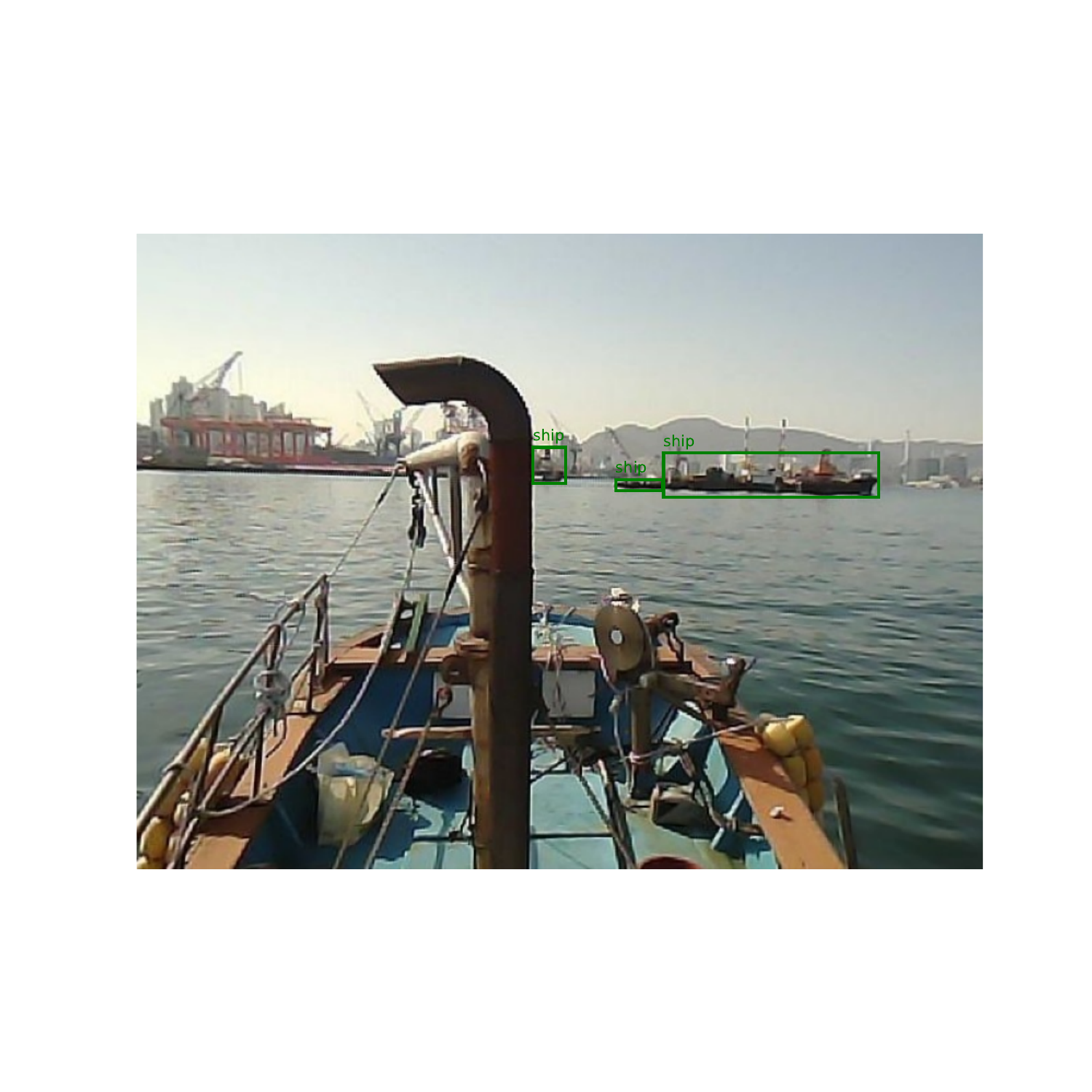}
        }
        \subfloat{
		\includegraphics[width=0.32\textwidth,trim={3cm 5.5cm 3cm 5.5cm},clip]{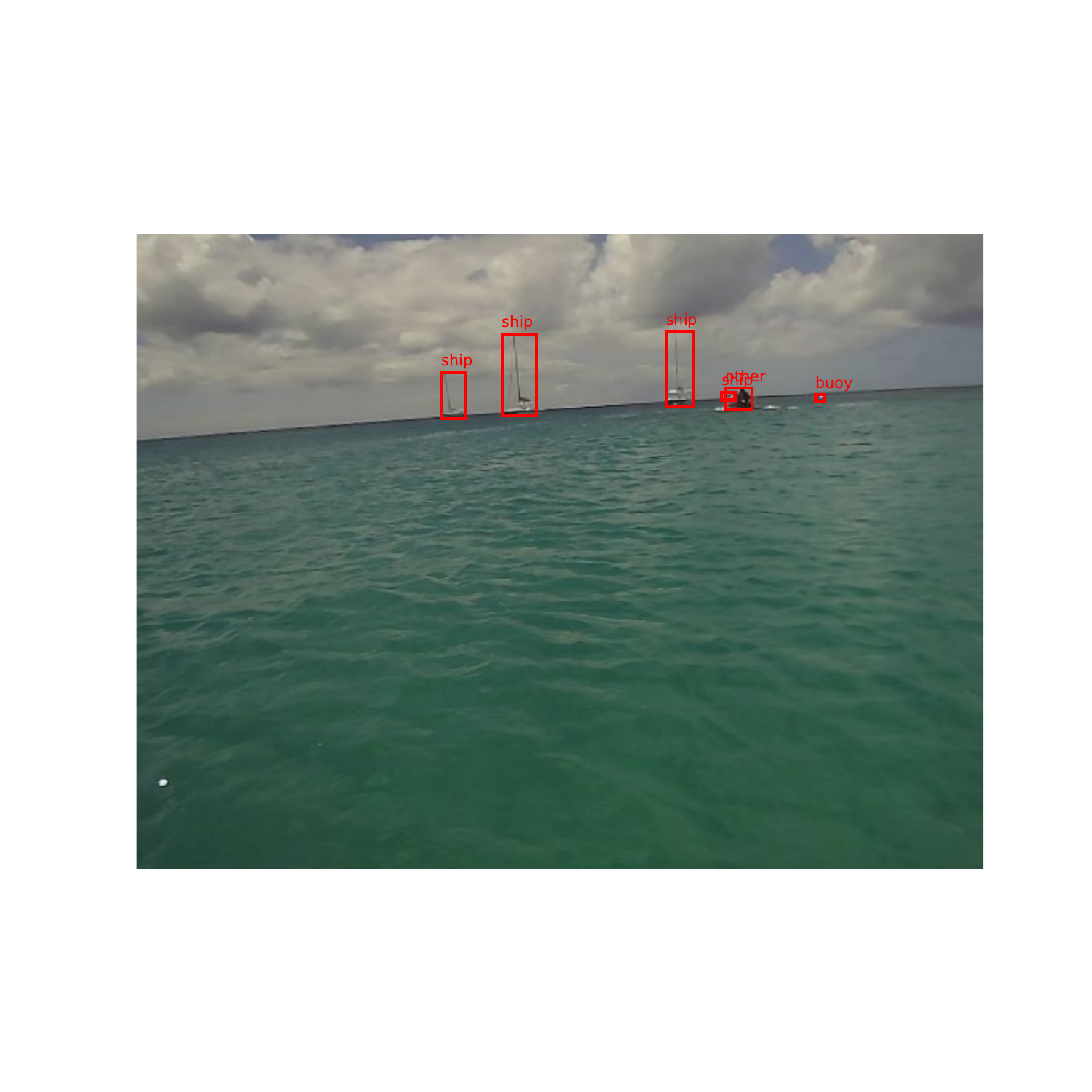}
        }
	~%
        \subfloat{
		\includegraphics[width=0.32\textwidth,trim={3cm 5.5cm 3cm 5.5cm},clip]{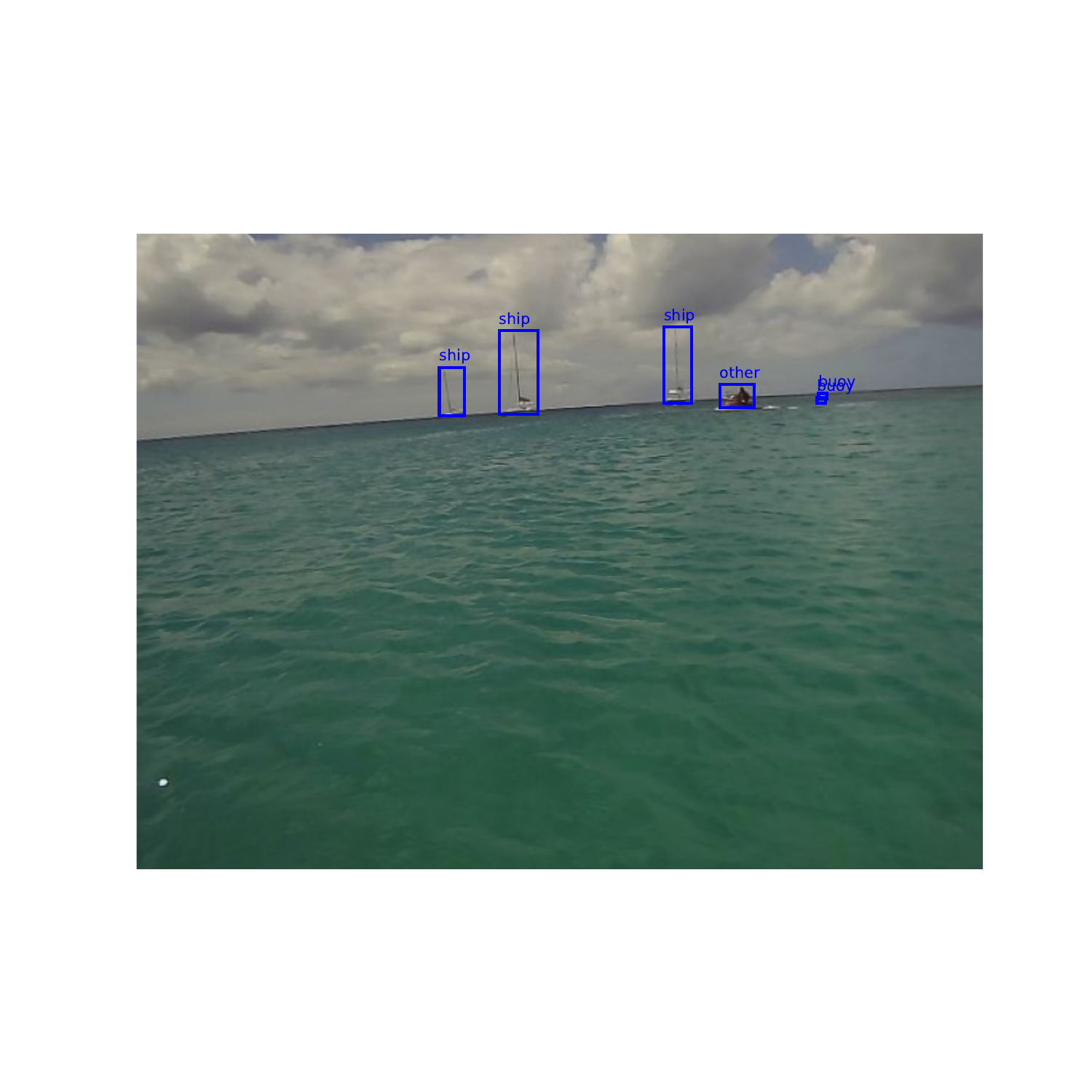}
        }
	~%
        \subfloat{
		\includegraphics[width=0.32\textwidth,trim={3cm 5.5cm 3cm 5.5cm},clip]{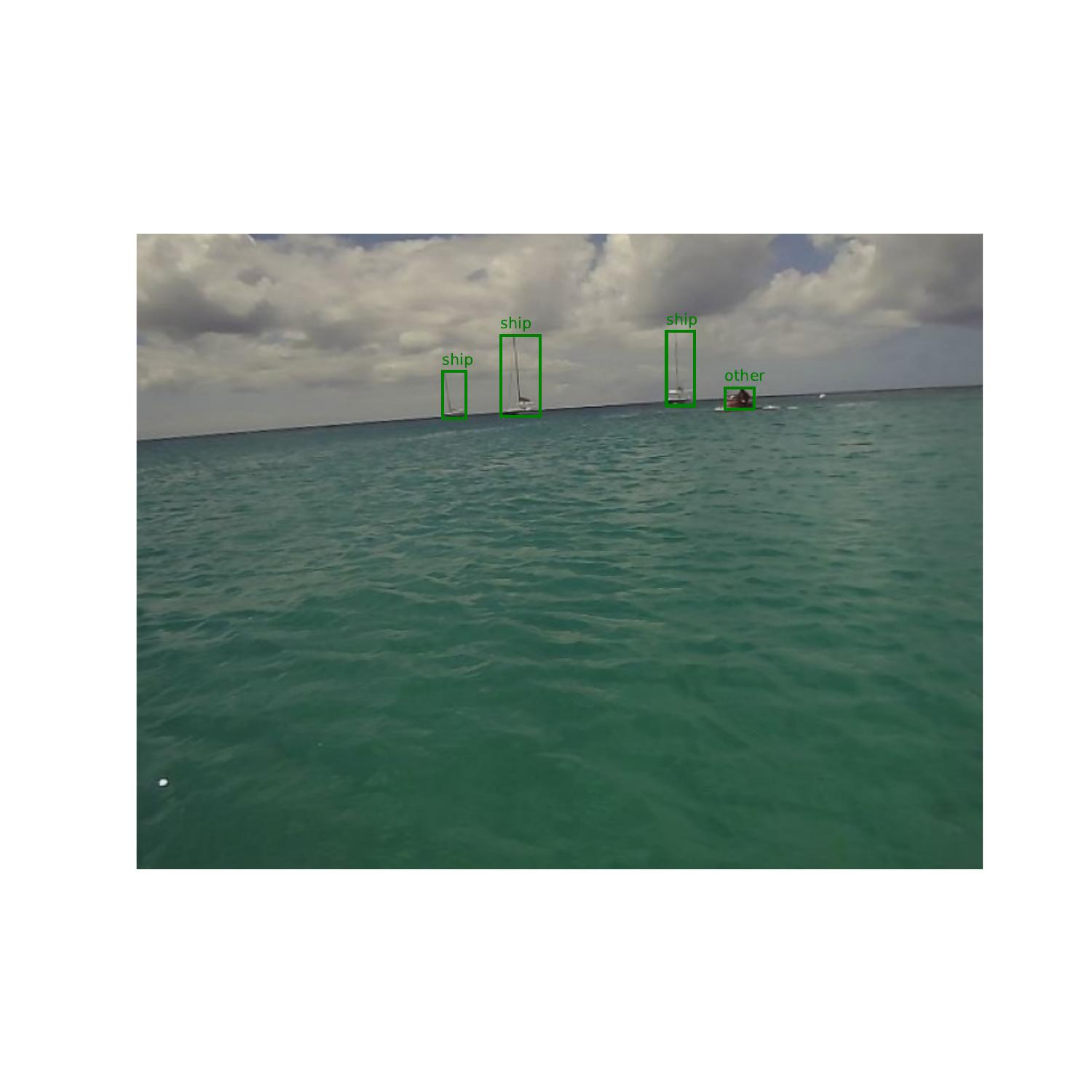}
        }

	\caption{Detection examples from the benchmark image models trained on the RGB image portion of the dataset, where the left column is groundtruth (\textit{red}), the middle column is RT-DETR (\textit{blue}), and the right column is YOLOv9 (\textit{green}). While the models did learn to predict many of the classes, there is still much room for improvement that robotic perception methods adapted to the ASV domain could begin to address using this dataset.}
	\label{fig:qual-benchmark-image}
\end{figure*}

We trained both models for 300 epochs and used the default hyperparameters from the YOLOv9 \cite{yolov9-github} and RT-DETR \revised{with HGNetv2 backbone} \cite{rtdetr-github} open-source implementations. From their reference implementation, we applied a confidence threshold of $0.25$ and an Intersection over Union (IoU) threshold of $0.45$ for non-maximum suppression to post-process outputs before compiling results. For consistent comparison across 2D object detection methods, we used the mean Average Precision (mAP) metric. Validation and test set results are in \tab{tab:image-class-comparison} and example detections are in \fig{fig:qual-benchmark-image}. The qualitative examples are from 3 of the dataset's locations: Barbados, Lake Mascoma and Busan Port, to show several multi-object encounters with ship, buoy, and other labeled objects. 

From the results of both models, qualitative and quantitative, out-of-the box models have room for improvement -- especially on the buoy and other classes. %
It is clear that (a) there are relevant image-only features to train object detection models and (b) that this dataset represents a challenging detection task, characteristic of the maritime ASV environment. 

The heterogeneity of maritime objects, variable environmental conditions, and in-water dynamics make this a difficult RGB-camera-only robotic vision problem -- one that the addition of LiDAR data can help address.  

\subsection{LiDAR-based deep learning benchmarks}

\begin{figure*}[t!]
    \centering
    
    \begin{minipage}[t]{0.3\textwidth}
        \subfloat[PointPillars]{
           \includegraphics[width=\linewidth]{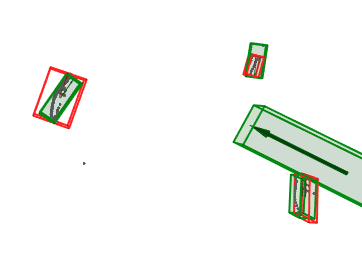}
        }
    \end{minipage}
    \hfill %
    \begin{minipage}[t]{0.3\textwidth}
        \subfloat[SECOND]{
           \includegraphics[width=\linewidth]{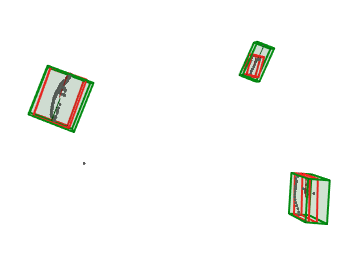}
        }
    \end{minipage}
    \hfill %
    \begin{minipage}[t]{0.3\textwidth}
        \subfloat[PointRCNN]{
           \includegraphics[width=\linewidth]{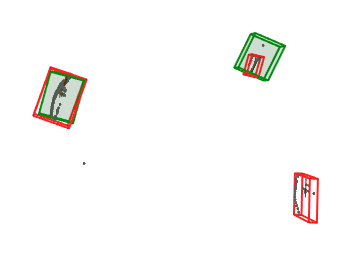}
        }
    \end{minipage}

    \begin{minipage}[t]{0.3\textwidth}
        \subfloat[PV-RCNN]{
           \includegraphics[width=\linewidth]{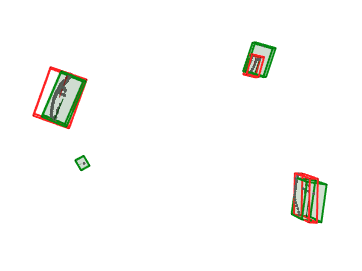}
        }
    \end{minipage}
    \hfill %
    \begin{minipage}[t]{0.3\textwidth}
        \subfloat[Voxel-RCNN]{
           \includegraphics[width=\linewidth]{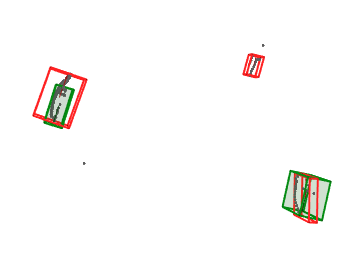}
        }
    \end{minipage}
    \hfill %
    \begin{minipage}[t]{0.3\textwidth}
        \subfloat[TED-S]{
           \includegraphics[width=\linewidth]{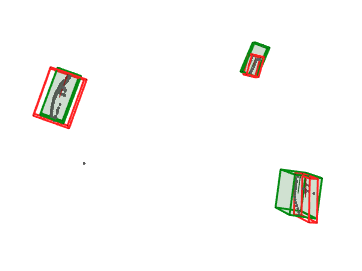}
        }
    \end{minipage}
    \hfill %
    \begin{minipage}[t]{0.3\textwidth}
        \subfloat[PointPainting]{
           \includegraphics[width=\linewidth]{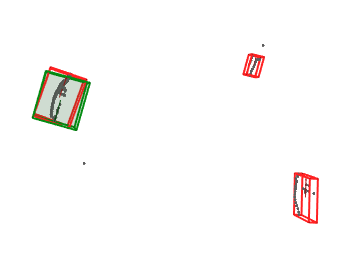}
        }
    \end{minipage}    
    \begin{minipage}[t]{0.3\textwidth}
        \subfloat[CLOCs]{
           \includegraphics[width=\linewidth]{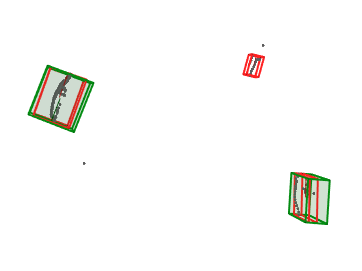}
        }
    \end{minipage}
    \hfill %
    \begin{minipage}[t]{0.3\textwidth}
        \subfloat[Focals Conv-F]{
           \includegraphics[width=\linewidth]{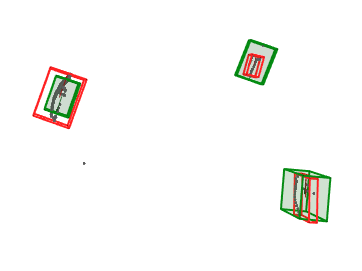}
        }
    \end{minipage}
    \hfill
    \begin{minipage}[t]{0.3\textwidth}
        \subfloat[TED-M]{
           \includegraphics[width=\linewidth]{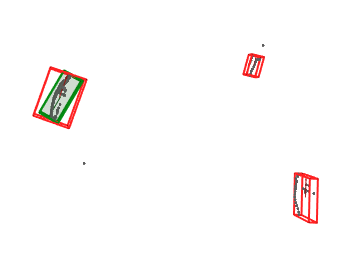}
        }
    \end{minipage}
    \hfill
    \begin{minipage}[t]{0.3\textwidth}
        \subfloat[Image of evaluated objects]{
           \includegraphics[width=\linewidth]{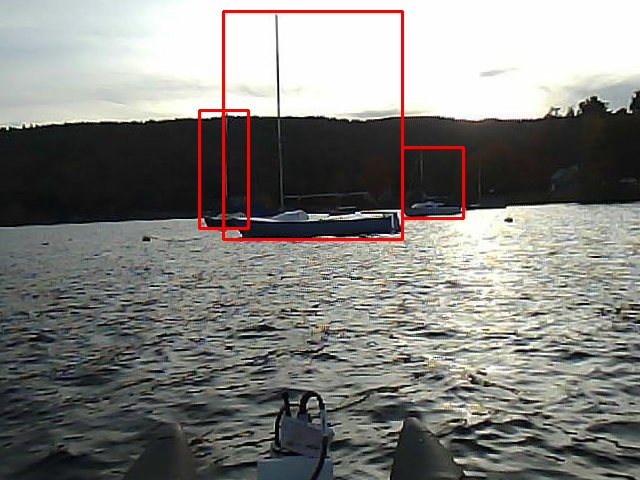}
        }
    \end{minipage}
    
    \caption{\revised{Qualitative comparison of LiDAR-based and fusion object detection benchmarks tested on our dataset, shown from a bird's-eye view. The evaluated objects belong to the \textbf{ship} class within the FoV of both the camera and LiDAR, with ground truth bounding boxes depicted in red and predicted bounding boxes in green.}   }
    \label{fig:lidar-qual}
\end{figure*}

We analyzed the performance of LiDAR based methods on   3D and Bird's Eye View (BEV) detection in the maritime domain. We selected $6$ state-of-the-art LiDAR-only 3D object detection models  -- based on the following categorizations: 

\begin{itemize}
    \item \textbf{Voxel-based}: \pointpillars~\cite{Lang2018PointPillarsFE}, \second~\cite{second-2018}, \voxelRCNN~\cite{deng2020voxel-rcnn}, \revised{\tedSingle~\cite{TED-2023}};
    \item \textbf{Point-based}: \pointRCNN~\cite{point-rcnn-2019};
    \item \textbf{Point-voxel-based}: \pvRCNN~\cite{pv-rcnn-2020}.
\end{itemize}
\noindent \revised{We adapted open-source libraries, including OpenPCDet~\cite{openpcdet2020}, as well as implementations from other repositories~\cite{second-2018, CLOCs-2020, TED-2023}, to enable benchmark comparisons tailored to our maritime dataset.}

We followed each paper's guideline on setting the hyperparameters and used the suggested values when possible. We increased the point cloud range and the voxel size to account for the longer distances between the ASV and obstacles.
We trained each method for 200 epochs with early stopping once the model stopped improving.

For consistent comparisons across LiDAR-only and fusion methods, we evaluated and reported performance for objects within both the camera and LiDAR FoV, consistent with KITTI benchmarks~\cite{KITTI-raw-2013}. Note that our ground truth labeling provides a \SI{360}{\degree} FoV from the LiDAR used on our platforms. We compared performance based on Average Precision (AP) at IoU thresholds of $0.7$ and $0.5$, evaluated for both BEV and 3D detection. We focused on the \textbf{ship} class for performance comparison due to the sparsity and challenges posed by features associated with small objects in the \textbf{buoy} and \textbf{other} classes, which LiDAR typically returns as $1–2$ points, as shown in \fig{fig:pixel-area-point-count}(bottom).

The evaluation results (\tab{tab:lidar-comparison}) for BEV detection are comparable to those of previous work \cite{lin-maritime-lidar-2022}, which used simulation results tested on 2D. Instead, our benchmark comparison extends applicability to the 3D domain with real-world data. Among LiDAR-only methods, \revised{\tedSingle~achieved the highest performance across both BEV AP and 3D AP metrics, outperforming other state-of-the-art approaches. These strong results may be attributed to its transformation-equivariant sparse convolution pooling and transformation-invariant voxel pooling modules, which enable learning of robust, transformation-equivariant voxel features. Additionally, its distance-aware data augmentation strategy enhances detection of distant objects—an important characteristic for in-water maritime scenarios. Aside from \tedSingle, \second~consistently demonstrated strong BEV AP performance.} This may be attributed to its voxel-based representation and efficient sparse convolution, which effectively captures large-scale geometric features. These characteristics make \second~particularly robust for BEV representations, where preserving spatial structure is critical. In contrast, \voxelRCNN~performed relatively well in 3D AP metrics. Its performance stems from leveraging high-resolution voxel grids combined with an accurate region proposal network, enabling more precise object localization in 3D space. On the other hand, \pointRCNN, which relies solely on raw point clouds and bypasses voxelization, is limited in its ability to efficiently extract global features, making it less effective in sparse maritime environments. Meanwhile, \pvRCNN, employing a hybrid approach that combines voxel-based feature extraction (for global context) with raw point-cloud features (for local precision), was better than \pointRCNN~by balancing global and local feature extraction. \pointpillars~showed relatively lower performance, particularly in 3D AP.. This is likely due to its reliance on a pillar-based pseudo-image representation that flattens vertical structure early in the pipeline.

\subsection{Fusion-based deep learning benchmarks}
We evaluated the following $3$ state-of-the-art 3D object detection fusion methods: 
\begin{itemize}
    \item \textbf{Sequential fusion}: \pointpainting~\cite{pointpainting-2020} based on DeepLabV3 \cite{chen2017rethinking} and \pointpillars;
    \item \textbf{Decision-level fusion}: \clocs~\cite{CLOCs-2020} based on the detection of YOLOv9~\cite{YOLOv9-2024} and \second; and
    \item \textbf{Feature-level fusion}: \focalconvf~\cite{focal-sparse-2022}, \revised{\tedMulti~\cite{TED-2023}}.
\end{itemize}

\noindent \revised{As shown in Table~\ref{tab:lidar-comparison}, \focalconvf~and \tedMulti~achieved the best overall results among fusion-based methods across both BEV AP and 3D AP metrics. \focalconvf's effectiveness may be attributed to the integration of complementary sensor modalities through focal sparse convolutions, enabling robust spatial reasoning and precise object localization. \tedMulti~builds upon \tedSingle~by incorporating appearance features from RGB images, offering further improvements in some cases. However, as noted in \cite{TED-2023, Valid-2025}, our results indicate that incorporating camera data does not uniformly enhance detection performance. Notably, TED-M’s marginal gains come at the cost of increased system complexity, as it requires generating pseudo-LiDAR points from camera images. These image-derived points depend on depth estimation~\cite{PENet-2021}, which can be particularly noisy for distant objects. This noise partly explains why distant or hard-to-detect targets sometimes see minimal benefit—or even slight performance degradation—with fusion. Such drops can also be attributed to sensor misalignment~\cite{mislaighment1-2023, mislaignment2-2022}, a challenge observed in the maritime domain and further discussed in Section~\ref{sec:discussion}.} Other fusion methods such as \clocs~and \pointpainting~performed competitively but their performance lagged at stricter IoU thresholds for 3D AP. \clocs, which integrates predictions from multiple backbones, showed reduced performance in scenarios requiring high precision, likely due to a weaker emphasis on fine-grained feature alignment. Similarly, \pointpainting's reliance on segmentation quality and alignment resulted in lower performance in 3D AP metrics compared to \focalconvf.

We also provide a qualitative analysis across the 3D Object Detection benchmarks. \fig{fig:lidar-qual} illustrates the results of a sequence (Mascoma Lake) on our \textit{open-set} test split, which were excluded from all training steps. Consistent with the quantitative evaluation, \tedSingle~and \focalconvf~exhibited strong performance for ship detection.

\section{Discussion}\label{sec:discussion}

\begin{figure}[t!]
    \centering
    \begin{minipage}{0.46\columnwidth}
        \subfloat{
            \includegraphics[width=\textwidth]{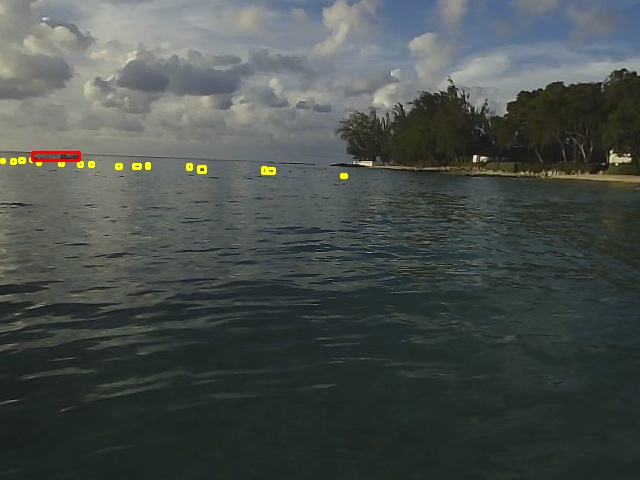}
        }
    \end{minipage}
    \centering
    \begin{minipage}{0.485\columnwidth}
        \subfloat{
            \includegraphics[width=\textwidth]{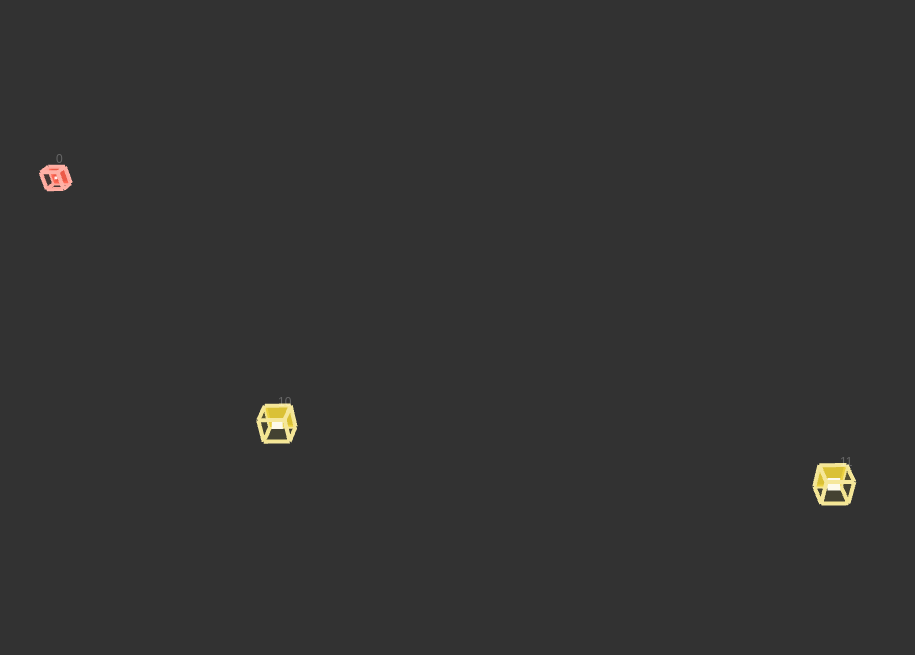}
        }
    \end{minipage}
    \caption{Challenging sparsity example from Barbados sequence in our dataset -- (\textit{yellow}) buoys, (\textit{orange}) floating dock. Although there are many objects in the image (\textit{left}), the LiDAR measurement has only $3$ objects while each of them has $1$ point inside the bounding box (\textit{right}).}
    \label{fig:sparse-point-cloud}
\end{figure}

Based on our contribution of the first multi-modal dataset in the maritime domain and its utility for deep learning-based approaches, we identify and provide insights into the challenges and open problems for future tasks aimed at enhancing robust perception systems in maritime environments. Furthermore, we hope this dataset will provide the research community with a starting point to develop robust, novel methods for ASV perception. Given this work, it is our continuing hypothesis that multi-modal methodologies are essential for the development of robust ASV situational awareness given in-water dynamics, environment heterogeneity, and failures being inadmissible. In the following paragraphs, we will discuss open challenges to the development of these methods, which include: sparsity, generalizability, and misalignment. 

Maritime environments often feature \textbf{sparse} point clouds due to objects located at long distances and unstable measurements affected by the motions of both ego and target vehicles, as noted by \cite{jeong-iros2021}. Current detection methods struggle to learn features from such minimal data, particularly for buoys and small objects. This highlights the need for models capable of accurately detecting and classifying objects even under sparse conditions. For example, as shown in \fig{fig:sparse-point-cloud}, even a single LiDAR-detected point representing an object such as a buoy could lead to a collision if ignored. This differs from other domains where they often use thresholds for a minimum number of points.

\begin{figure}[t!]
    \centering

    \subfloat[ASV in water: sensor alignment (\textit{left}) vs. misalignment (\textit{right}) due to motion-induced effects.
        \label{fig:in-water-comparison}]{
        \centering
        \begin{minipage}{0.49\columnwidth}
            \includegraphics[width=\textwidth]{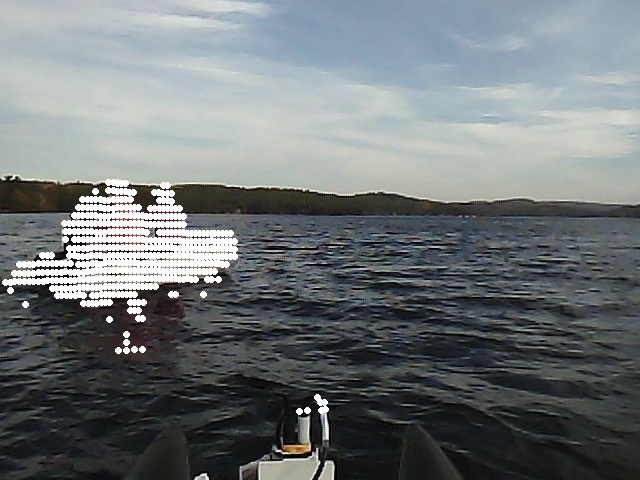}
        \end{minipage}
        \hfill
        \begin{minipage}{0.49\columnwidth}
            \includegraphics[width=\textwidth]{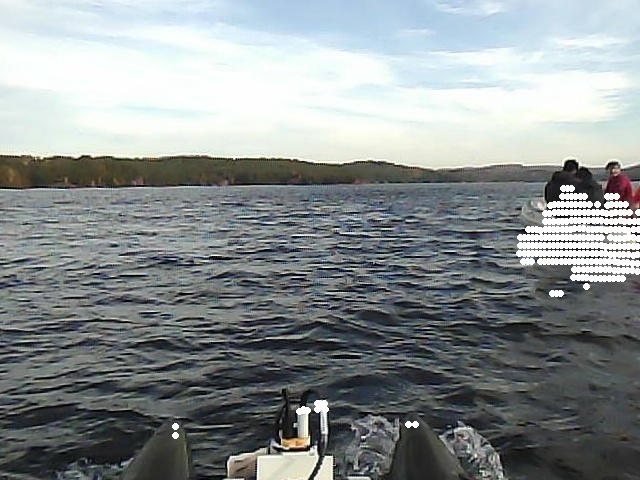}
        \end{minipage}
        }

    \subfloat[\revised{ASV on shore: consistent alignment (\textit{left} vs. \textit{right}) of fused sensors in a static environment.} \label{fig:shore-comparison}]{
        \centering
        \begin{minipage}{0.49\columnwidth}
            \includegraphics[width=\textwidth]{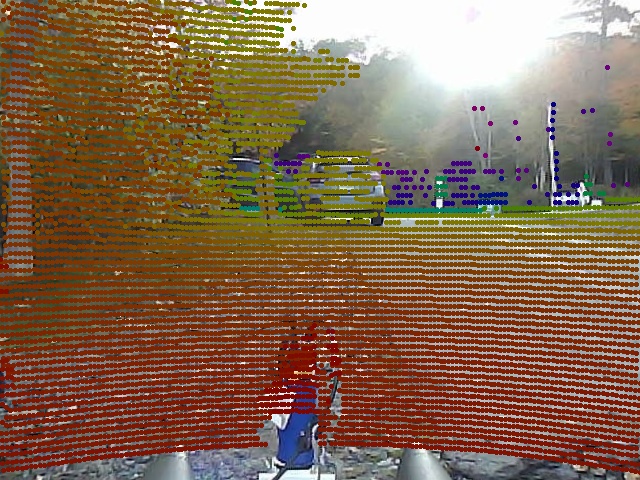}
        \end{minipage}
        \hfill
        \begin{minipage}{0.49\columnwidth}
            \includegraphics[width=\textwidth]{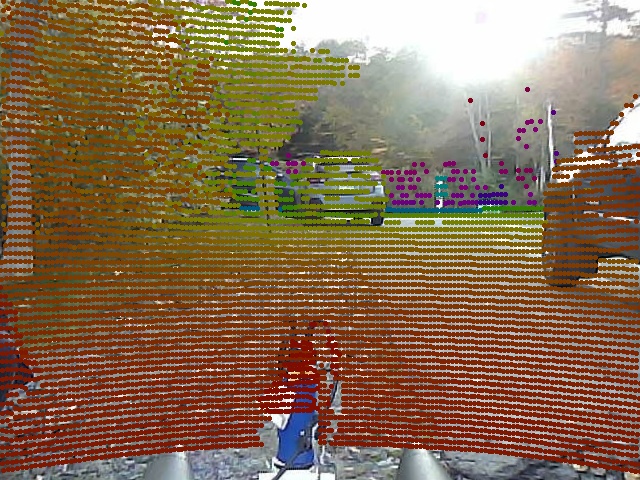}
        \end{minipage}
        }

    \subfloat[\revised{Angular motion: roll (\textit{left}) and pitch (\textit{right}) comparison between in-water and on-shore conditions.}
        \label{fig:angular-comparison}]{
        \centering
        \begin{minipage}{0.493\columnwidth}
            \includegraphics[width=\textwidth]{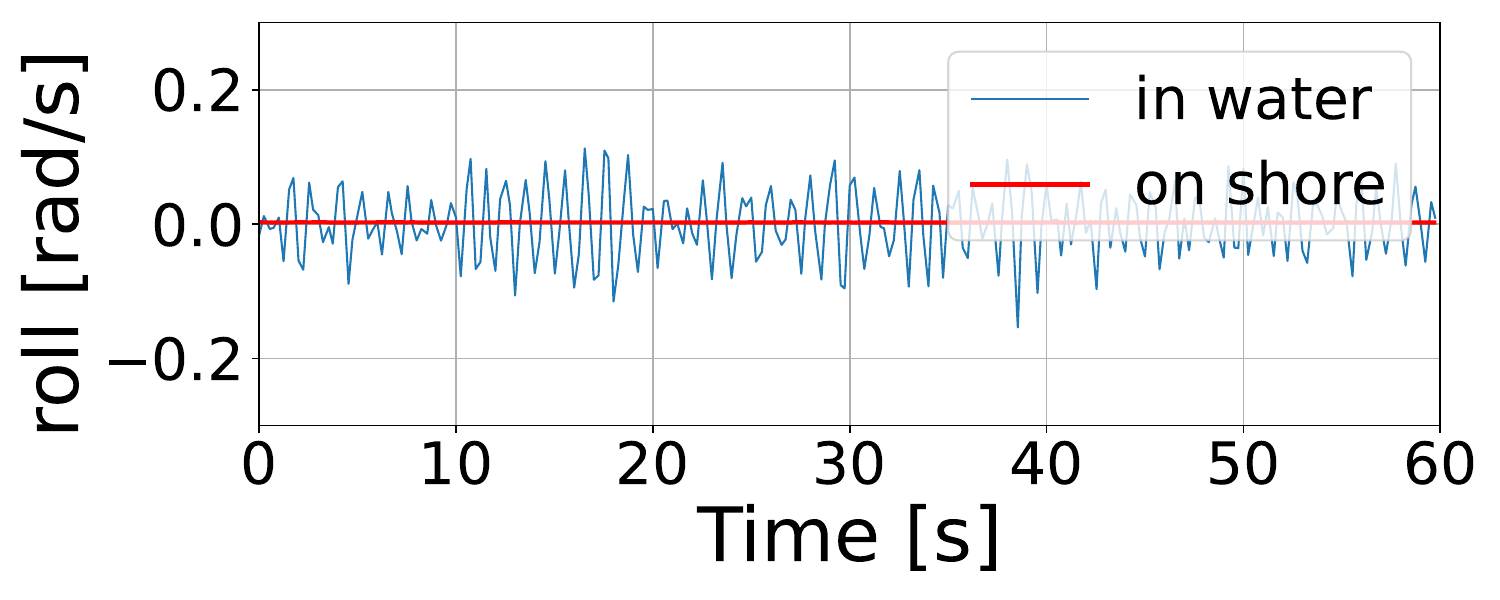}
        \end{minipage}
        \hfill
        \begin{minipage}{0.493\columnwidth}
            \includegraphics[width=\textwidth]{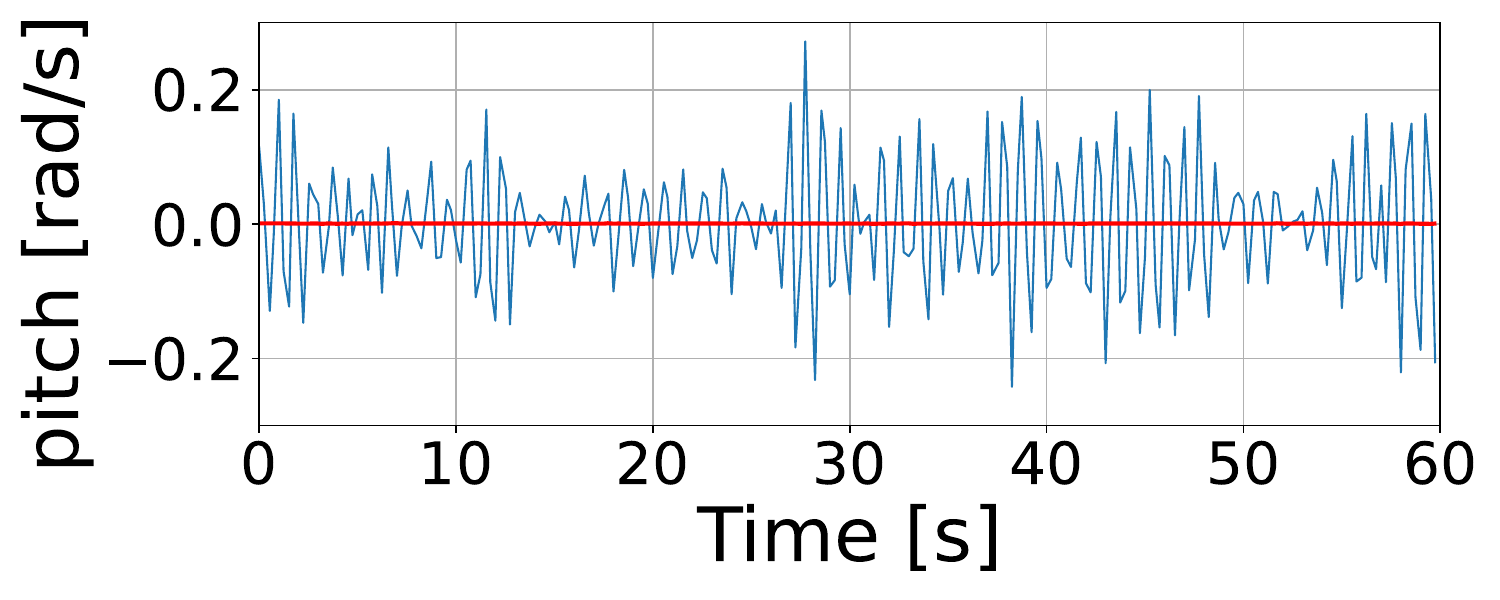}
        \end{minipage}
        }

    \caption{\revised{Comparison of LiDAR-camera alignment and motion-induced effects during the same deployment date. (a) Sensor misalignment occurs in water due to motion, despite the same calibration; (b) Stable alignment is preserved on shore; and (c) quantitative analysis of motions (roll, pitch) during $\SI{60}{\second}$ as a sequence time. Note that the point clouds are colored for the best visibility.}}
    \label{fig:misalignment}
\end{figure}

\textbf{Generalizability} remains another significant open challenge. For instance, LiDAR-based deep learning models (and many real-time image-based deep learning models) use anchors, which represent the predefined dimensions of bounding boxes, to enhance the accuracy and efficiency of object predictions. However, as shown by the range of the length (\SI{0.1}{\meter} - \SI{123.5}{\meter}), width (\SI{0.1}{\meter} - \SI{81.1}{\meter}), and height (\SI{0.1}{\meter} - \SI{35.7}{\meter}) of the LiDAR annotations, object sizes in the maritime domain vary greatly -- from small fishing boats to large commercial ships -- all defined as ``ships'' under international maritime traffic rules~\cite{colreg}. Therefore, if one does not carefully choose hyperparameter values, such as anchor dimensions, it may degrade the performance of detection benchmarks. Furthermore, all the point cloud detection benchmarks we utilized in our study relied on preset point cloud ranges. However, we observed cases where detected point clouds lay beyond the sensor's nominal range (e.g., exceeding \SI{120}{\meter}), particularly in open sea conditions. Aligned with maritime navigation principles on focusing on early detection and taking large actions in ample time, one must thoughtfully select predefined ranges and sizes. These parameters strictly constrain current learning-based methods, underscoring the need for models, such as anchor-free approaches, which can adapt to varying detection ranges and object dimensions.

\revised{Another challenge for generalization is \textbf{class imbalance}, as noted in Section~\ref{sec:dataset-composition}. This imbalance naturally reflects real-world coastal navigation environments. To mitigate its effects during model development, we recommend incorporating class-aware strategies such as targeted data augmentation (e.g., oversampling of rare object classes, copy-paste methods, or simulation-based generation) and loss reweighting approaches (e.g., focal loss or class-balanced loss functions). These methods can enhance detection performance on minority classes without compromising the dataset’s representativeness. Furthermore, although the current version represents an initial contribution to the community, there are strategies to address the limited number of samples collected under \textbf{low-light conditions}. It is possible to supplement the dataset with style-transfer methods or domain adaptation techniques to simulate nighttime environments and improve model robustness across varying illumination scenarios.}

Robustness against \textbf{misalignment} presents additional open challenges in the maritime domain. As observed in ground-based applications~\cite{mislaighment1-2023, mislaignment2-2022}, \revised{spatial and temporal misalignment is also prevalent in maritime environments. This misalignment arises from factors such as noisy extrinsic parameters and the relative motion between ego and target vehicles on the water surface, as illustrated in \fig{fig:misalignment}. We compared ASV behavior in in-water versus on-shore conditions. As shown in \fig{fig:angular-comparison}, during the same deployment operation, the in-water scenario exhibited significantly higher motion variability than the on-shore case (Levene’s test: $p$-value $< 0.01$ for both roll and pitch), leading to misalignments.} Furthermore, mechanical misalignments are particularly difficult to correct onboard due to the lack of fixed environmental features and the continuous motion caused by hydrodynamic forces. These challenges underscore the need for online, in-water calibration methods to improve system robustness.

In addition, annotations in the maritime domain naturally suffer from misalignment. Our dataset primarily considers $z$-axis orientation (i.e., yaw) during the labeling process. However, pitch and roll can significantly impact object detection and state estimation---especially in maritime settings, where dynamic and non-stationary conditions differ greatly from those in other domains (e.g., flat road surfaces). Generating accurate ground truth for pitch and roll remains a major challenge but is essential for improving detection and tracking performance in such environments. Addressing this open problem is likely to be a key prerequisite for developing robust multimodal fusion methods in maritime surface applications.

\section{Conclusion}\label{sec:conclusion}
This paper introduces the first publicly accessible multi-modal perception dataset for autonomous maritime navigation, focusing on in-water obstacles within aquatic environments to enhance situational awareness for ASVs. Our dataset, which includes a diverse range of in-water objects encountered under varying environmental conditions, aims to bridge the research gap in marine robotics by providing a multi-modal, annotated, and ego-centric perception dataset for object detection and classification. We also demonstrate the applicability of the proposed dataset using open-source deep learning-based perception algorithms that have proven successful in other domains. Additionally, the development and analysis of this dataset offer foundational insights for advancing perception tasks in the maritime domain.

\revised{Future work will focus on designing adaptable and robust deep learning models and systems capable of addressing domain-specific complexities while aligning with maritime best practices.   Alongside this, we also plan to integrate additional sensor configurations under diverse weather conditions (e.g., rain, snow), such as marine RADAR and wide-field-of-view cameras for supplementary data collection.   Furthermore, we plan to extend this work to the object tracking task and continue to explore multi-modal modeling for ASV perception using this dataset. These advancements will be crucial for enhancing situational awareness, safety, and efficiency in real-world autonomous maritime systems, addressing high-impact societal needs such as search and rescue, environmental monitoring, and transportation.}

\section*{ACKNOWLEDGMENT}
We would like to thank Kizito Masaba, Julien Blanchet, Haesang Jeong, and Capt. Jinmyeong Lee for help with field experiments, and Korea Maritime and Ocean University, McGill Bellairs Research Institute, and the Eliassen family for our access to the experimental sites. We would like to also thank Dongha Chung and Jinhwan Kim from KAIST for providing tips on publicly available datasets. 

{\renewcommand{\markboth}[2]{}\printbibliography}

\begin{IEEEbiography}[{\includegraphics[width=1in,height=1.25in,clip,keepaspectratio]{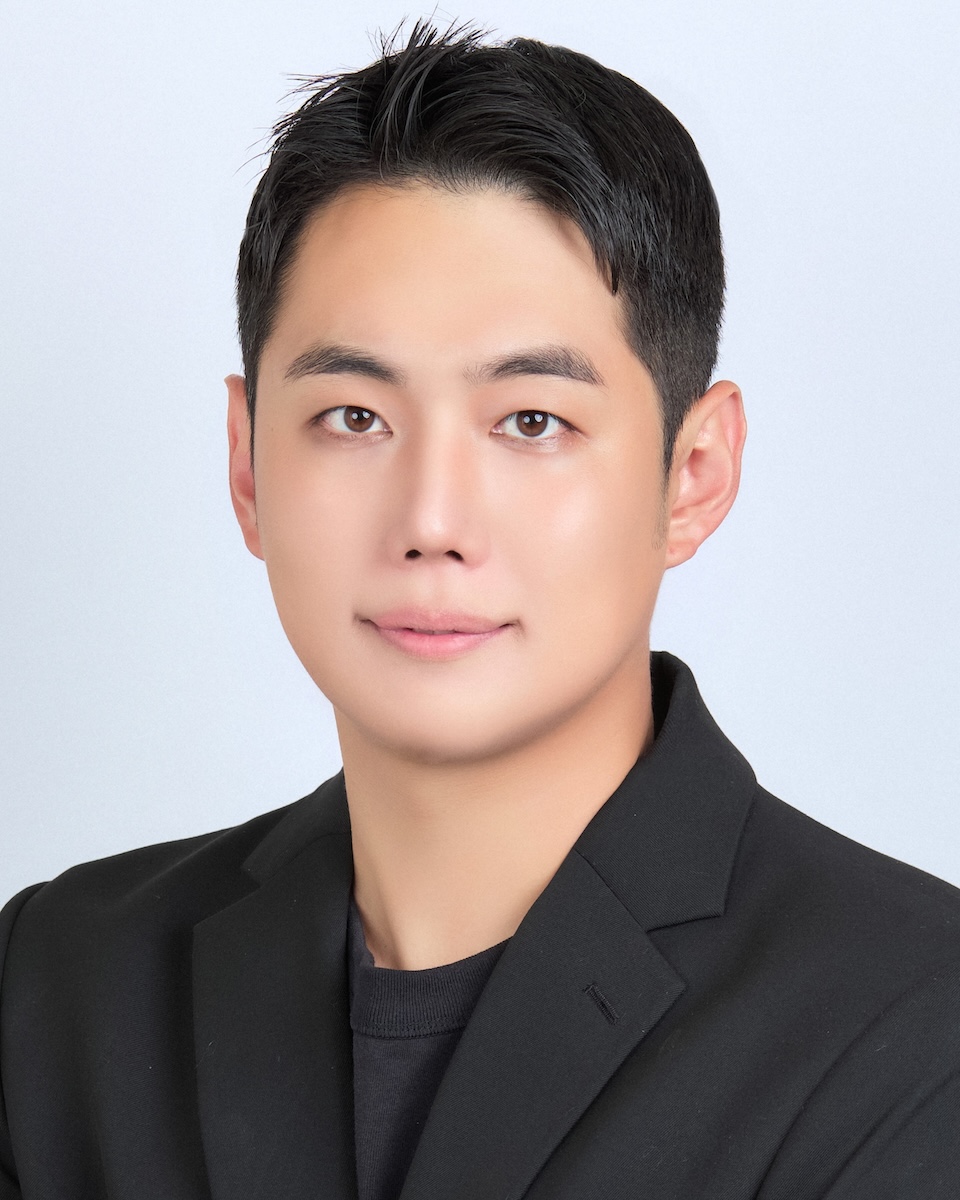}}]
{Mingi Jeong}~(Graduate Student Member, IEEE) is currently pursuing the Ph.D. degree of Reality and Robotics Lab with the Department of Computer Science, Dartmouth College, Hanover, NH, USA. 

His current interests are autonomous navigation, multirobot system, and maritime collision avoidance decision making.
\end{IEEEbiography}
\begin{IEEEbiography}[{\includegraphics[width=1in,height=1.25in,clip,keepaspectratio]{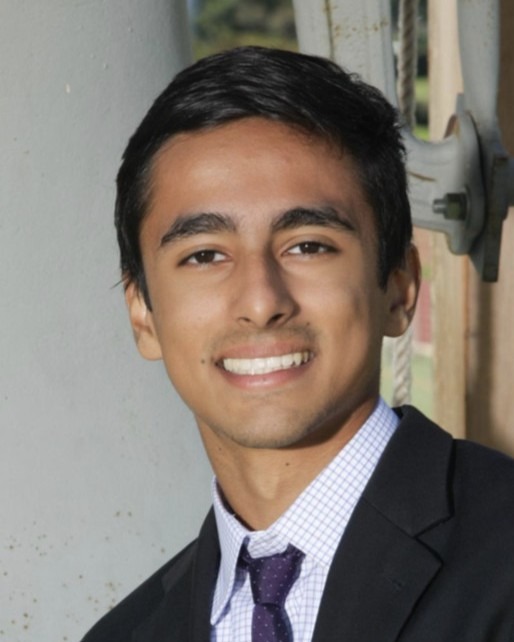}}]
{Arihant Chadda}~is a Senior Data Scientist, Applied Research at In-Q-Tel in Tysons, VA, USA. He received his Bachelors of Arts in Computer Science from Dartmouth College, where he also did research in the Reality and Robotics Lab. 

His current research interests are in developing robust digital and physical autonomous systems. 
\end{IEEEbiography}
\begin{IEEEbiography}[{\includegraphics[width=1in,height=1.25in,clip,keepaspectratio]{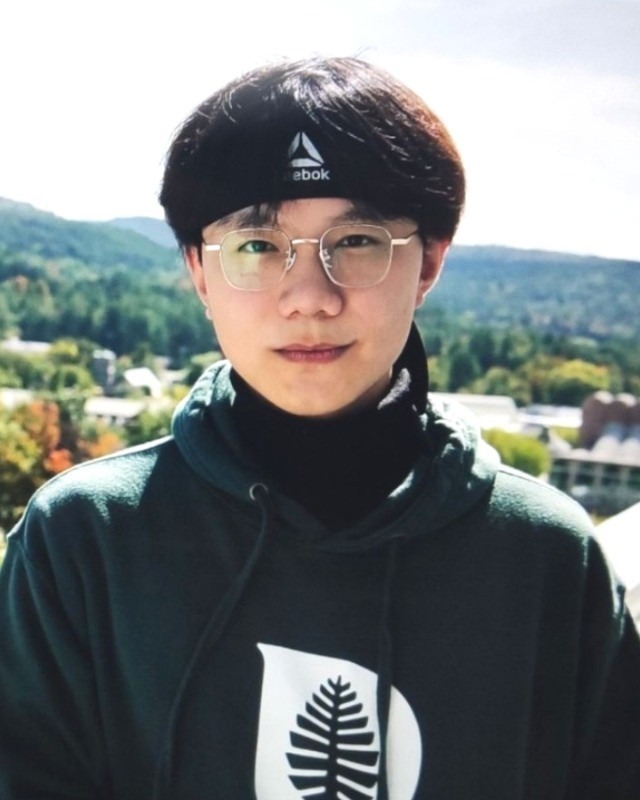}}]
{Ziang Ren}~is currently a Ph.D. student in computer science at MobileX lab at Columbia University. He received his M.S. degree at Dartmouth College, while working with Reality and Robotics Lab. 

His research interests are 3D vision, computational imaging, and robot perception.
\end{IEEEbiography}
\begin{IEEEbiography}[{\includegraphics[width=1in,height=1.25in,clip,keepaspectratio]{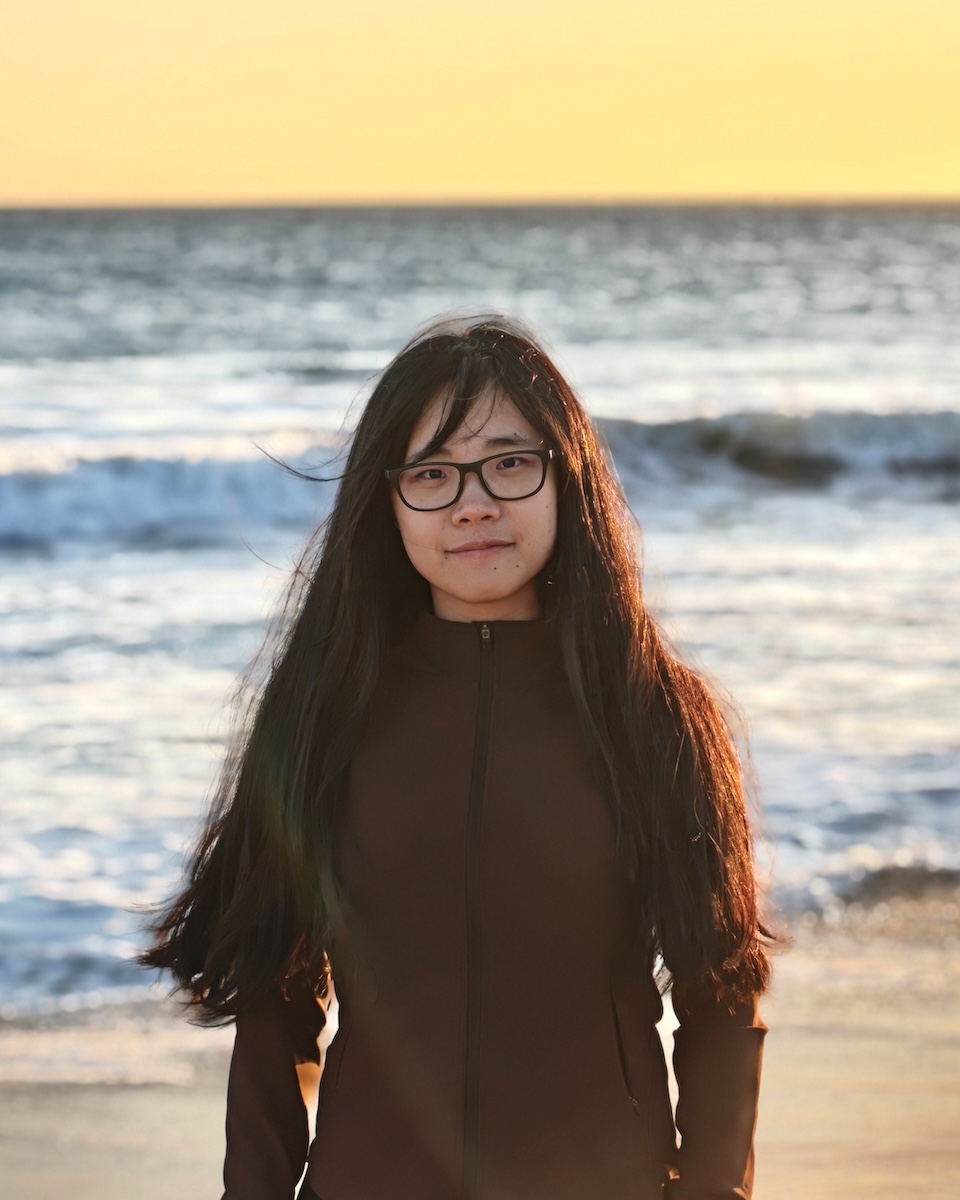}}]
{Luyang Zhao}~(Member, IEEE) received her Ph.D. degree in Computer Science from Dartmouth College, Hanover, NH, USA. She will join the Department of Electrical and Computer Engineering at Clemson University, Clemson, SC, USA, as an Assistant Professor. 

Her research interests include soft modular robotics, computational design, robot perception, and learning-based control. 
\end{IEEEbiography}

\begin{IEEEbiography}[{\includegraphics[width=1in,height=1.25in,clip,keepaspectratio]{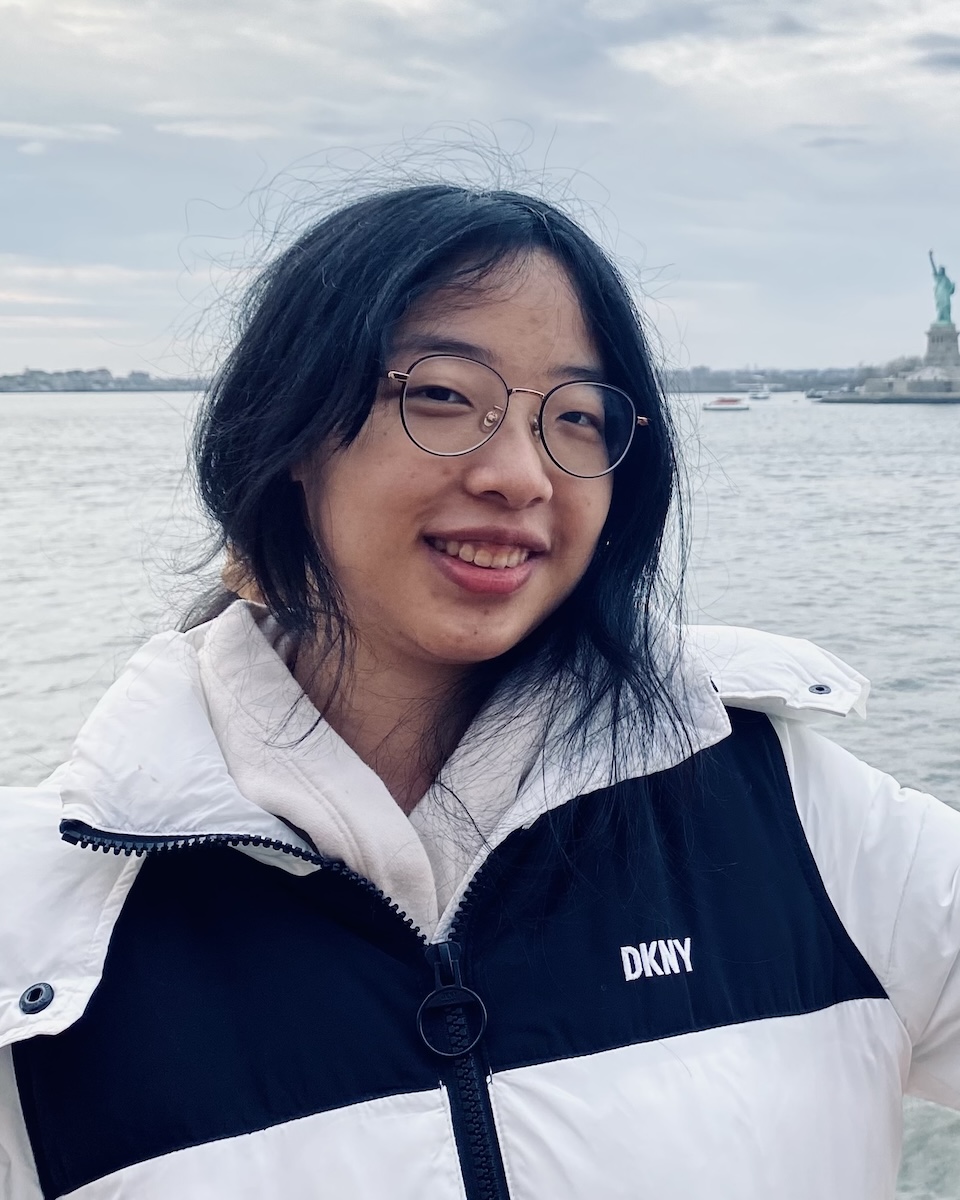}}]
{Haowen Liu}~is currently a Ph.D. student in computer science at University of Maryland College Park. She received her M.S. degree at Dartmouth College, while working with Reality and Robotics Lab. 

Her research interests are in video understanding and embodied AI.
\end{IEEEbiography}

\begin{IEEEbiography}[{\includegraphics[width=1in,height=1.25in,clip,keepaspectratio]{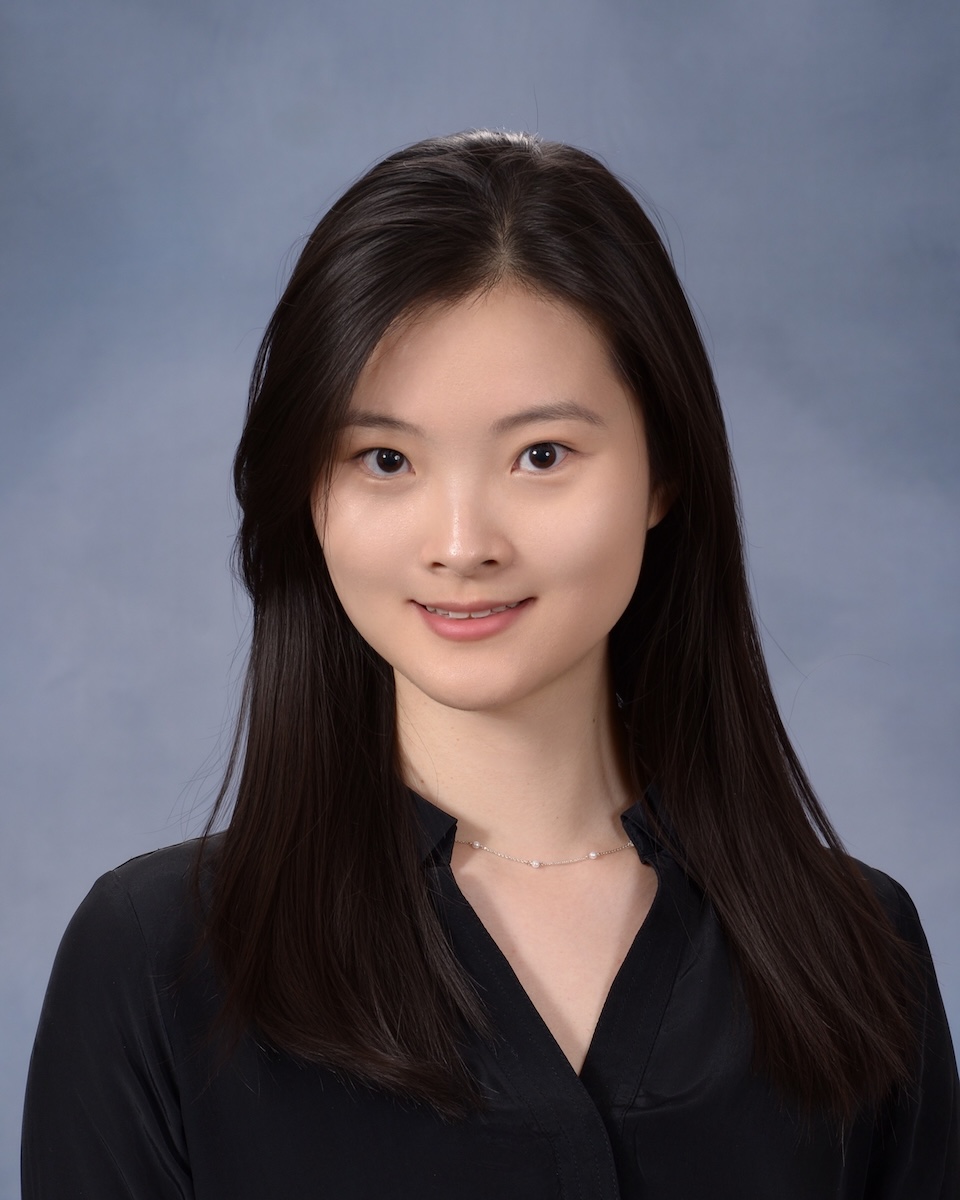}}]
{Aiwei Zhang}~received her Bachelors of Arts in Computer Science from Dartmouth College, Hanover NH, USA where she did research in the Reality and Robotics Lab. 

Her current research interests include social computing and machine learning with applications in healthcare, community-centered mental health support, and digital well-being. 
\end{IEEEbiography}
\begin{IEEEbiography}[{\includegraphics[width=1in,height=1.25in,clip,keepaspectratio]{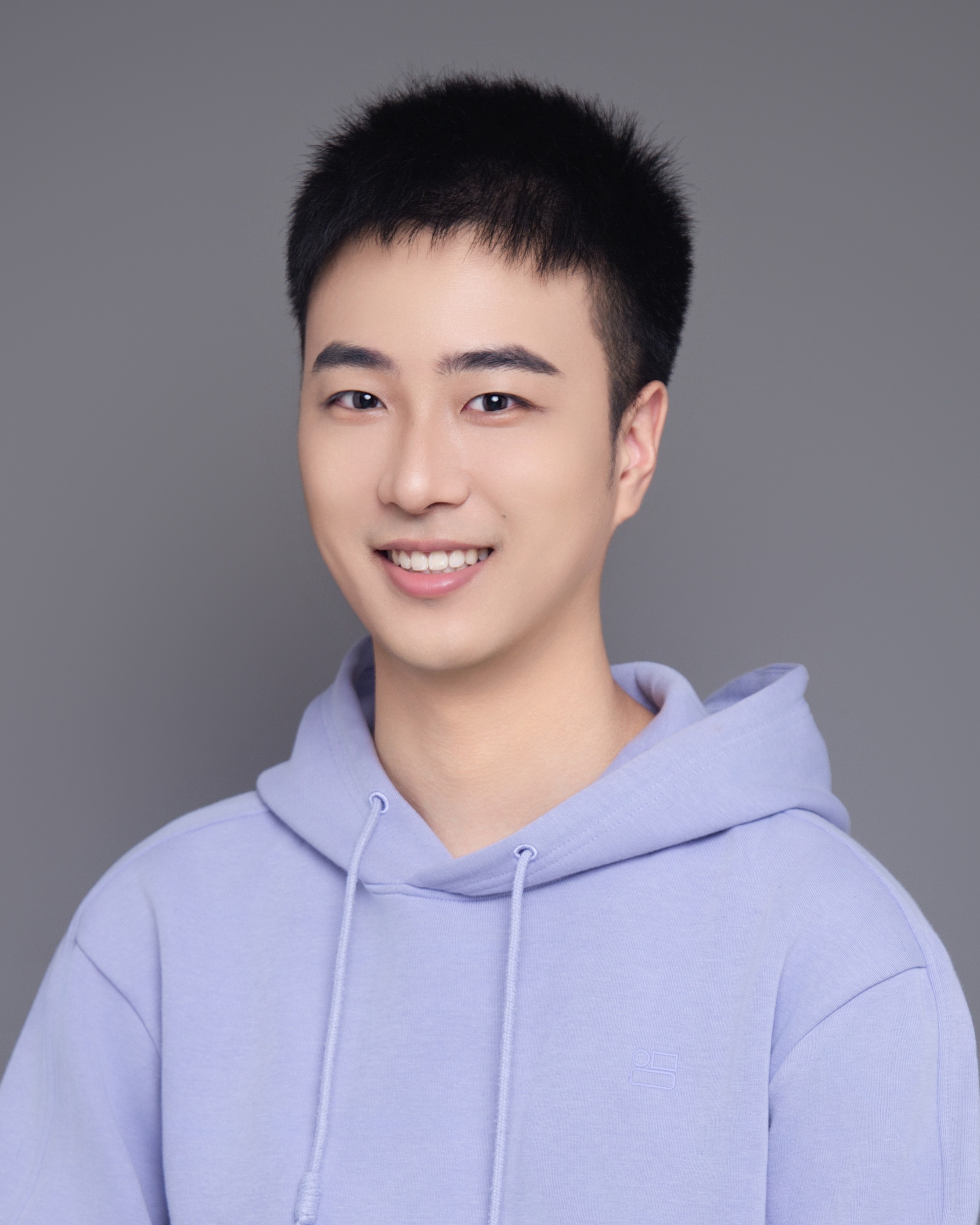}}]
{Yitao Jiang}~is currently a Ph.D. student Dartmouth College, Hanover NH, USA. He received a Master of Engineering Management from Dartmouth's Thayer School of Engineering.

His research focuses on actuators, modular tensegrity and soft robotic systems, swarm robotics driven by large language models, and robots inspired by biological locomotion. 

\end{IEEEbiography}

\begin{IEEEbiography}[{\includegraphics[width=1in,height=1.25in,clip,keepaspectratio]{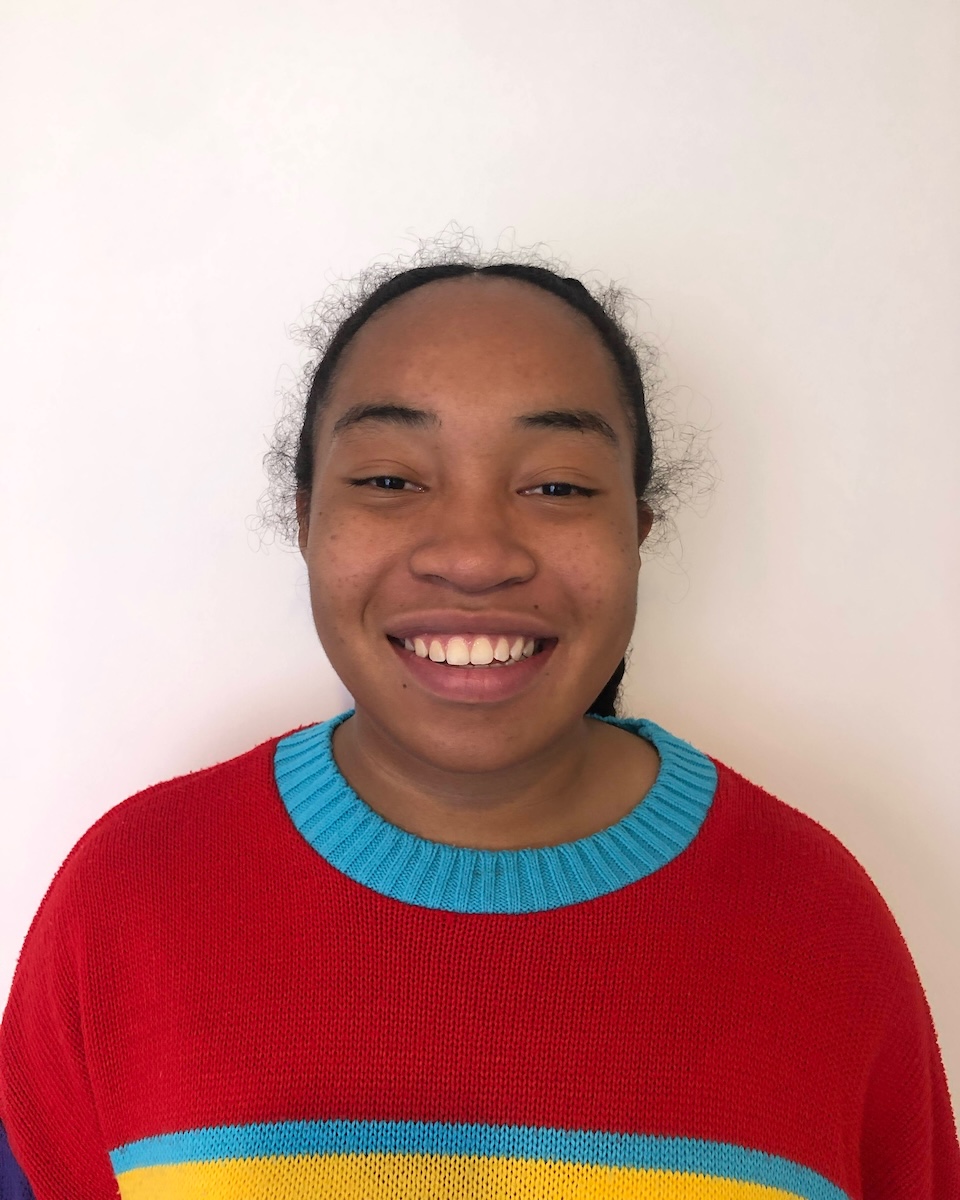}}]
{Sabriel Achong}~is currently pursuing a bachelor’s degree in Economics at Dartmouth College, Hanover, NH, USA. 

Her current interests include artificial intelligence, applied statistics, and agent-based modeling.
\end{IEEEbiography}

\begin{IEEEbiography}[{\includegraphics[width=1in,height=1.25in,clip,keepaspectratio]{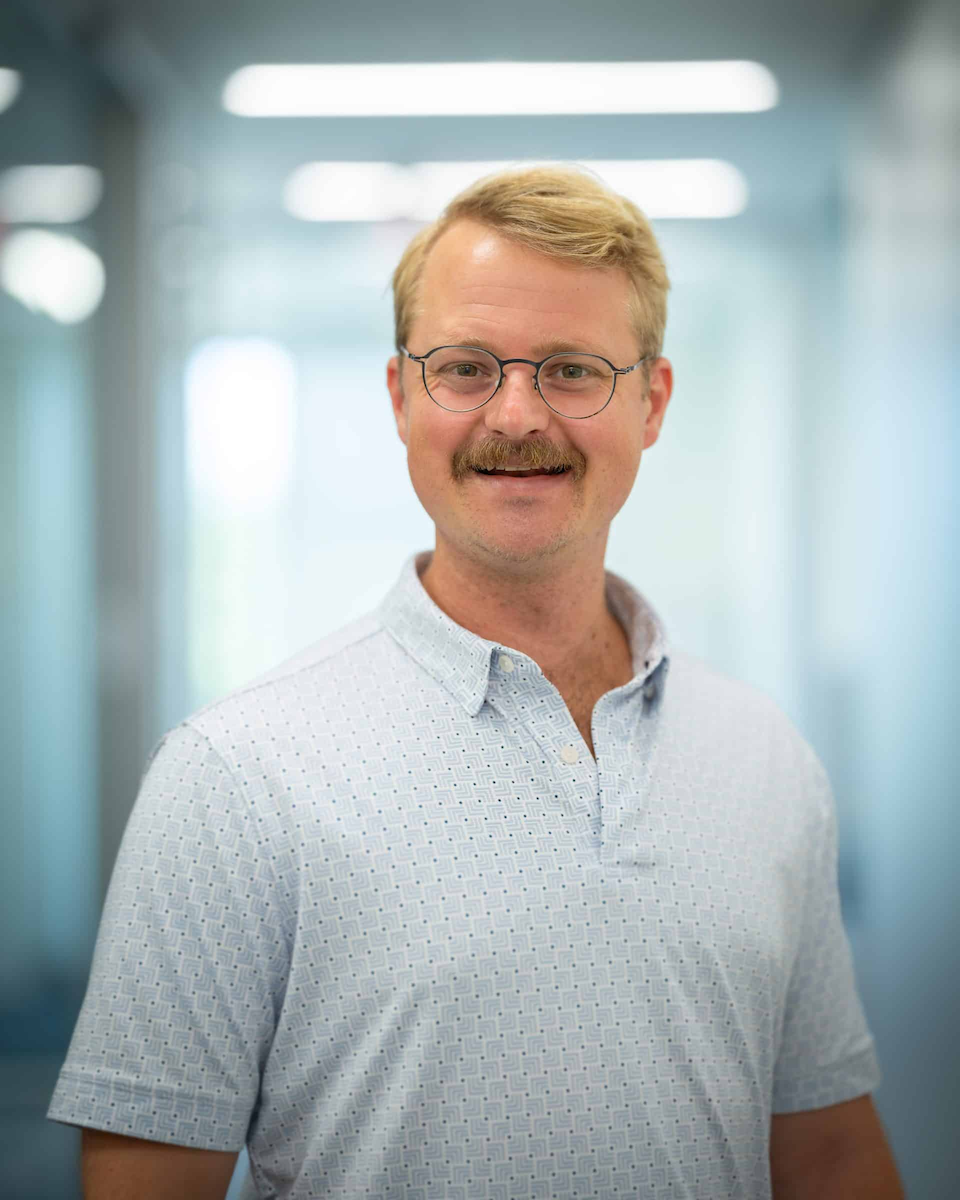}}]
{Samuel Lensgraf}~is a Research Scientist at the Florida Institute for Human and Machine Cognition where he works on autonomous underwater systems. He received his Ph.D. in Computer Science from Dartmouth College, Hanover, NH, USA.

His research interests include bimanual underwater manipulation and underwater human-machine teaming.
\end{IEEEbiography}

\begin{IEEEbiography}[{\includegraphics[width=1in,height=1.25in,clip,keepaspectratio]{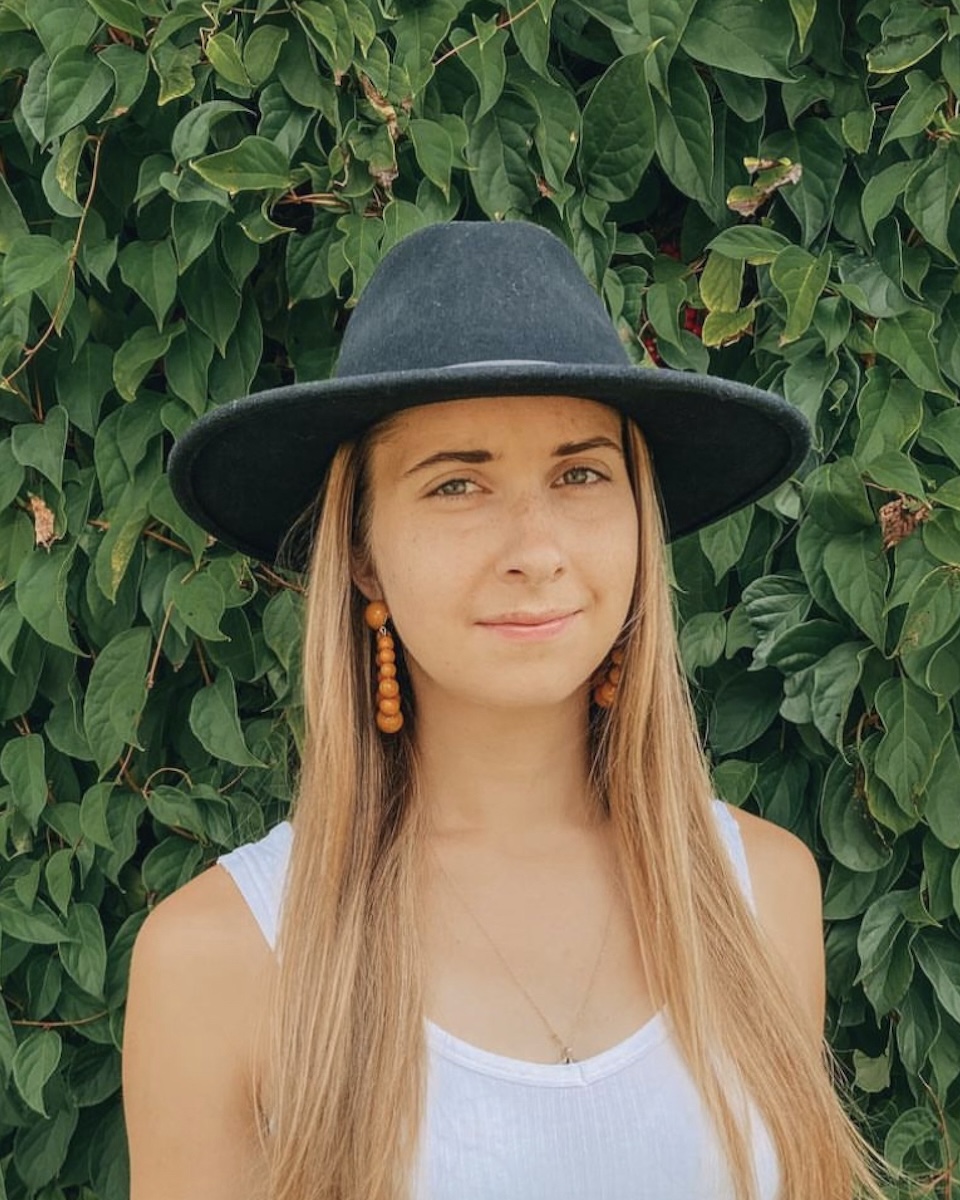}}]
{Monika Roznere}~received her Ph.D. degree in Computer Science from Dartmouth College in Hanover, NH, USA.

She is an Assistant Professor in the School of Computing at Binghamton University in Binghamton, NY, USA, and leads the Marine Robotics Lab. Her research interests include active perception, sensor fusion, and autonomous exploration.
\end{IEEEbiography}

\begin{IEEEbiography}[{\includegraphics[width=1in,height=1.25in,clip,keepaspectratio]{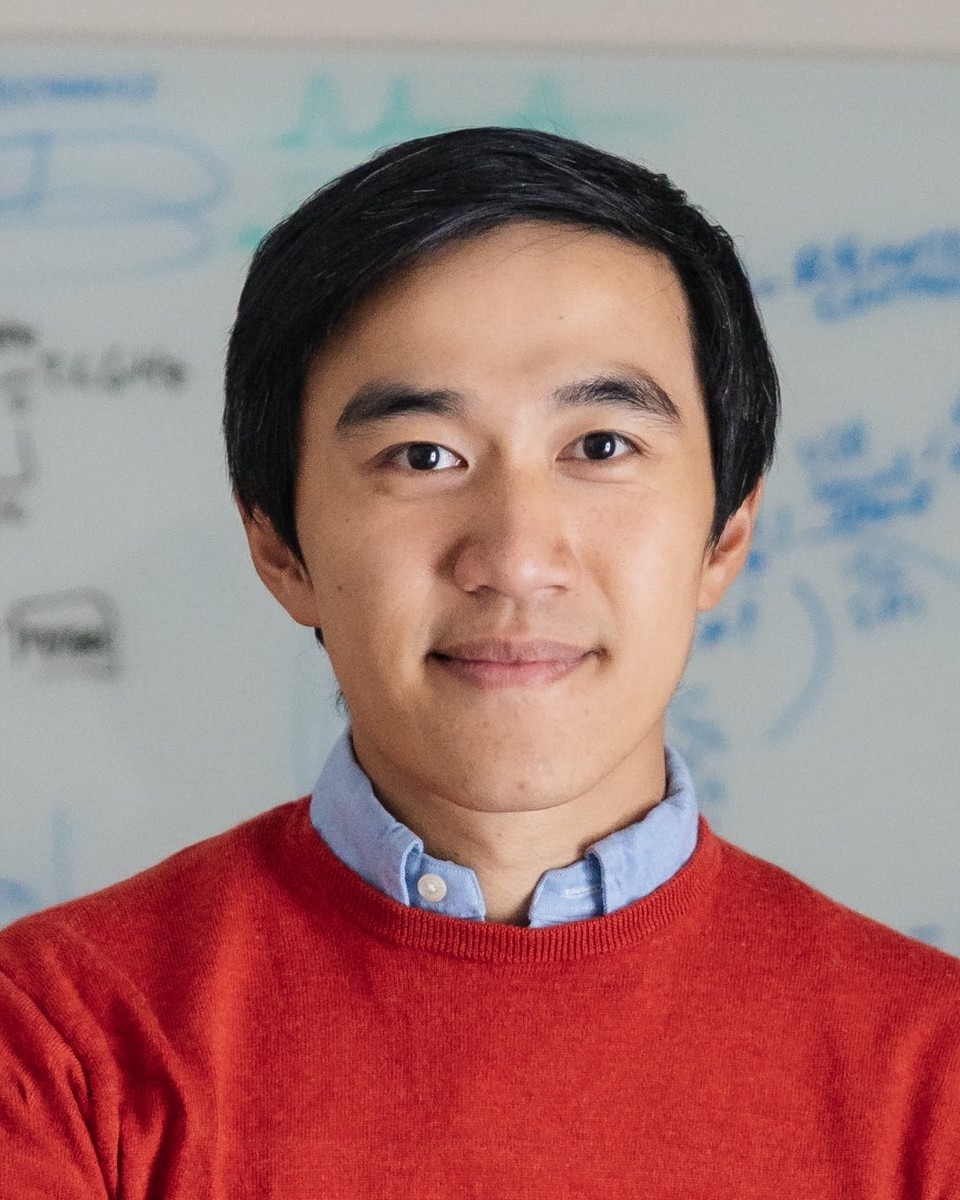}}]
{Alberto Quattrini Li}~received the Ph.D. degree in computer science and engineering from Polytechnic of Milan, Milan, Italy. He is an Associate Professor with the Computer Science Department at Dartmouth College in Hanover, NH, USA, and a Co-Director of the Reality and Robotics Lab. 

His research interests include autonomous mobile robotics, artificial intelligence, and agents and multiagent systems.
\end{IEEEbiography}

\vfill\pagebreak

\end{document}